\documentclass[letterpaper, 10 pt, conference]{ieeeconf}  

\IEEEoverridecommandlockouts                              
\usepackage[utf8]{inputenc}
\usepackage[T1]{fontenc}
\usepackage{amsmath}
\usepackage{multirow}
\usepackage[export]{adjustbox}

\usepackage[table,xcdraw]{xcolor}
\usepackage{amssymb}

\definecolor{rblue}{rgb}{0,0.5,1}
\definecolor{awesome}{rgb}{1.0, 0.13, 0.32}
\definecolor{hollywoodcerise}{rgb}{0.96, 0.0, 0.63}
\definecolor{lasallegreen}{rgb}{0.03, 0.47, 0.19}
\definecolor{hanpurple}{rgb}{0.32, 0.09, 0.98}
\definecolor{green(pigment)}{rgb}{0.0, 0.65, 0.31}

\definecolor{mygray}{gray}{.9}
\usepackage{booktabs}

\usepackage{graphicx}
\usepackage{array}
\usepackage{caption}
\usepackage{subcaption}

\makeatletter
\let\NAT@parse\undefined
\makeatother
\usepackage[pagebackref=false,breaklinks=true,colorlinks,bookmarks=false]{hyperref}
\hypersetup{colorlinks=true,linkcolor={red},citecolor={hanpurple},urlcolor={magenta}}

\title{\LARGE \bf
Spherical-GOF: Geometry-Aware Panoramic Gaussian Opacity Fields for 3D Scene Reconstruction}

\author{Zhe Yang$^{1}$, Guoqiang Zhao$^{1}$, Sheng Wu$^{1}$, Kai Luo$^{1}$, and Kailun Yang$^{1,2,\dag}$%
\thanks{This work was supported in part by the National Natural Science Foundation of China (Grant No. 62473139), in part by the Hunan Provincial Research and Development Project (Grant No. 2025QK3019), and in part by the State Key Laboratory of Autonomous Intelligent Unmanned Systems (the opening project number ZZKF2025-2-10).}
\thanks{$^{1}$The authors are with the School of Artificial Intelligence and Robotics, Hunan University, China (email: kailun.yang@hnu.edu.cn).}%
\thanks{$^{2}$The author is also with the National Engineering Research Center of Robot Visual Perception and Control Technology, Hunan University, China.}%
\thanks{$^{\dag}$Corresponding author: Kailun Yang.}
}

\let\oldtwocolumn\twocolumn
\renewcommand\twocolumn[1][]{%
    \oldtwocolumn[{#1}{
    \begin{center}
    \vskip -3ex
        \centering
        \hspace*{-6mm}\includegraphics[width=0.99\textwidth]{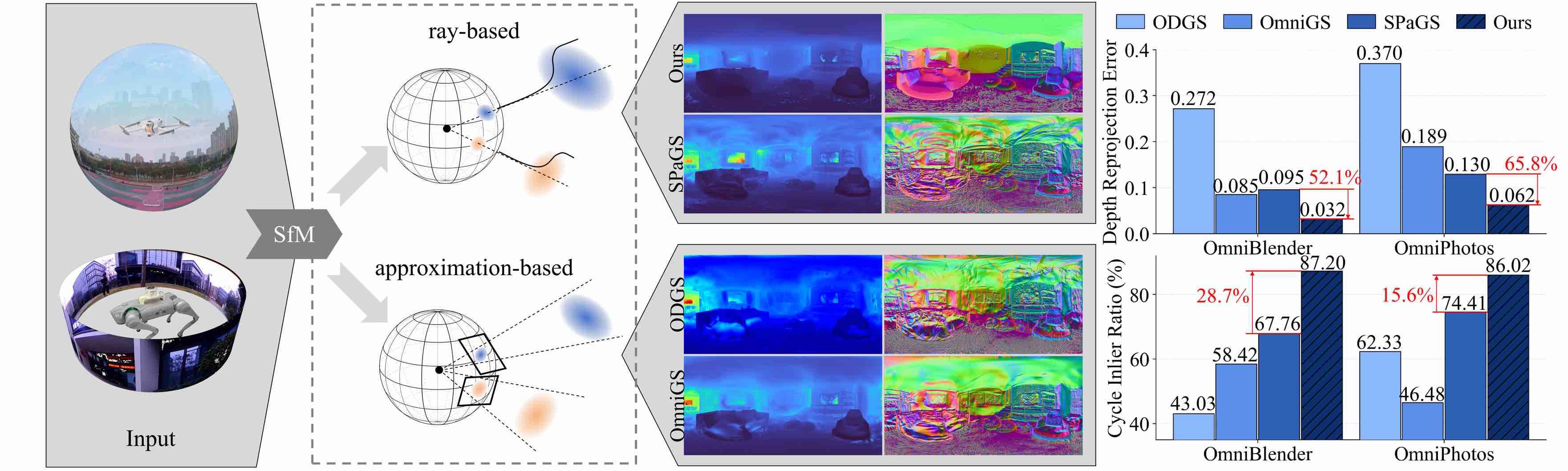}
        \vskip -1ex
        \captionof{figure}{{Our ray casting-based method enables accurate panoramic Gaussian rendering from an input panorama and SfM poses. 
        Compared to projection-based approaches (ODGS~\cite{lee2024odgs} and OmniGS~\cite{li2025omnigs}) and prior ray-based methods (SPaGS~\cite{li2025spags}), our method produces cleaner and more geometrically consistent depth and normals, avoiding texture-like ripple artifacts.
        }}
        \label{fig:teaser}
    \end{center}
    }]
}

\begin{document}

\maketitle
\thispagestyle{empty}
\pagestyle{empty}

\begin{abstract}
Omnidirectional images are increasingly used in robotics and vision due to their wide field of view. However, extending 3D Gaussian Splatting (3DGS) to panoramic camera models remains challenging, as existing formulations are designed for perspective projections and naive adaptations often introduce distortion and geometric inconsistencies.

We present Spherical-GOF, an omnidirectional Gaussian rendering framework built upon Gaussian Opacity Fields (GOF). Unlike projection-based rasterization, Spherical-GOF performs GOF ray sampling directly on the unit sphere in spherical ray space, enabling consistent ray-Gaussian interactions for panoramic rendering. To make the spherical ray casting efficient and robust, we derive a conservative spherical bounding rule for fast ray-Gaussian culling and introduce a spherical filtering scheme that adapts Gaussian footprints to distortion-varying panoramic pixel sampling.

Extensive experiments on standard panoramic benchmarks (OmniBlender and OmniPhotos) demonstrate competitive photometric quality and substantially improved geometric consistency. Compared with the strongest baseline, Spherical-GOF reduces depth reprojection error by $57\%$ and improves cycle inlier ratio by $21\%$. Qualitative results show cleaner depth and more coherent normal maps, with strong robustness to global panorama rotations. We further validate generalization on OmniRob, a real-world robotic omnidirectional dataset introduced in this work, featuring UAV and quadruped platforms. The source code and the OmniRob dataset will be released at \url{https://github.com/1170632760/Spherical-GOF}.
\end{abstract}

 
\section{Introduction}
Advances in robotics and embodied intelligence, together with the rise of AR/VR and digital-twin applications, have made 3D scene reconstruction increasingly important for perception, simulation, and environment understanding~\cite{selvaratnam20253d,bao20253d}.
Accurate geometric representations are essential for building realistic simulation environments and digital twins, enabling perception-driven analysis and supporting downstream embodied AI and human-robot interaction research~\cite{melnik2025digital,jin2024gs}.
Conventional reconstruction pipelines~\cite{schonberger2016structure,moulon2016openmvg} relying on pinhole cameras typically require the acquisition of extensive image datasets, followed by the recovery of feature-based point cloud maps via Structure-from-Motion (SfM). 
Panoramic cameras offer an expansive Field of View (FoV), enabling efficient 360{\textdegree} scene coverage with far fewer images under practical sensing constraints, and have therefore attracted increasing attention in robotics and embodied vision~\cite{ai2025survey_learning,huang2022360vo}.

Simultaneously, methods based on Neural Radiance Fields (NeRF)~\cite{mildenhall2021nerf} and 3DGS~\cite{kerbl20233d} have garnered significant attention due to their ability to not only recover scene geometry but also render high-fidelity photorealistic textures. 
Consequently, extending these paradigms to support panoramic cameras has become a focal point of research. 
Given the inherent ray-sampling nature of NeRF, adapting it to panoramic models is relatively straightforward, and several studies have demonstrated impressive performance in this domain~\cite{muller2022instant,chen2022tensorf,choi2023balanced}. 
However, such methods inevitably inherit NeRF's intrinsic limitations, specifically low rendering efficiency and protracted training times. Conversely, while 3DGS offers a significant leap in rendering speed, adapting it to panoramic imaging presents substantial challenges. 
Standard 3DGS~\cite{kerbl20233d}explicitly represents scenes using 3D Gaussians and relies on a projection mechanism tailored for pinhole camera models. 
Specifically, the projected covariance is computed by applying the Jacobian of the camera model to linearly approximate the inherently non-linear projection, under the assumption that a projected 3D Gaussian remains an ellipsoidal Gaussian on the image plane. 
Such geometric assumptions substantially impede the direct extension of 3D Gaussian Splatting to panoramic imagery; moreover, even spherical-aware splatting variants that achieve strong visual quality often prioritize optical appearance, leaving geometry-oriented accuracy less explicitly emphasized~\cite{lee2024odgs,li2025omnigs,li2025spags}. 

To address these challenges, we propose Spherical-GOF, a ray-space framework for omnidirectional Gaussian rendering.
Building on GOF~\cite{yu2024gaussian}, Spherical-GOF performs ray sampling directly on the unit sphere, enabling projection-consistent ray-Gaussian interactions for panoramic imaging without relying on planar projection approximations.
To ensure robust and efficient rendering, we further derive a conservative spherical bounding strategy for Gaussian primitives, which supports reliable ray–Gaussian culling in omnidirectional settings. 
In addition, we introduce a spherical filtering scheme that adapts Gaussian footprints to panoramic pixel sampling, effectively mitigating aliasing artifacts induced by spherical distortion and improving rendering stability. 
Quantitatively, Spherical-GOF reduces depth reprojection error by 57\% and improves cycle consistency by 21\% over the strongest baseline, while maintaining competitive rendering performance. 
These improvements translate into more accurate geometry and more reliable mesh extraction, making the method better suited for downstream tasks that require consistent surface reconstruction.

In summary, this paper makes the following contributions:
\begin{itemize}
\item We propose \textbf{Spherical-GOF}, a spherical ray-space GOF sampling framework for equirectangular projection (ERP) panoramas, which improves geometric reconstruction accuracy of omnidirectional Gaussian rendering by avoiding local linearization errors introduced by planar projection.
\item We introduce a panoramic filter and sphere-metric-consistent geometric regularization that stabilize training and reduce the influence of high-frequency appearance textures on geometry, leading to cleaner depth and more coherent normal estimates.
\item We conduct extensive experiments on public panoramic benchmarks to evaluate both photometric and geometric quality, and further validate camera-model adaptation on our robot-captured omnidirectional data, showing that our formulation can be applied to diverse omnidirectional camera setups with only minor modifications.
\end{itemize}

\section{Related Work}
\noindent\textbf{3D Gaussian splatting.}
Recent advances in 3D reconstruction have been largely driven by NeRF~\cite{mildenhall2021nerf} and 3DGS~\cite{kerbl20233d}, both of which enable high-quality novel view synthesis. 
In particular, 3DGS has gained popularity for its real-time rendering speed. It represents a scene as a set of anisotropic 3D Gaussians with learnable attributes such as scale, rotation, opacity, and color, and renders them efficiently via rasterization. However, this efficiency relies on a local affine approximation of the projection, which can become unreliable for wide-FoV or highly distorted camera models.

To improve rendering correctness and geometric reliability, several works modify the Gaussian representation or evaluation strategy. 
For example, 2D Gaussian Surfels (2DGS)~\cite{huang20242d} represent primitives as local surface patches, enabling more accurate depth and normal estimation. 
GOF~\cite{yu2024gaussian} instead formulates Gaussian rendering in a volumetric occupancy-style manner, producing well-defined depth and normal outputs. 
HTGS~\cite{hahlbohm2025efficient} further improves 3DGS by introducing perspective-correct, ray-based splat evaluation that avoids matrix inversion and stabilizes optimization. 
These developments highlight the importance of projection-consistent formulations, especially when extending Gaussian-based reconstruction to omnidirectional panoramic imagery.

\noindent\textbf{Panoramic 3D reconstruction.}
Thanks to the wide FoV of panoramic cameras~\cite{ai2025survey_learning}, panorama-based 3D reconstruction has received increasing attention. 
In NeRF, extending volumetric rendering to panoramic imagery is relatively straightforward since radiance is evaluated along rays; representative examples include EgoNeRF~\cite{choi2023balanced} and other NeRF-based methods for omnidirectional images~\cite{huang2022real,kulkarni2023360fusionnerf}.
In contrast, adapting 3DGS to panoramic images is more challenging because its rasterization pipeline is tightly coupled with the camera projection model and typically assumes that a 3D Gaussian remains an elliptical 2D Gaussian after projection.

To support panoramic rendering, some methods derive the Jacobian of panoramic projection and propagate covariance accordingly, such as OmniGS~\cite{li2025omnigs}, ErpGS~\cite{ito2025erpgs}, and 360-GS~\cite{bai2025360}. 
However, they still rely on a local affine approximation, which can break down in highly distorted regions such as the polar areas of equirectangular panoramas and lead to projection inconsistencies. 
To avoid a single global projection, other approaches render on intermediate surfaces. For instance, ODGS~\cite{lee2024odgs} projects Gaussians onto locally defined tangent planes and maps them back to the panorama, while face-based methods render on cubemap-like piecewise perspective surfaces, optionally introducing additional transition planes to reduce cross-face discontinuities, before mapping back to the panorama~\cite{shen2025you}.
While such strategies alleviate severe distortion, they may introduce additional overhead and can suffer from approximation/stitching issues across pieces.

More recently, SPaGS~\cite{li2025spags} extends the ray casting formulation of HTGS~\cite{hahlbohm2025efficient} to spherical panoramas via omnidirectional ray-splat intersection and bounding-box-based rasterization, enabling projection-consistent panoramic rendering without local affine projection approximations. 
Different from SPaGS, which mainly improves projection-correct splat evaluation under spherical camera models, our method extends the GOF formulation to spherical ray space. 
By deriving depth and normal estimates from a GOF-style volumetric opacity formulation, our method focuses more on geometry-consistent panoramic reconstruction rather than only projection-correct panoramic rendering.

\section{Methodology}
\subsection{Preliminary: Classical Gaussian Splatting and Gaussian Opacity Fields}
3DGS is an emerging approach to 3D reconstruction. It takes as input a set of posed images and a sparse point cloud, which is typically obtained from SfM; it initializes a collection of anisotropic 3D Gaussians and optimizes their parameters by differentiable rendering under a photometric reconstruction loss. Each Gaussian $i$ is parameterized by its mean position $\boldsymbol{\mu}_i\in\mathbb{R}^3$, color $\mathbf{c}_i$, opacity $o_i$, and covariance $\boldsymbol{\Sigma}_i\in\mathbb{R}^{3\times 3}$.
The covariance of each Gaussian is commonly represented using a rotation matrix $\mathbf{R}_i\in SO(3)$ and per-axis scales $\mathbf{s}_i\in\mathbb{R}^3$. Let $\mathbf{S}_i=\mathrm{diag}(\mathbf{s}_i)$ denote the scaling matrix. The covariance is then given by
\begin{align}
\boldsymbol{\Sigma}_i=\mathbf{R}_i\mathbf{S}_i\mathbf{S}_i^{\top}\mathbf{R}_i^{\top}.
\end{align}
Each Gaussian defines a 3D density
\begin{align}
G_i(\mathbf{x})=\exp\!\left(-\frac{1}{2}(\mathbf{x}-\boldsymbol{\mu}_i)^{\top}\boldsymbol{\Sigma}_i^{-1}(\mathbf{x}-\boldsymbol{\mu}_i)\right).
\end{align}
Since perspective projection is inherently non-linear, classical 3DGS adopts the EWA-splatting formulation~\cite{zwicker2002ewa}, which approximates the projected footprint by a 2D Gaussian via local first-order linearization. Let $\mathbf{x}_i^c$ denote the mean transformed to the camera coordinate system, and let $\pi(\cdot)$ be the perspective projection. The Jacobian of the projection evaluated at $\mathbf{x}_i^c$ is denoted as $\mathbf{J}_i\in\mathbb{R}^{2\times 3}$. The screen-space covariance is computed as
\begin{align}
\boldsymbol{\Sigma}_i^{'}
= \mathbf{J}_i\,
\Bigl(\mathbf{R}^{c}_{w}\,\boldsymbol{\Sigma}_i\,(\mathbf{R}^{c}_{w})^{\top}\Bigr)\,
\mathbf{J}_i^{\top}.
\end{align}
where $\mathbf{R}^{c}_{w}$ denotes the rotation from the world coordinate system to the camera coordinate system.

The resulting 2D Gaussian footprint at pixel coordinate $\mathbf{u}\in\mathbb{R}^2$ is
\begin{align}
G_i^{'}(\mathbf{u})
= \exp\left(
-\frac{1}{2}
(\mathbf{u}-\boldsymbol{\mu}_i^{'})^{\top}
(\boldsymbol{\Sigma}_i^{'})^{-1}
(\mathbf{u}-\boldsymbol{\mu}_i^{'})
\right),
\end{align}
where $\boldsymbol{\mu}_i^{'}=\pi(\mathbf{x}_i^c)$ denotes the projected mean.
Finally, the pixel color is computed as:
\begin{equation}
\mathbf{C}(\mathbf{u})=\sum_{j=1}^{N}\mathbf{c}_{j}\,\alpha_{j}(\mathbf{u})\,T_{j}(\mathbf{u}).
\end{equation}
Here, $\alpha_{j}(\mathbf{u})=o_{j}\,G_{j}^{'}(\mathbf{u})$, and 
$T_{j}(\mathbf{u})=\prod_{k=1}^{j-1}\left(1-\alpha_{k}(\mathbf{u})\right)$ denotes the accumulated transmittance.

\subsection{Panoramic Rendering via Gaussian Opacity Field}
Gaussian Opacity Fields ~\cite{yu2024gaussian} offer a ray-based rendering formulation that avoids screen-space projection approximations. GOF evaluates each Gaussian's opacity accumulation directly along camera rays, improving geometric consistency and yielding more accurate geometry estimation.

Specifically, given the camera center $\mathbf{o}\in\mathbb{R}^3$ and a ray direction $\mathbf{r}\in\mathbb{R}^3$, 
any point along the ray can be written as $\mathbf{x}=\mathbf{o}+t\mathbf{r}$, where $t$ denotes the ray depth. 
For a 3D Gaussian $i$ with mean $\mathbf{x}_i$, rotation $\mathbf{R}_i$, and scaling matrix $\mathbf{S}_i=\mathrm{diag}(\mathbf{s}_i)$, 
we transform the ray into the local coordinate frame of the Gaussian,
\begin{align}
\mathbf{o}_{i}&=\mathbf{S}_{i}^{-1}\mathbf{R}_{i}(\mathbf{o}-\mathbf{x}_{i}),\\
\mathbf{r}_{i}&=\mathbf{S}_{i}^{-1}\mathbf{R}_{i}\mathbf{r},\\
\mathbf{x}_{i}(t)&=\mathbf{o}_{i}+t\mathbf{r}_{i}.
\end{align}
Substituting the point in the Gaussian local frame into the Gaussian formulation yields the 1D response along the ray:
\begin{align}
\mathcal{G}_{i}^{1\mathrm{D}}(t)
&= \exp\!\left(-\frac{1}{2}\,\mathbf{x}_{i}(t)^{\top}\mathbf{x}_{i}(t)\right)
\notag\\
&= \exp\!\left(
-\frac{1}{2}
\left(
\mathbf{r}_{i}^{\top}\mathbf{r}_{i}\, t^{2}
+ 2\,\mathbf{o}_{i}^{\top}\mathbf{r}_{i}\, t
+ \mathbf{o}_{i}^{\top}\mathbf{o}_{i}
\right)
\right). \label{eq:ray}
\end{align}
According to Eq.~\eqref{eq:ray}, the exponent is a quadratic function of $t$, whose maximum is attained at 
$t^{*}=-\frac{B}{A}$, where $A=\mathbf{r}_{i}^{\top}\mathbf{r}_{i}$ and $B=\mathbf{o}_{i}^{\top}\mathbf{r}_{i}$.

Since GOF evaluates Gaussian contributions along rays, the rendering no longer depends on a specific projection model. This property enables a seamless extension to panoramic imaging without introducing projection-approximation errors.
We define the camera coordinate system such that the $z$-axis points forward and the $x$-axis points to the right, consistent with the pinhole convention. 
For an equirectangular panorama, the longitude and latitude of a 3D point $\mathbf{x}$ are defined as
$\varphi=\mathrm{atan2}(x_x,x_z)\in[-\pi,\pi]$ and
$\phi=\arcsin\!\left(-x_y/\|\mathbf{x}\|_2\right)\in[-\pi/2,\pi/2]$.
Using these definitions, the panoramic projection is given by
\begin{equation}
\mathrm{Proj}(\mathbf{x})=\left(
\frac{W}{2\pi}\,\varphi + \frac{W}{2},
-\frac{H}{\pi}\,\phi + \frac{H}{2}
\right)^{\top}.
\end{equation}
In addition, during preprocessing, it is necessary to determine the tile range influenced by each Gaussian for efficient rendering. 
In the original 3DGS pipeline, this range is estimated using the projection Jacobian, which is not suitable for panoramic imaging. 
Since directly computing the exact longitudinal and latitudinal extent of an anisotropic Gaussian on the panorama is challenging, we approximate each Gaussian as a sphere whose diameter is determined by its longest principal axis. 
We then compute conservative upper and lower bounds of the longitude and latitude covered by this sphere. 
Although this estimated range is looser than the exact projection, it guarantees that no valid ray-Gaussian contributions are clipped.

\subsection{Optimization Strategies for Panoramic Rendering}
Due to latitude-dependent distortion in panoramic projections, Gaussians with identical 3D size can occupy markedly different areas on the image plane depending on their latitude.
As a result, Gaussians at higher latitudes tend to accumulate much larger gradients than those at lower latitudes. 
To balance the splitting thresholds across different latitudes, we modify the Gaussian gradients to be latitude-dependent.
We define
\begin{equation}
w_{\text{lat}}=\operatorname{clamp}\!\left(\cos(\phi),\, \epsilon\right),
\label{eq:lat_weight}
\end{equation}
This weight is multiplied by the densification score to suppress excessive splitting near the poles.

For ERP panoramas, the angular resolution varies with latitude, so identically sized Gaussians may have very different pixel footprints. 
To avoid sub-pixel footprints that cause aliasing and instability, we assign each Gaussian an isotropic filter radius $f_i$ according to the panorama's angular resolution.
For each visible camera, we compute the camera-space distance $r=\lVert\mathbf{x}_{\mathrm{cam}}\rVert_2$ and the latitude
$\phi=\arcsin\!\left(-x_y/r\right)$.
Given an image resolution $W\times H$, the vertical and horizontal angular resolutions are
\begin{equation}
\Delta\theta_{\text{lat}}=\frac{\pi}{H}, \qquad
\Delta\theta_{\text{lon}}=\frac{2\pi}{W}\cos\phi.
\end{equation}
and we set $\Delta\theta=\max(\Delta\theta_{\text{lat}},\Delta\theta_{\text{lon}})$.
This yields a pixel-support candidate $f_{\text{cand}}=r\,\Delta\theta$.
For a single Gaussian primitive, we take the maximum over all visible cameras and apply a constant factor $\kappa$:
\begin{equation}
f=\kappa\max_{\text{cams}} f_{\text{cand}}.
\end{equation}

Given the per-axis scale $\mathbf{s}\in\mathbb{R}^3$, we inflate the Gaussian scale with an isotropic radius $f$:
\begin{equation}
\tilde{\mathbf{s}}=\sqrt{\mathbf{s}\odot\mathbf{s} + f^2\mathbf{1}},
\end{equation}
where $\odot$ denotes the Hadamard product and $\mathbf{1}\in\mathbb{R}^3$ is an all-ones vector. This inflation enforces a stable lower bound on the Gaussian extent and avoids sub-pixel footprints.

Since the inflation changes the Gaussian volume, we compensate the opacity to preserve density consistency:
\begin{equation}
o \leftarrow o \cdot
\sqrt{\frac{\prod(\mathbf{s}\odot\mathbf{s})}{\prod(\tilde{\mathbf{s}}\odot\tilde{\mathbf{s}})}},
\end{equation}
where $\prod(\cdot)$ denotes the product of vector components. We apply this procedure to all Gaussians, which maintains sufficient image support in distant or high-latitude regions and suppresses numerical instability and aliasing.

\begin{table*}[!t]
\centering
\vskip-1ex
\footnotesize
\setlength{\tabcolsep}{3pt}
\renewcommand{\arraystretch}{1.08}
\resizebox{\textwidth}{!}{%
\begin{tabular}{l|ccccc|ccccc|ccccc}
\toprule
& \multicolumn{5}{c|}{OmniBlender-Indoor~\cite{choi2023balanced}}
& \multicolumn{5}{c|}{OmniBlender-Outdoor~\cite{choi2023balanced}}
& \multicolumn{5}{c}{OmniPhotos~\cite{bertel2020omniphotos}} \\
\cmidrule(lr){2-6} \cmidrule(lr){7-11} \cmidrule(lr){12-16}
Method & 
DRE$\downarrow$ & CIR$\uparrow$ & PSNR$\uparrow$ & SSIM$\uparrow$ & LPIPS$\downarrow$ & 
DRE$\downarrow$ & CIR$\uparrow$
& PSNR$\uparrow$ & SSIM$\uparrow$ & LPIPS$\downarrow$ & 
DRE$\downarrow$ & CIR$\uparrow$ 
& PSNR$\uparrow$ & SSIM$\uparrow$ & LPIPS$\downarrow$\\
\midrule
EgoNeRF~\cite{choi2023balanced}
       & 0.1254 & 49.47 & 31.427 & 0.8724 & 0.1916 
       & 0.4416 & 56.41 & 29.195 & 0.8746 & 0.1264 
       & 0.3443 & 51.93 & 26.283 & 0.8038 & 0.1951 \\
ODGS~\cite{lee2024odgs}
       & 0.2022 & 37.25 & 31.853 & 0.8873 & 0.1215 
       & 0.3116 & 46.34 & 28.884 & 0.8876 & 0.0885 
       & 0.3702 & 62.33 & 25.582 & 0.8247 & 0.1802 \\
OmniGS~\cite{li2025omnigs}
       & 0.0591 & 61.00 & \textbf{36.046} & 0.9233 & 0.0720 
       & 0.0965 & 58.59 & \textbf{32.683} & \textbf{0.9276} & \textbf{0.0479} 
       & 0.1891 & 46.48 & \textbf{29.241} & \textbf{0.9005} & \textbf{0.0905} \\
SPaGS~\cite{li2025spags}
       & 0.0453 & 73.86 & 35.353 & \textbf{0.9248} & 0.1093 
       & 0.0944 & 67.75 & 32.412 & 0.9270 & 0.0588 
       & 0.1295 & 74.41 & 28.493 & 0.8936 & 0.1171 \\
Ours   
       & \textbf{0.0169} & \textbf{90.56} & 34.684 & 0.9198 & \textbf{0.0709} 
       & \textbf{0.0416} & \textbf{85.28} & 31.403 & 0.9195 & 0.0682 
       & \textbf{0.0620} & \textbf{86.02} & 27.797 & 0.8872 & 0.0978 \\
\bottomrule
\end{tabular}%
}
\caption{Quantitative comparison on public omnidirectional benchmarks. 
We report rendering quality (PSNR/SSIM/LPIPS), Depth reprojection error (DRE), and Cycle inlier ratio (CIR).}
\label{tab:main}
\vskip-3ex
\end{table*}

\begin{figure*}[t]
\centering
\setlength{\tabcolsep}{1.5pt}
\begin{tabular}{ccccc}
\small EgoNeRF~\cite{choi2023balanced} &
\small ODGS~\cite{lee2024odgs} &
\small OmniGS~\cite{li2025omnigs} &
\small SPaGS~\cite{li2025spags} &
\small Ours \\
\includegraphics[width=.19\textwidth]{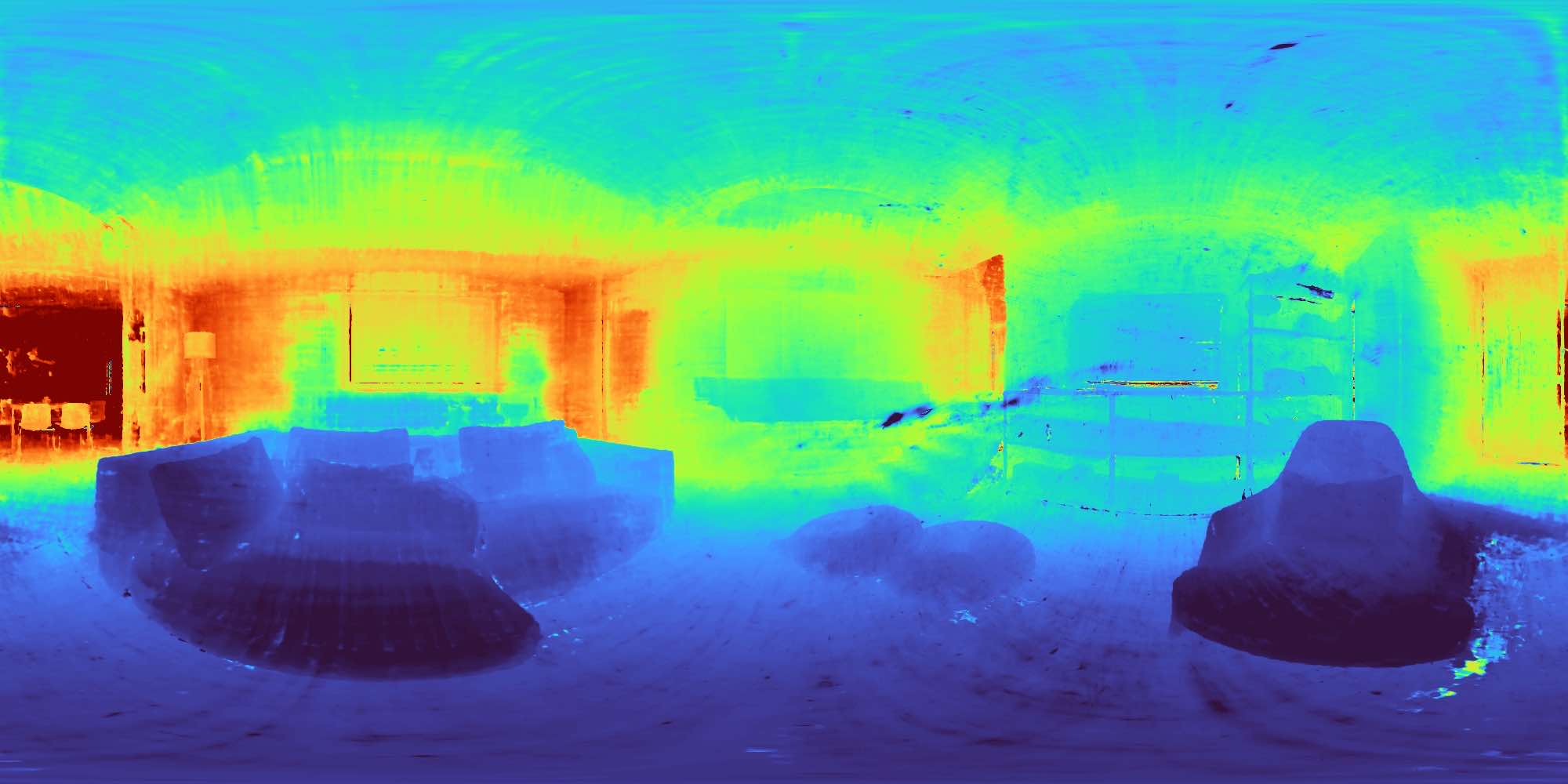} &
\includegraphics[width=.19\textwidth]{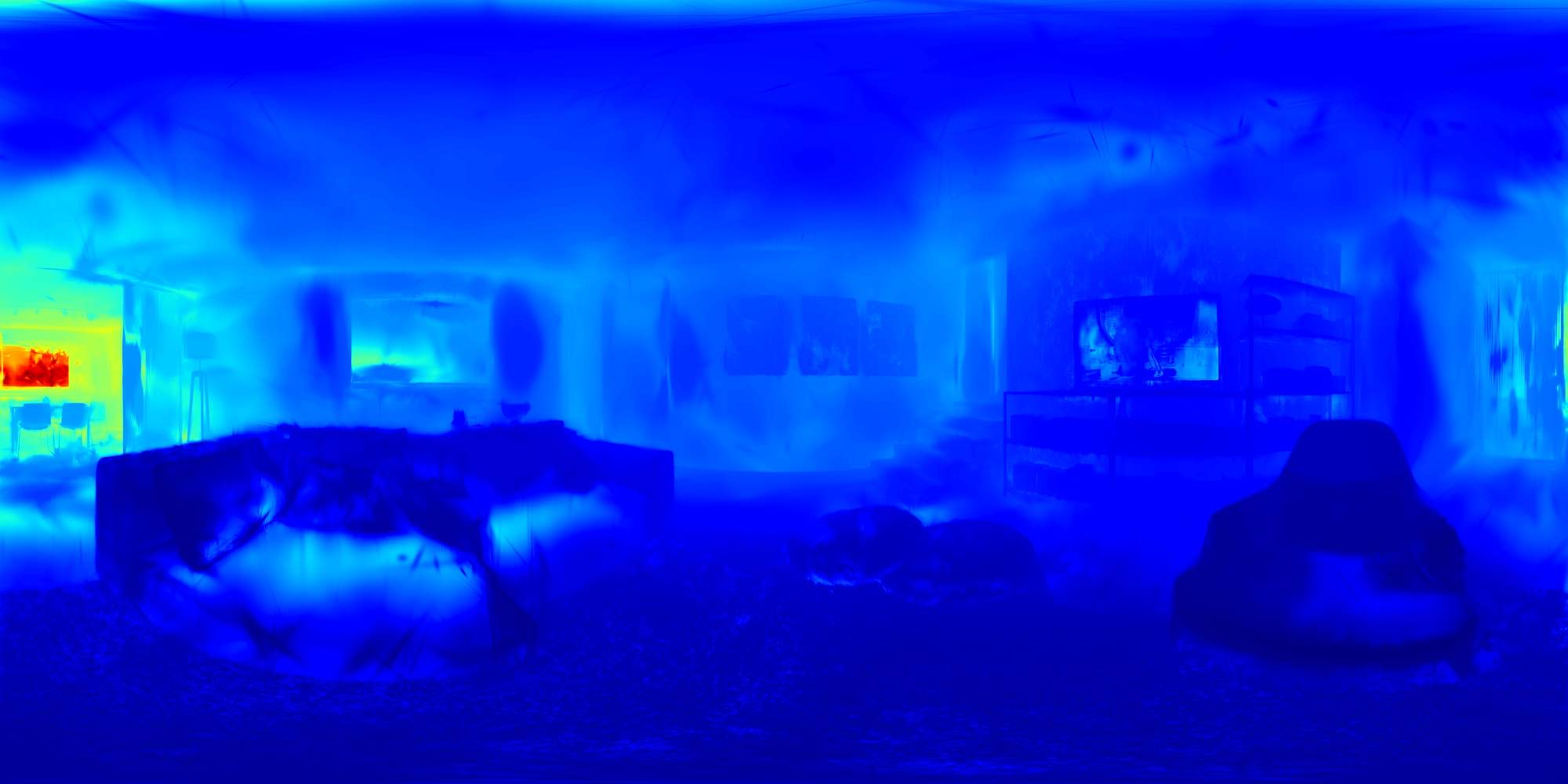} &
\includegraphics[width=.19\textwidth]{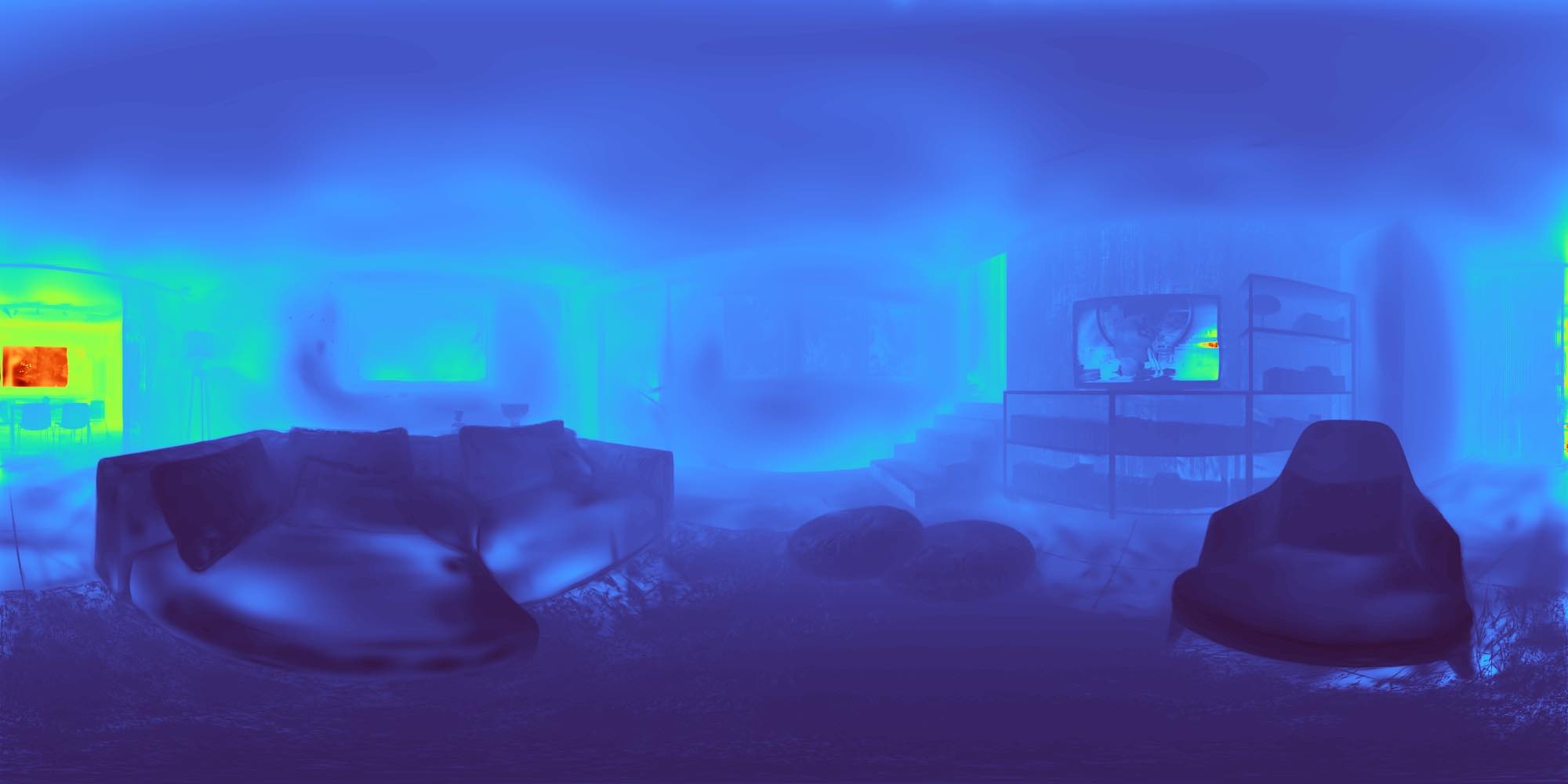} &
\includegraphics[width=.19\textwidth]{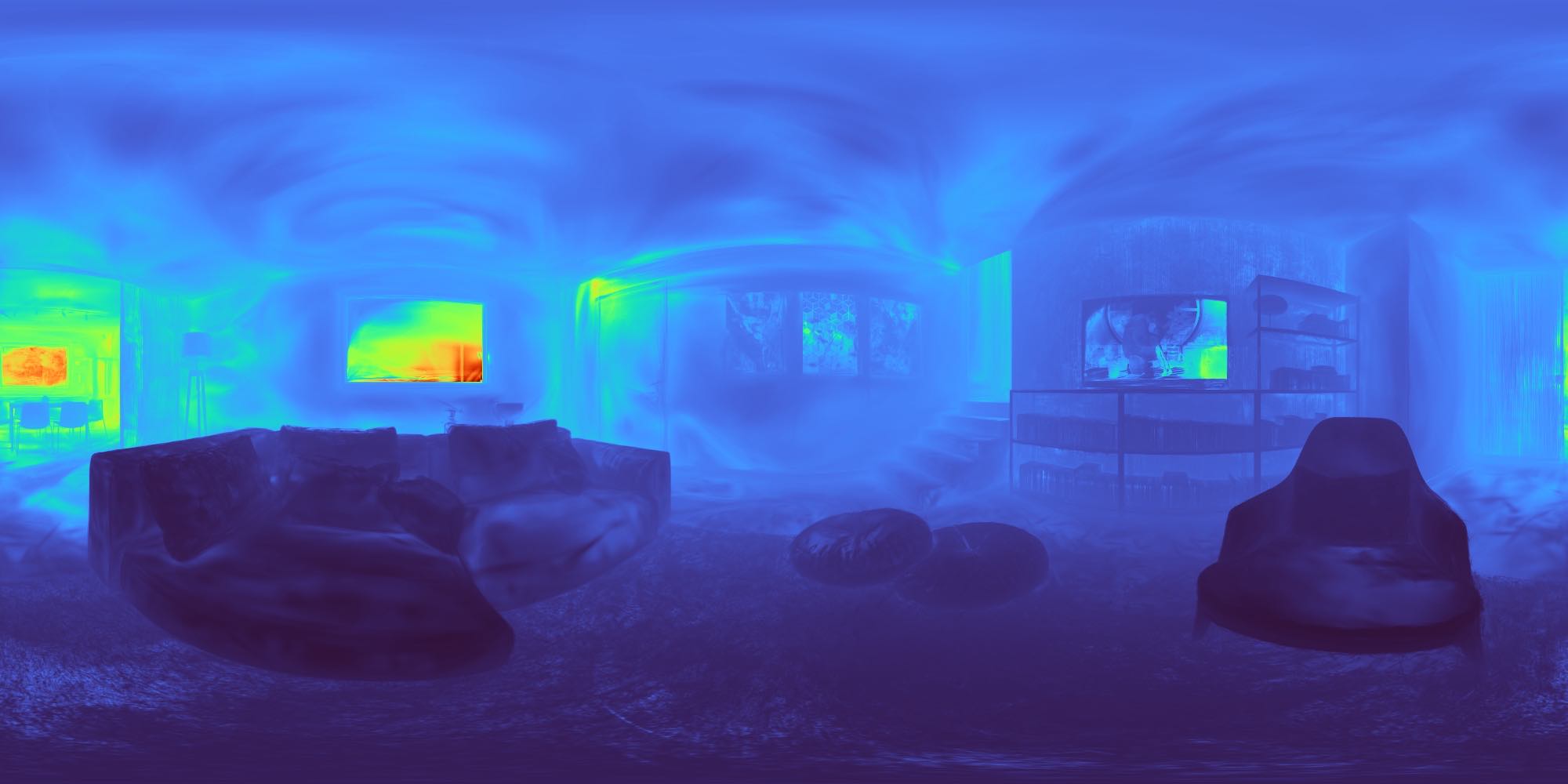} &
\includegraphics[width=.19\textwidth]{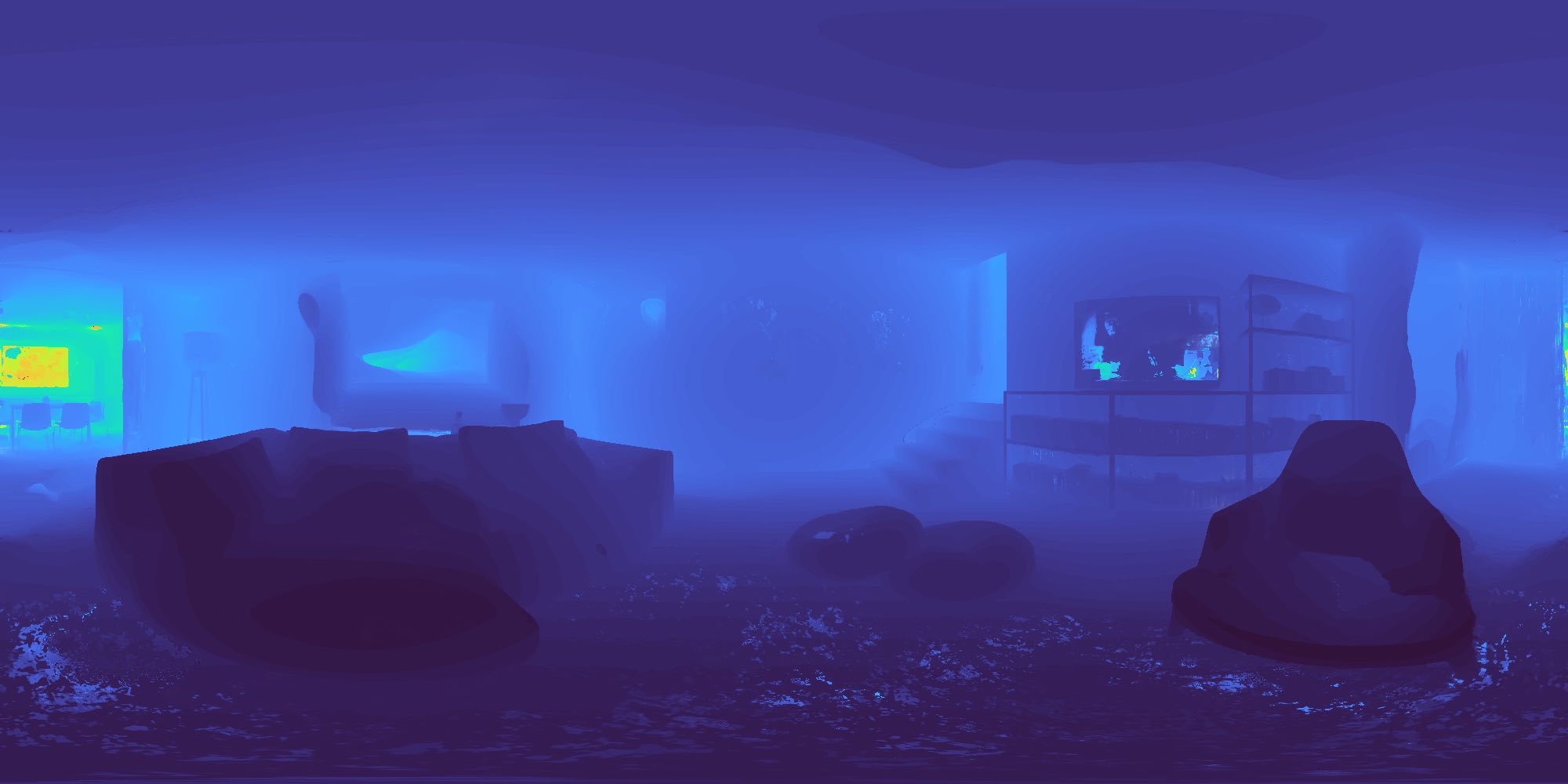} \\
\includegraphics[width=.19\textwidth]{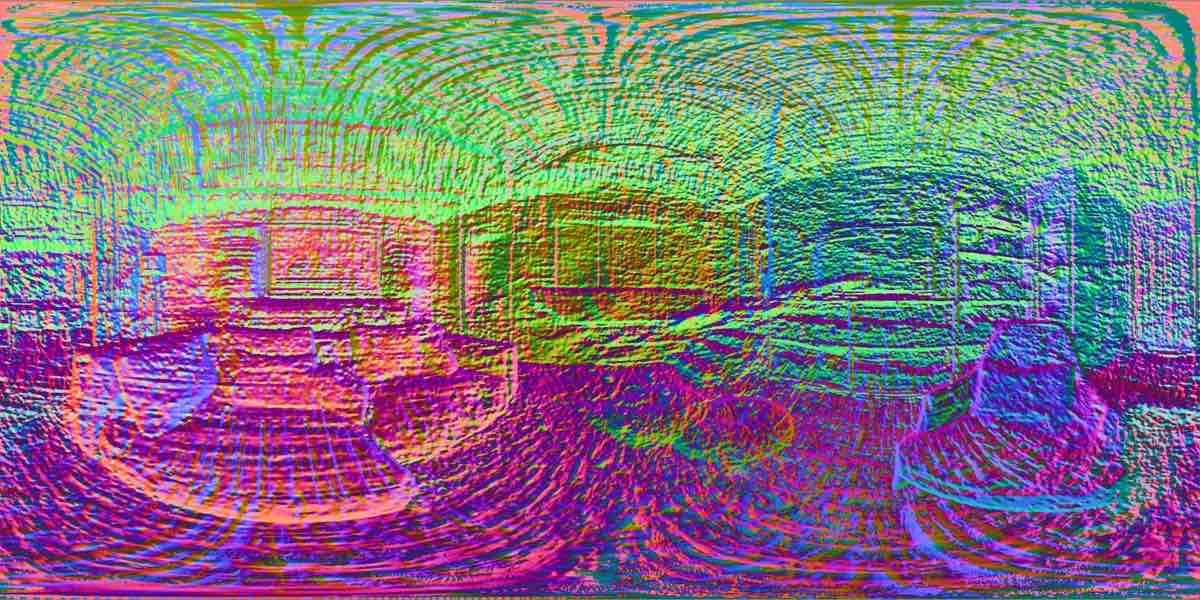} &
\includegraphics[width=.19\textwidth]{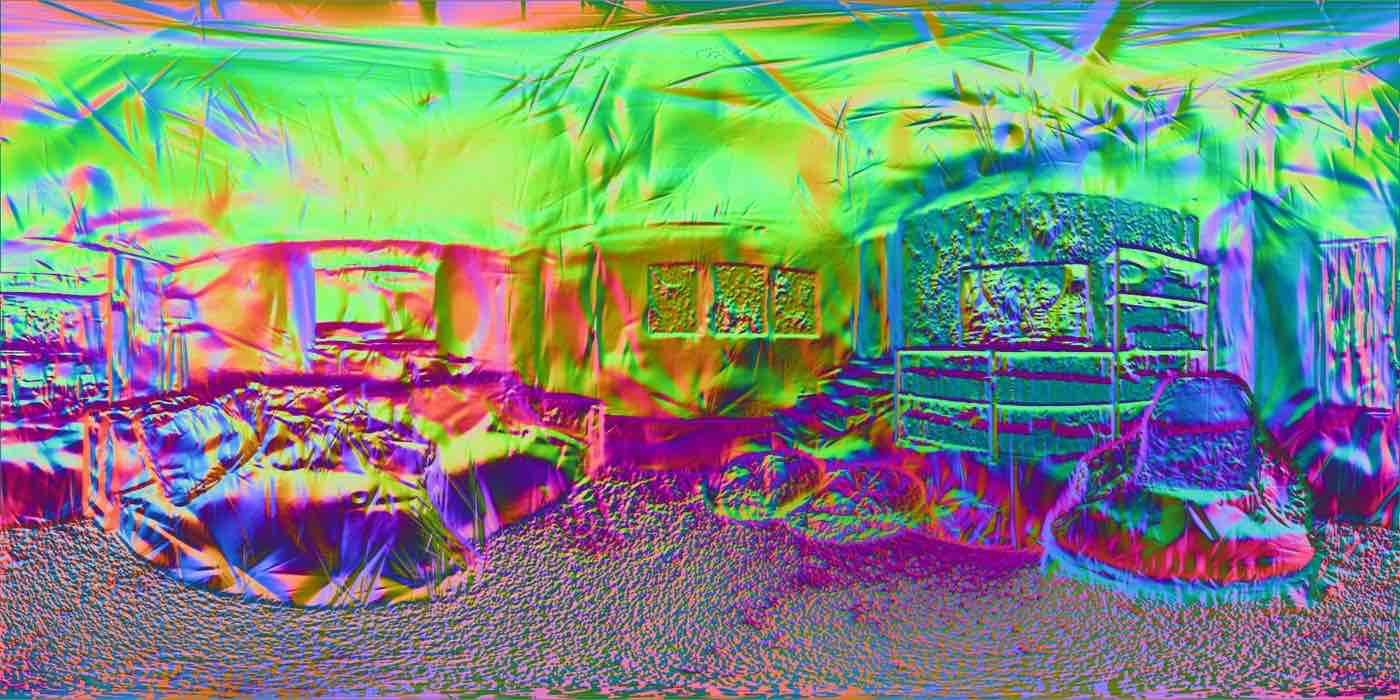} &
\includegraphics[width=.19\textwidth]{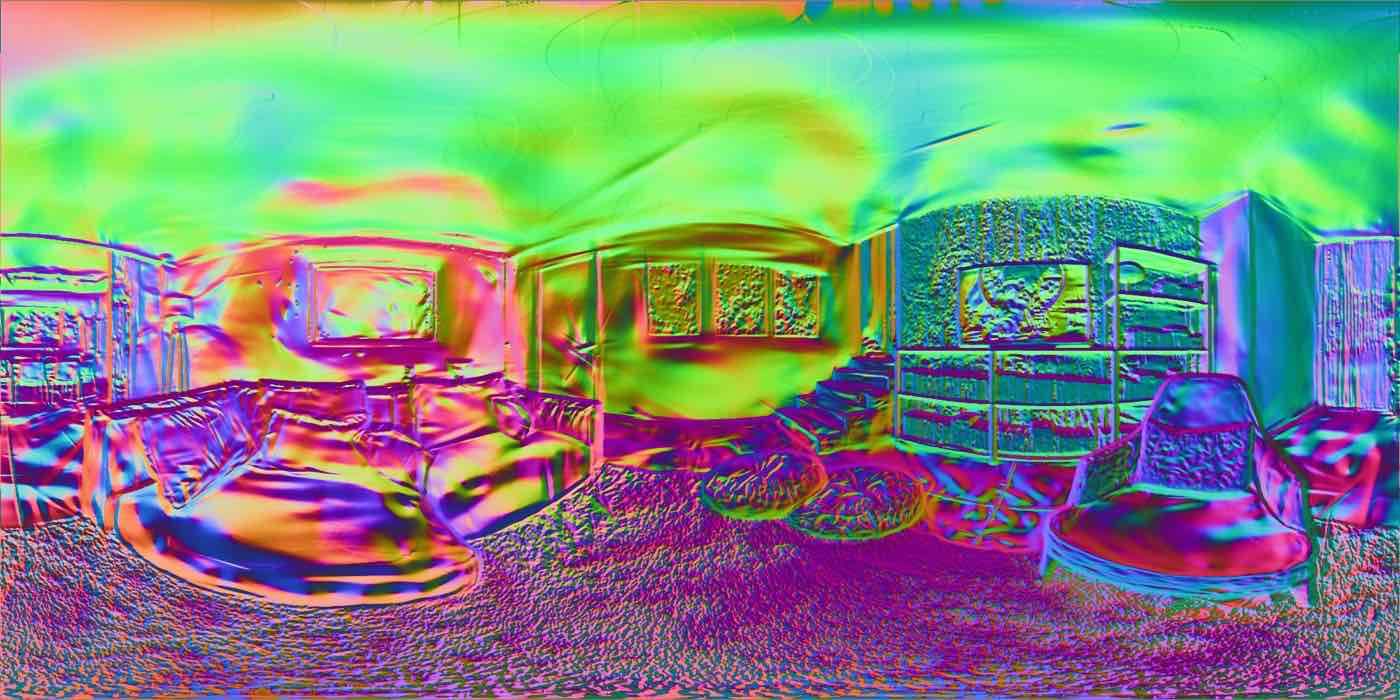} &
\includegraphics[width=.19\textwidth]{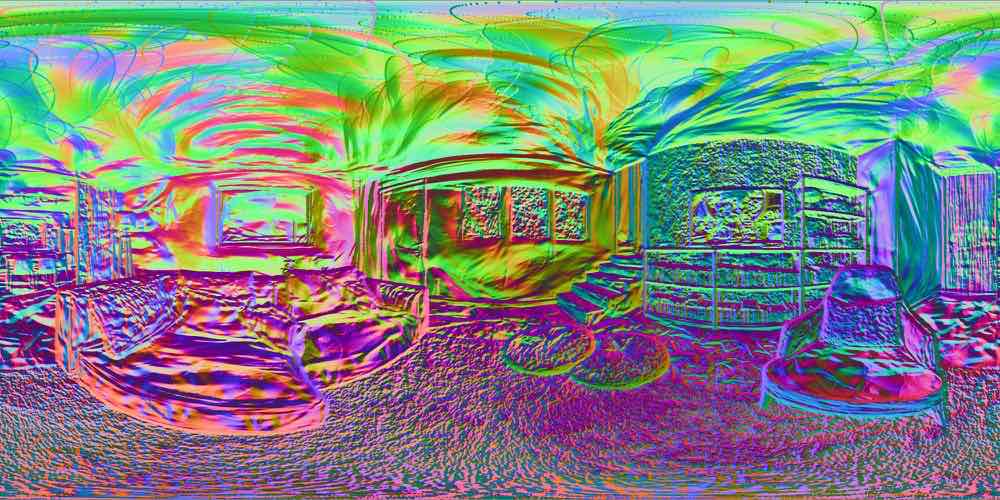} &
\includegraphics[width=.19\textwidth]{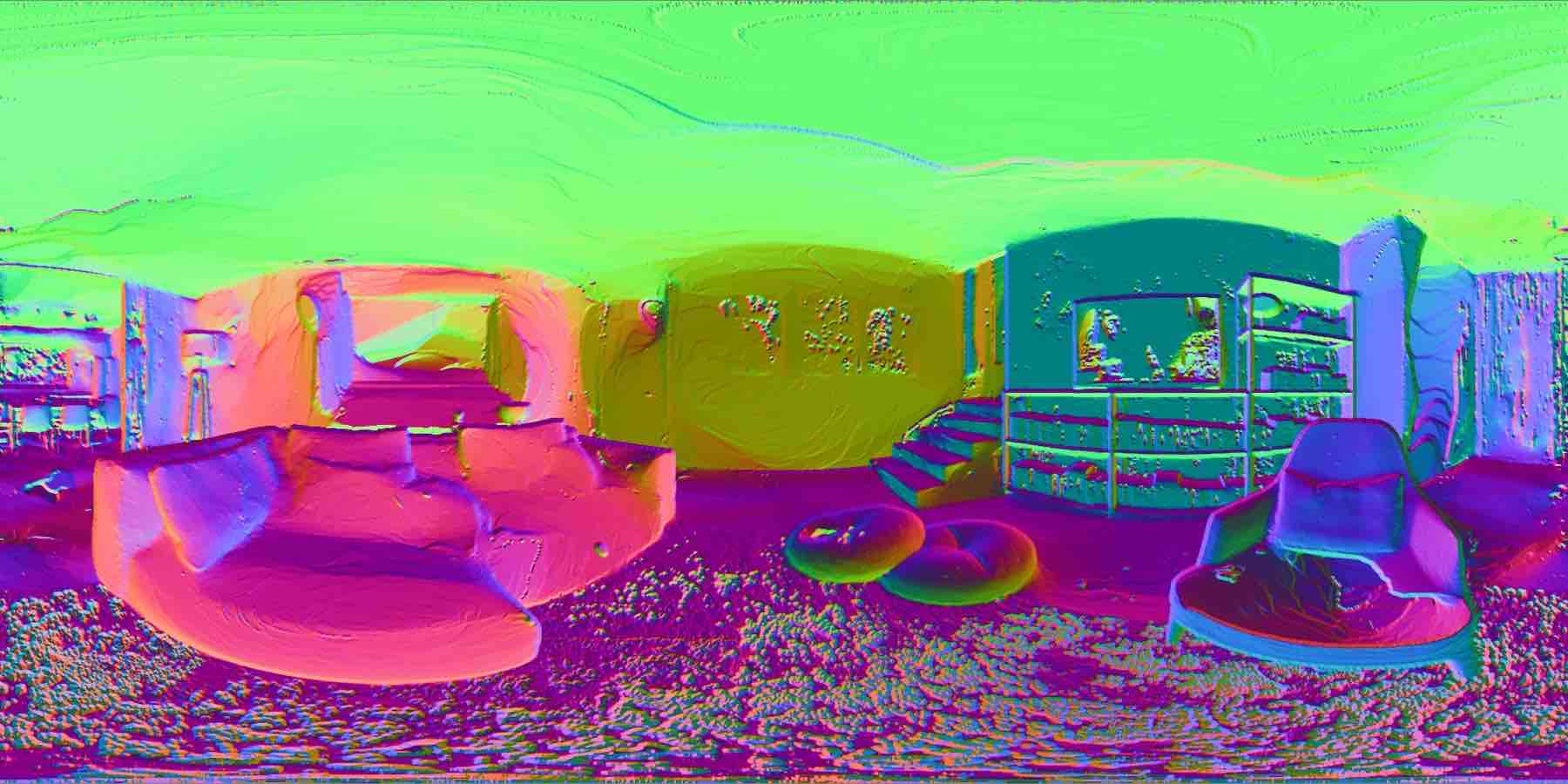} \\
\includegraphics[width=.19\textwidth]{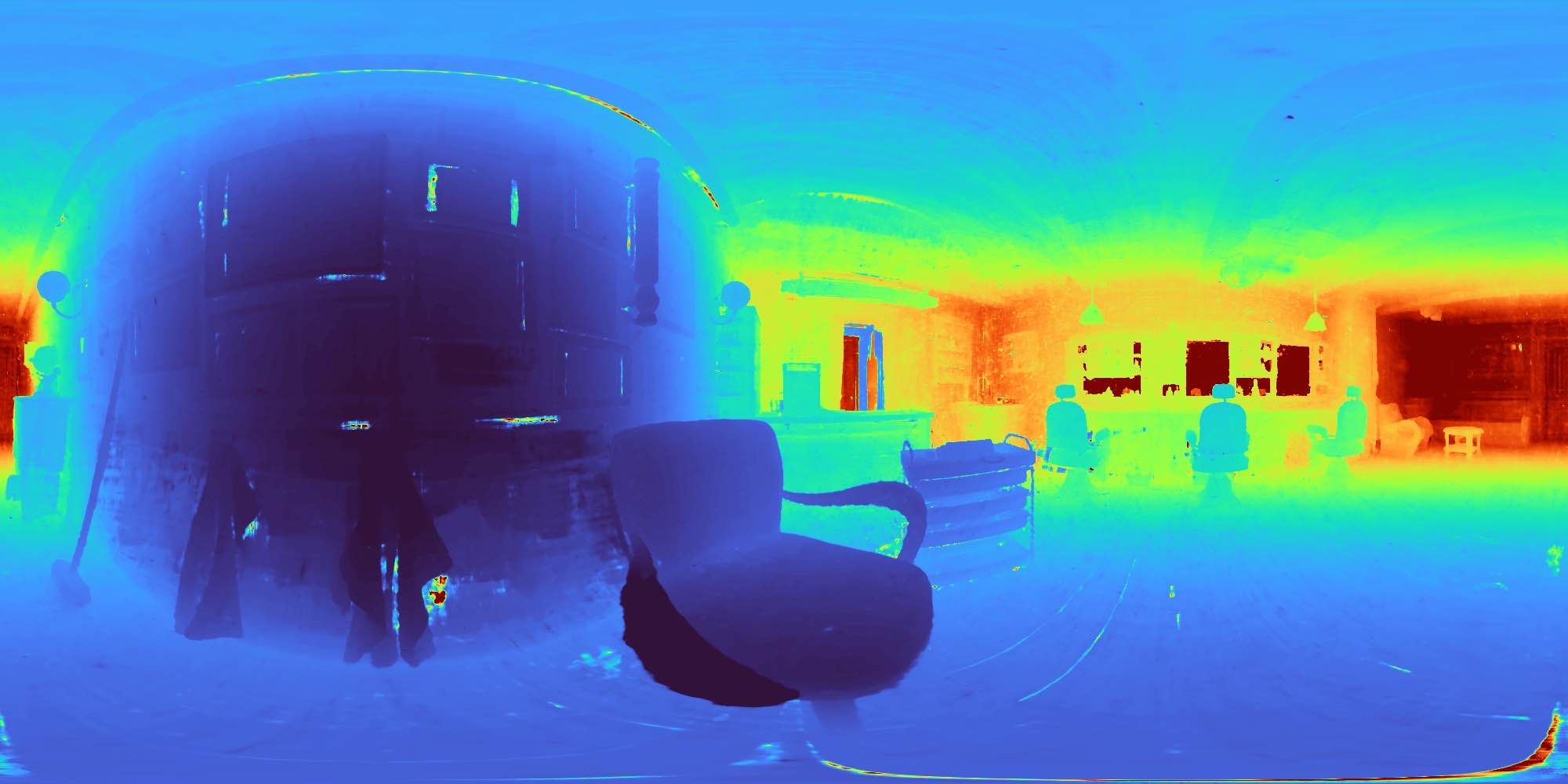} &
\includegraphics[width=.19\textwidth]{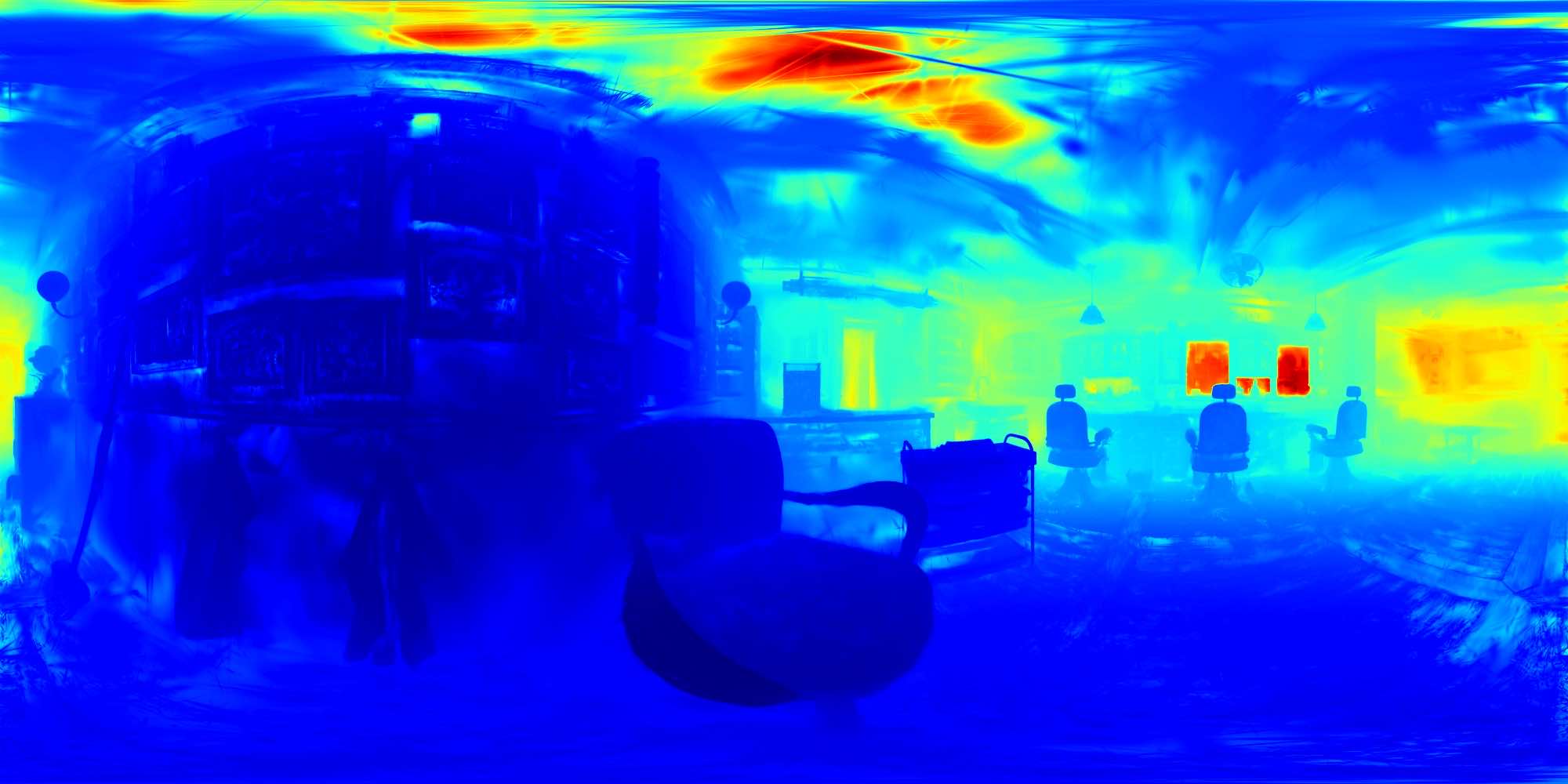} &
\includegraphics[width=.19\textwidth]{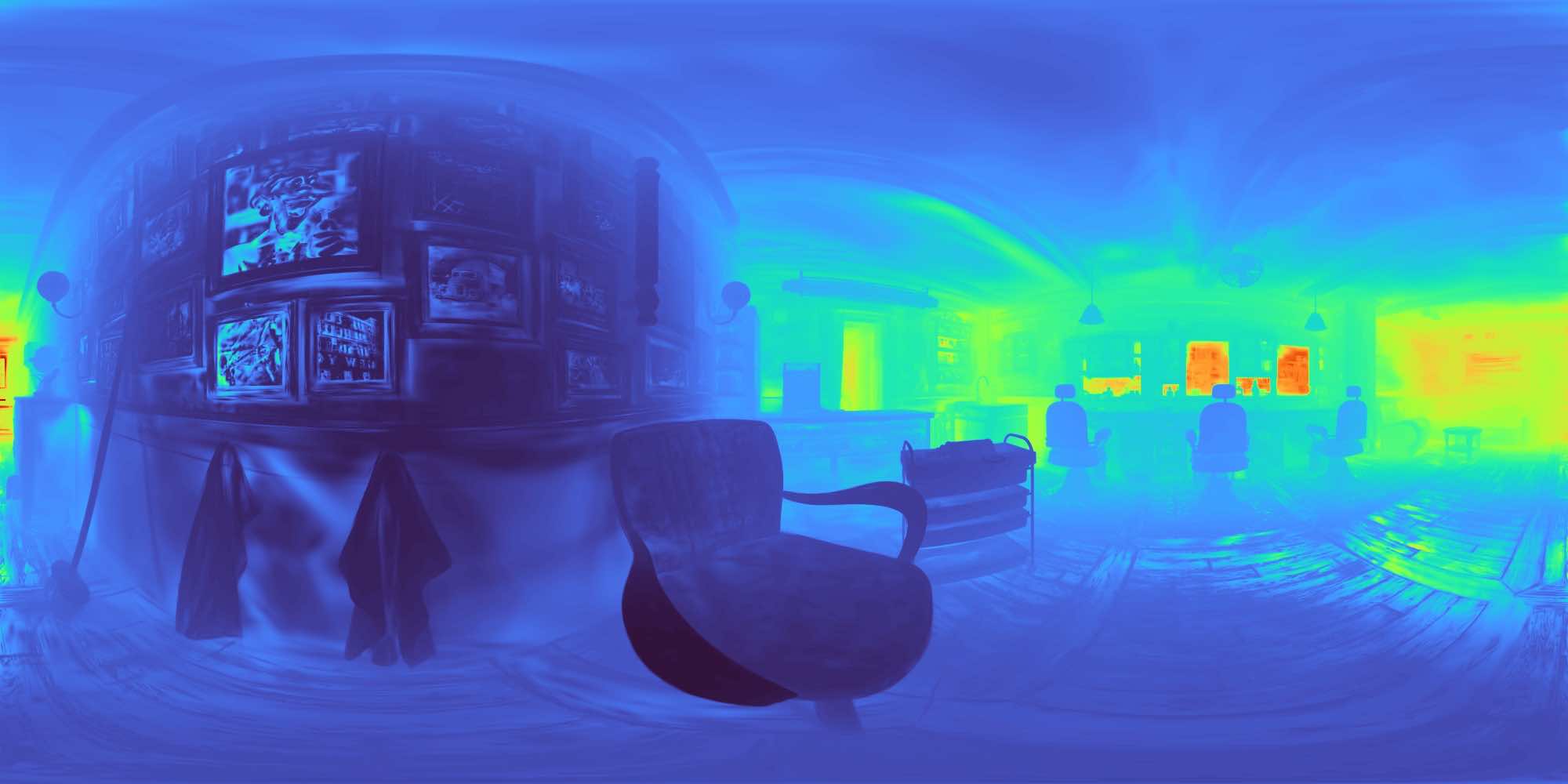} &
\includegraphics[width=.19\textwidth]{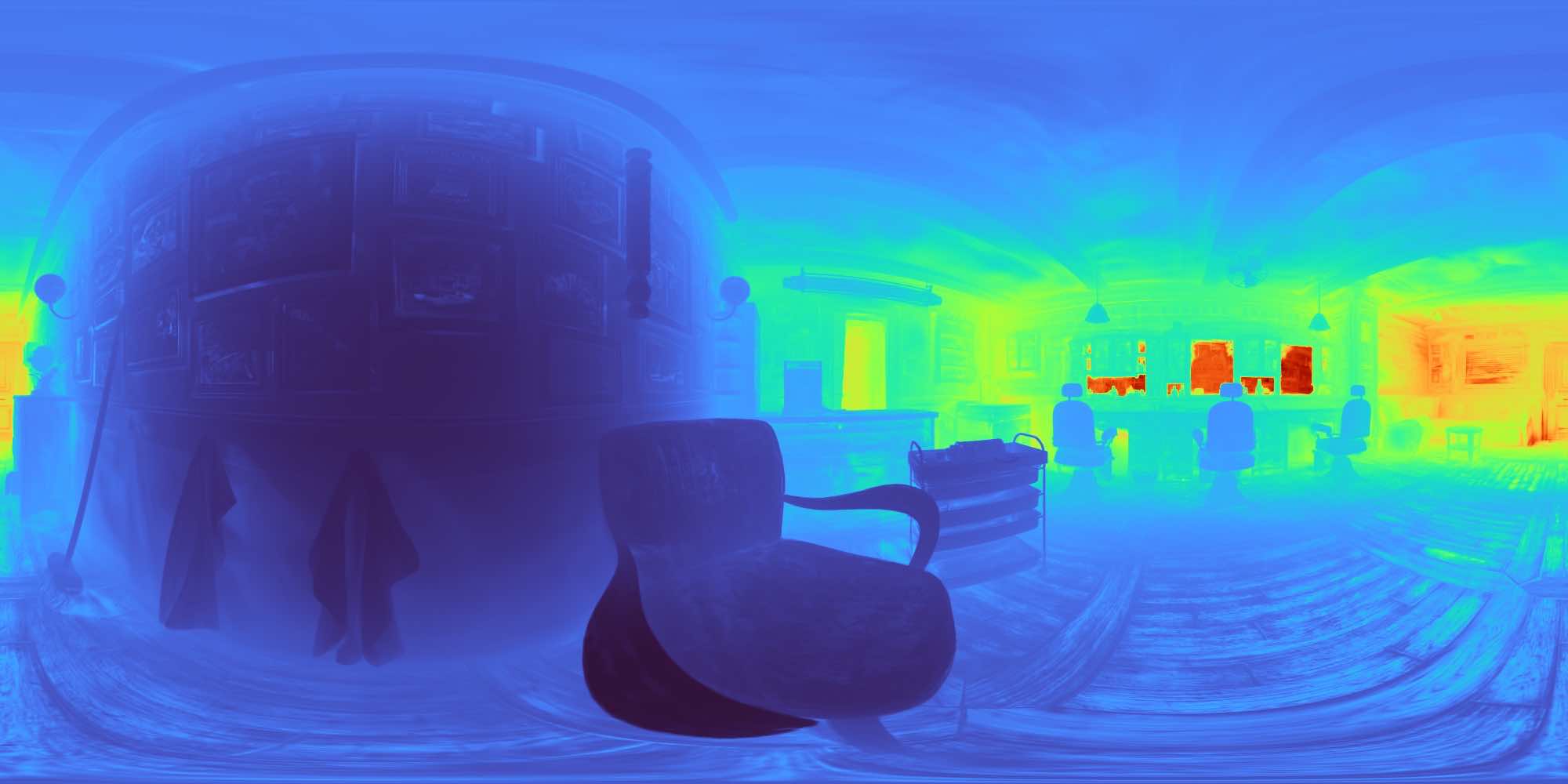} &
\includegraphics[width=.19\textwidth]{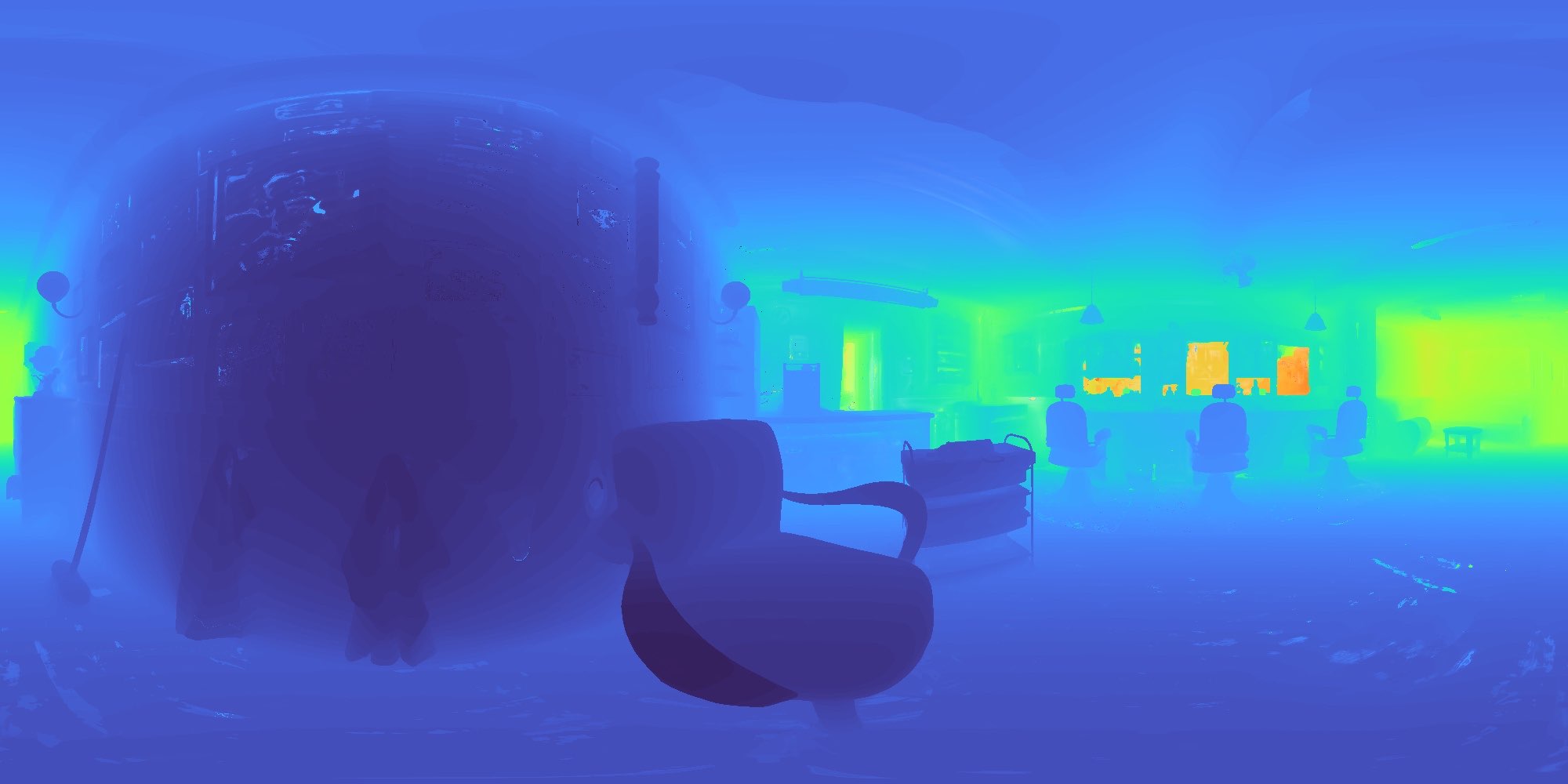} \\
\includegraphics[width=.19\textwidth]{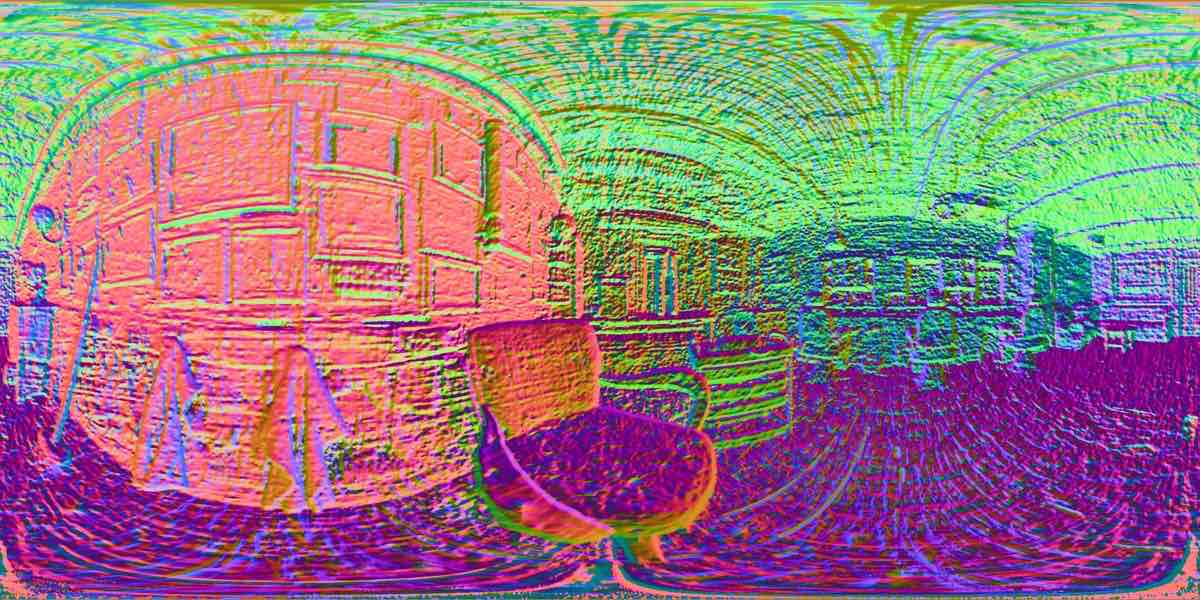} &
\includegraphics[width=.19\textwidth]{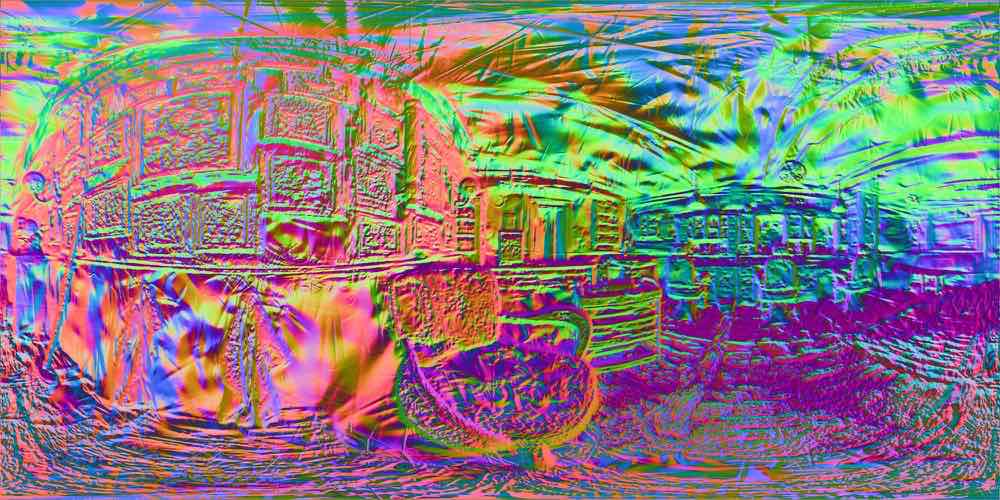} &
\includegraphics[width=.19\textwidth]{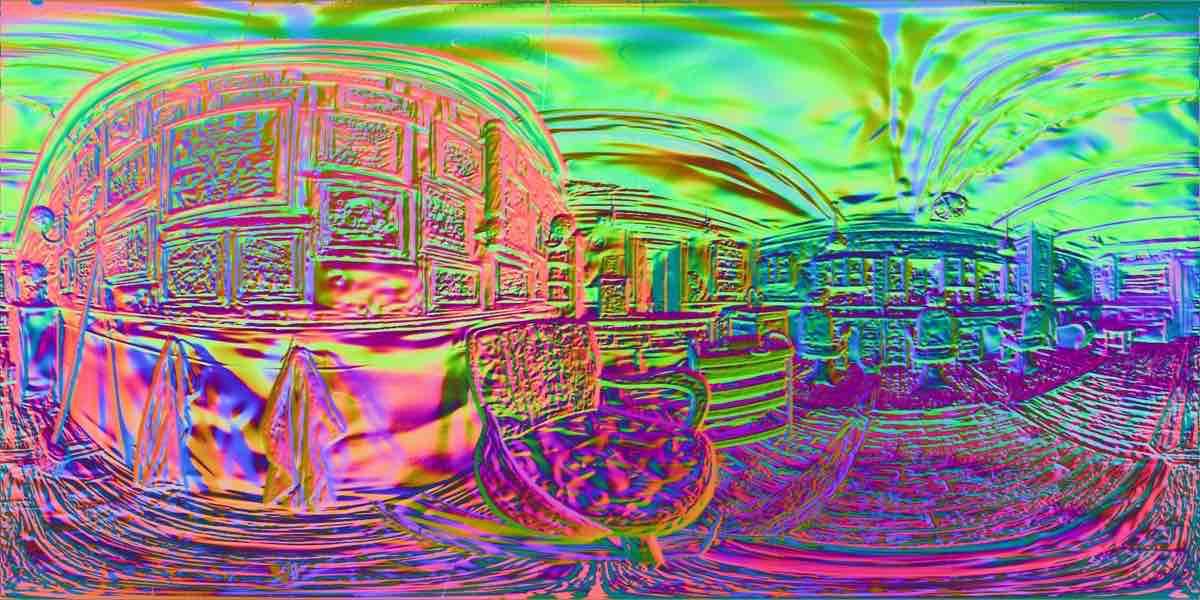} &
\includegraphics[width=.19\textwidth]{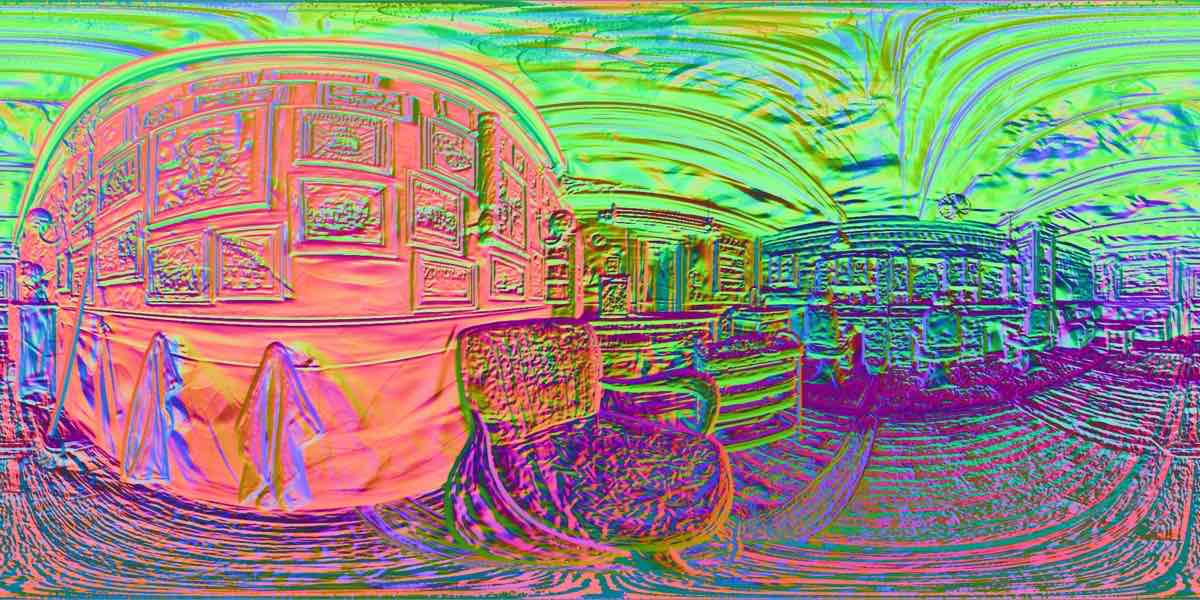} &
\includegraphics[width=.19\textwidth]{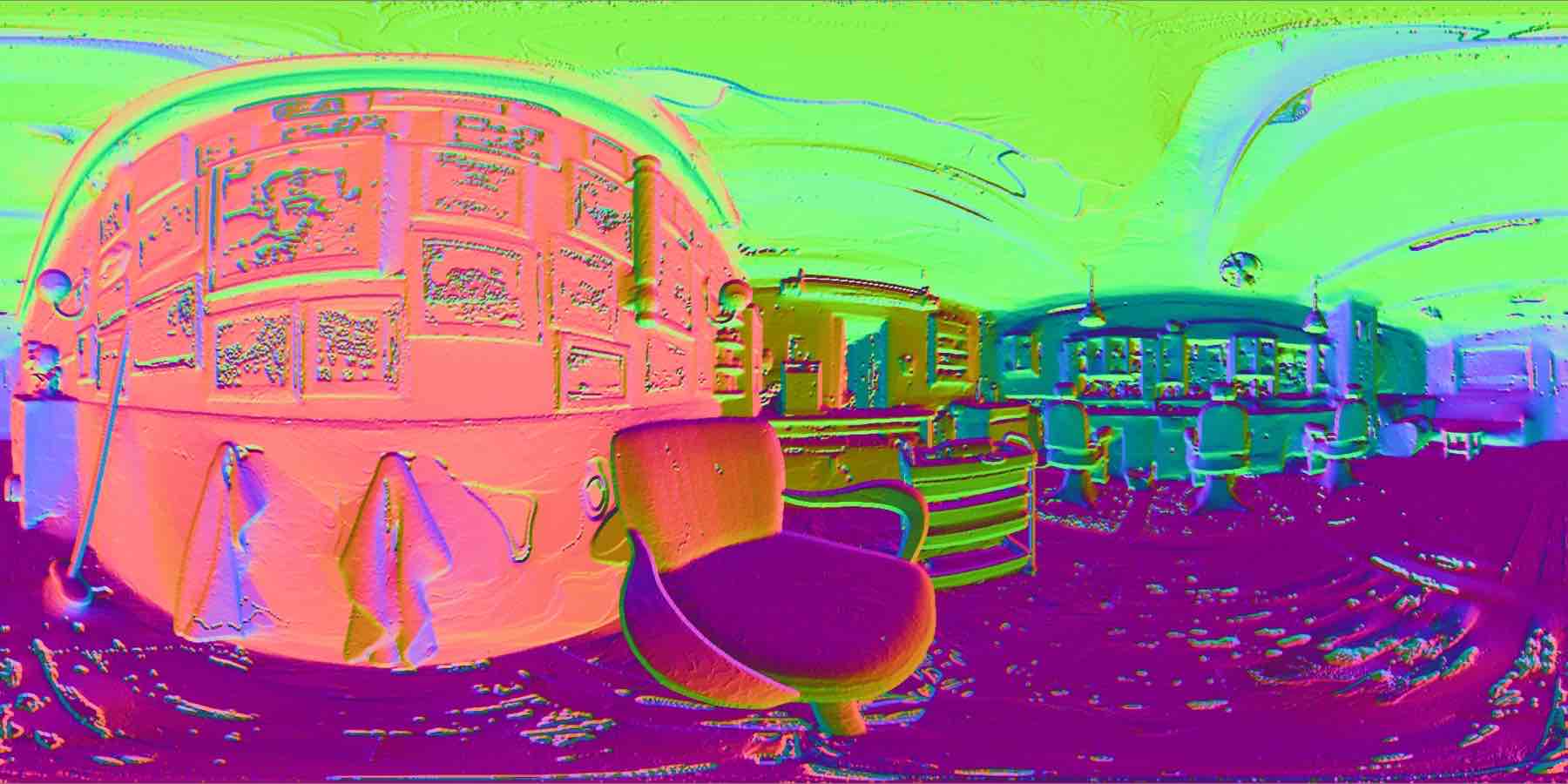} \\
\includegraphics[width=.19\textwidth]{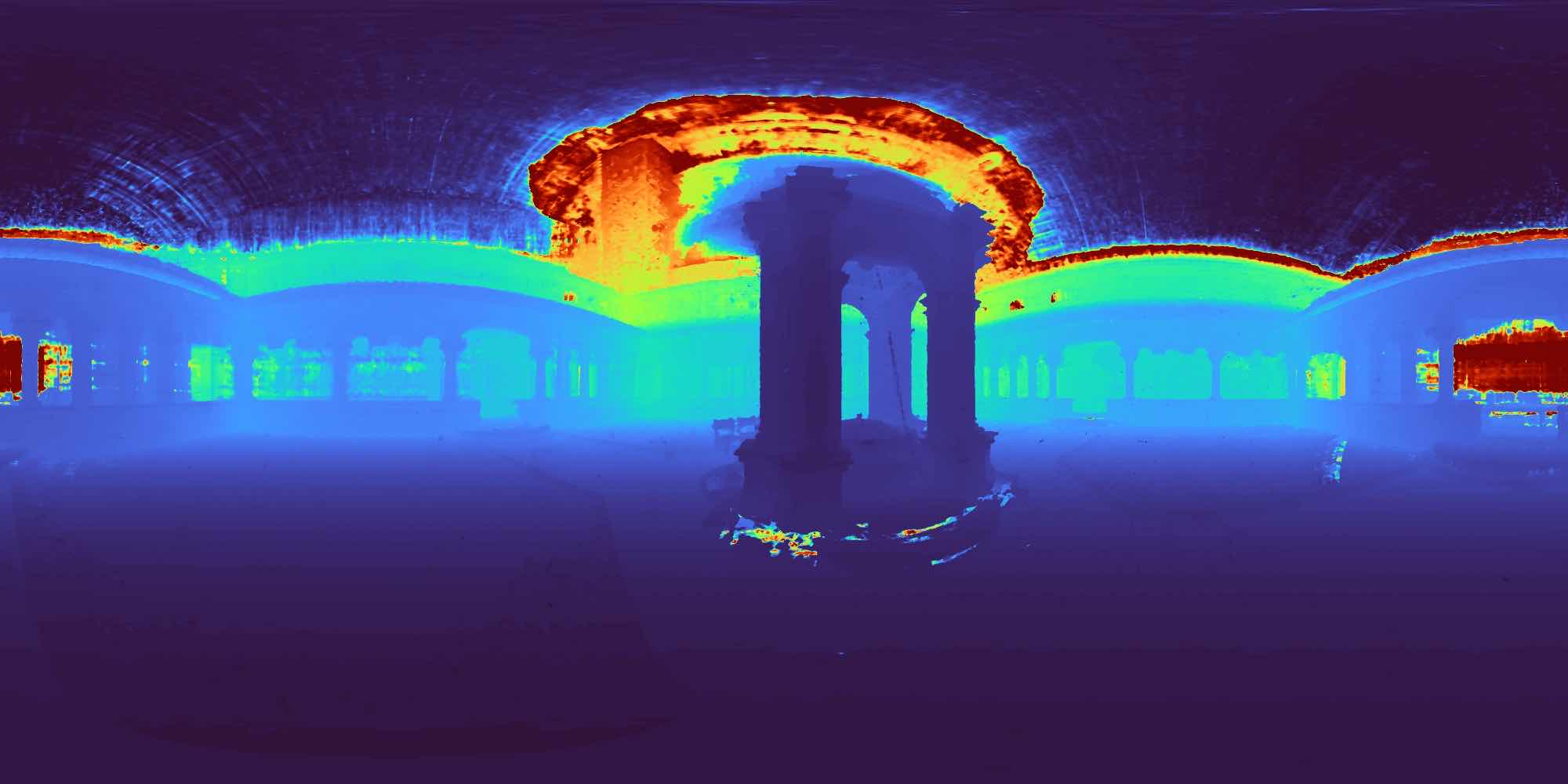} &
\includegraphics[width=.19\textwidth]{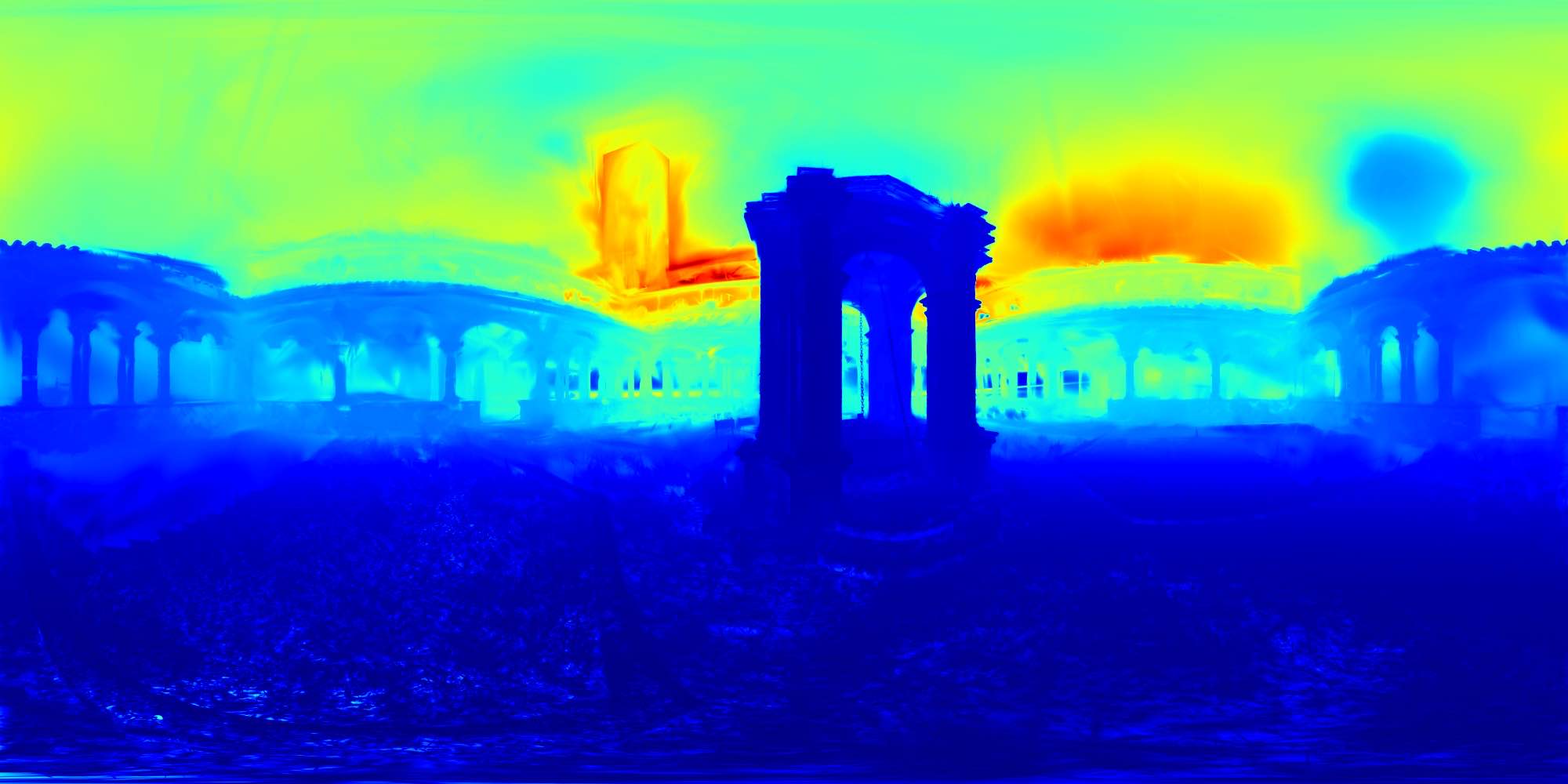} &
\includegraphics[width=.19\textwidth]{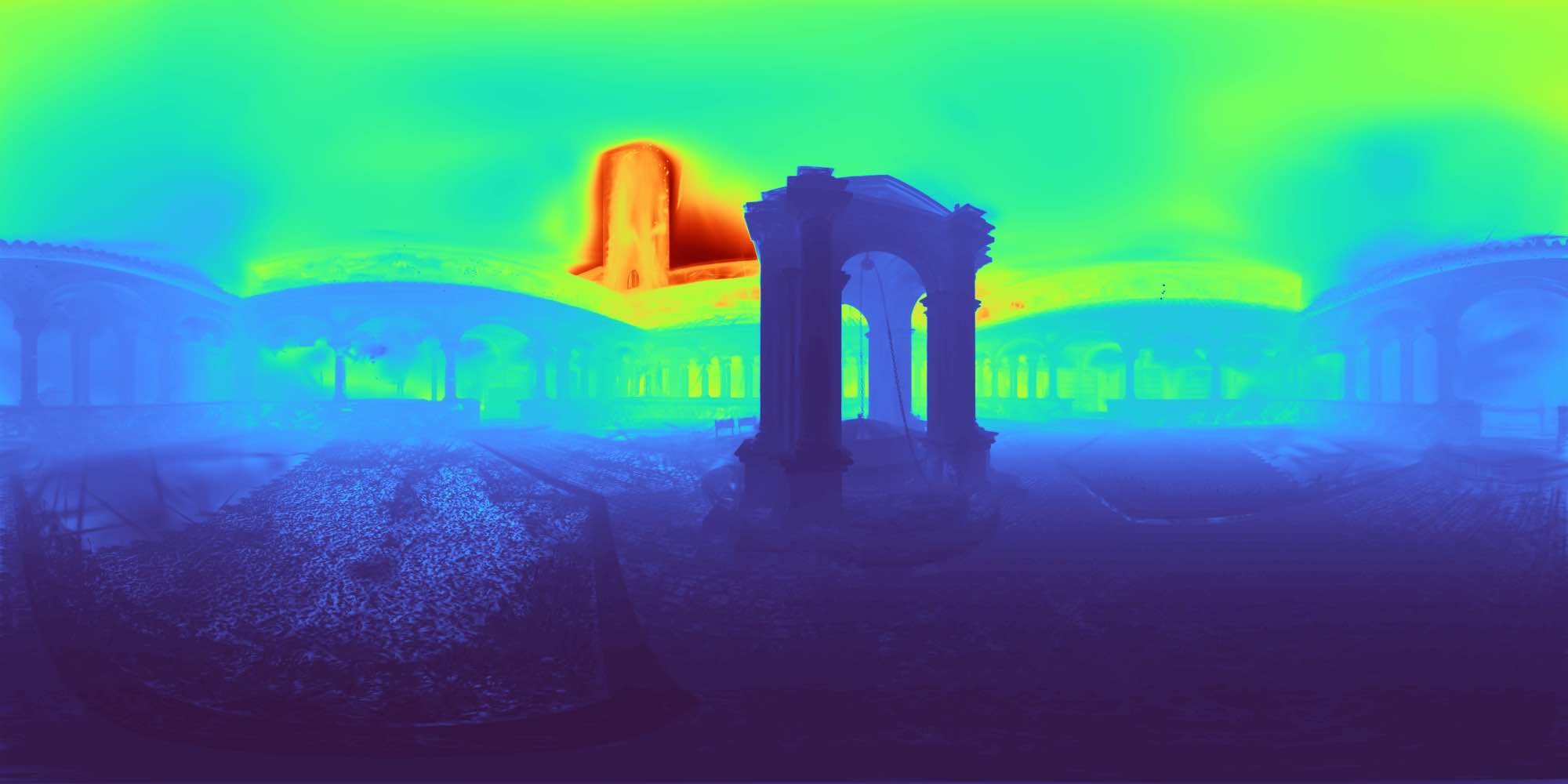} &
\includegraphics[width=.19\textwidth]{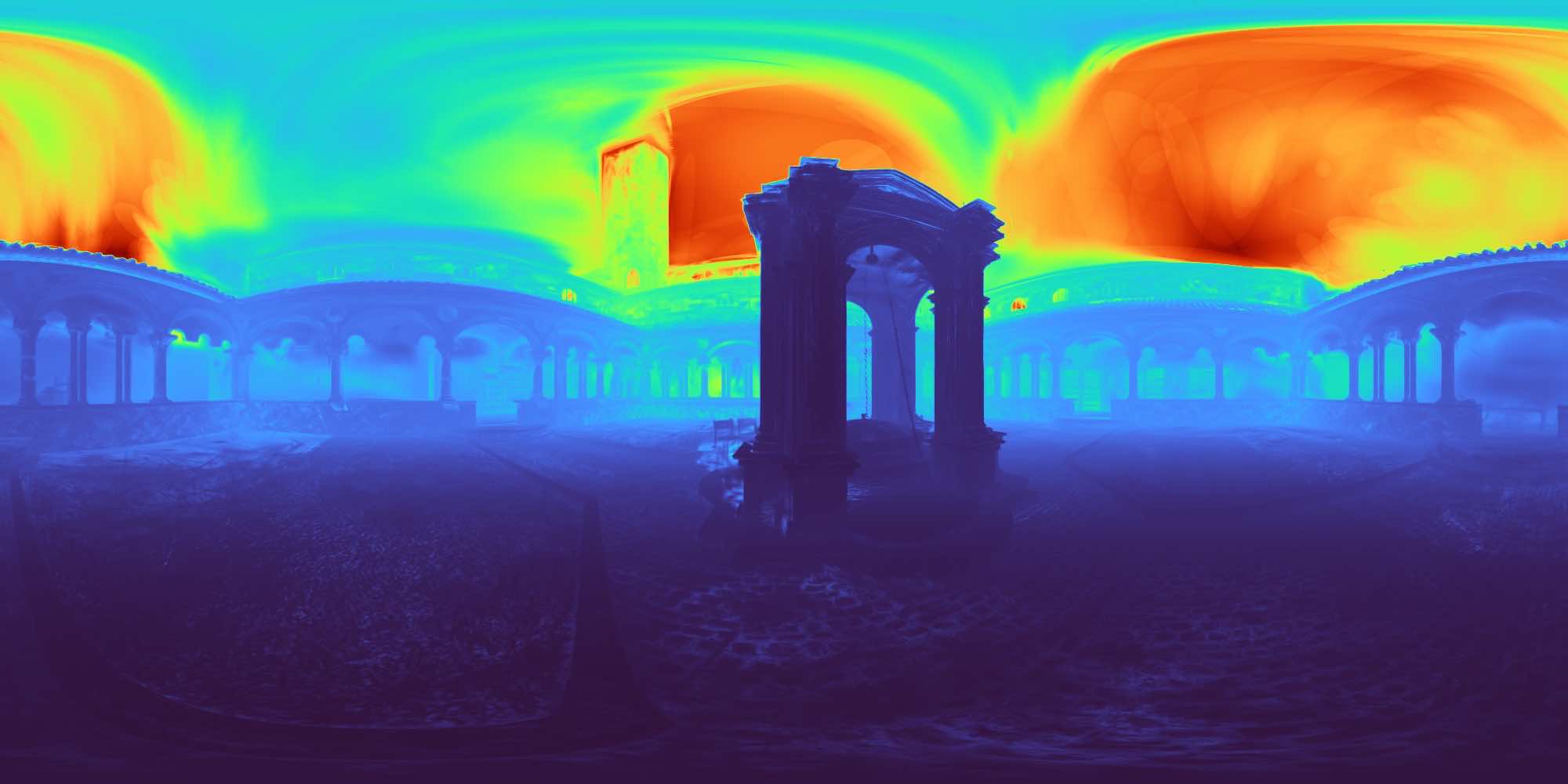} &
\includegraphics[width=.19\textwidth]{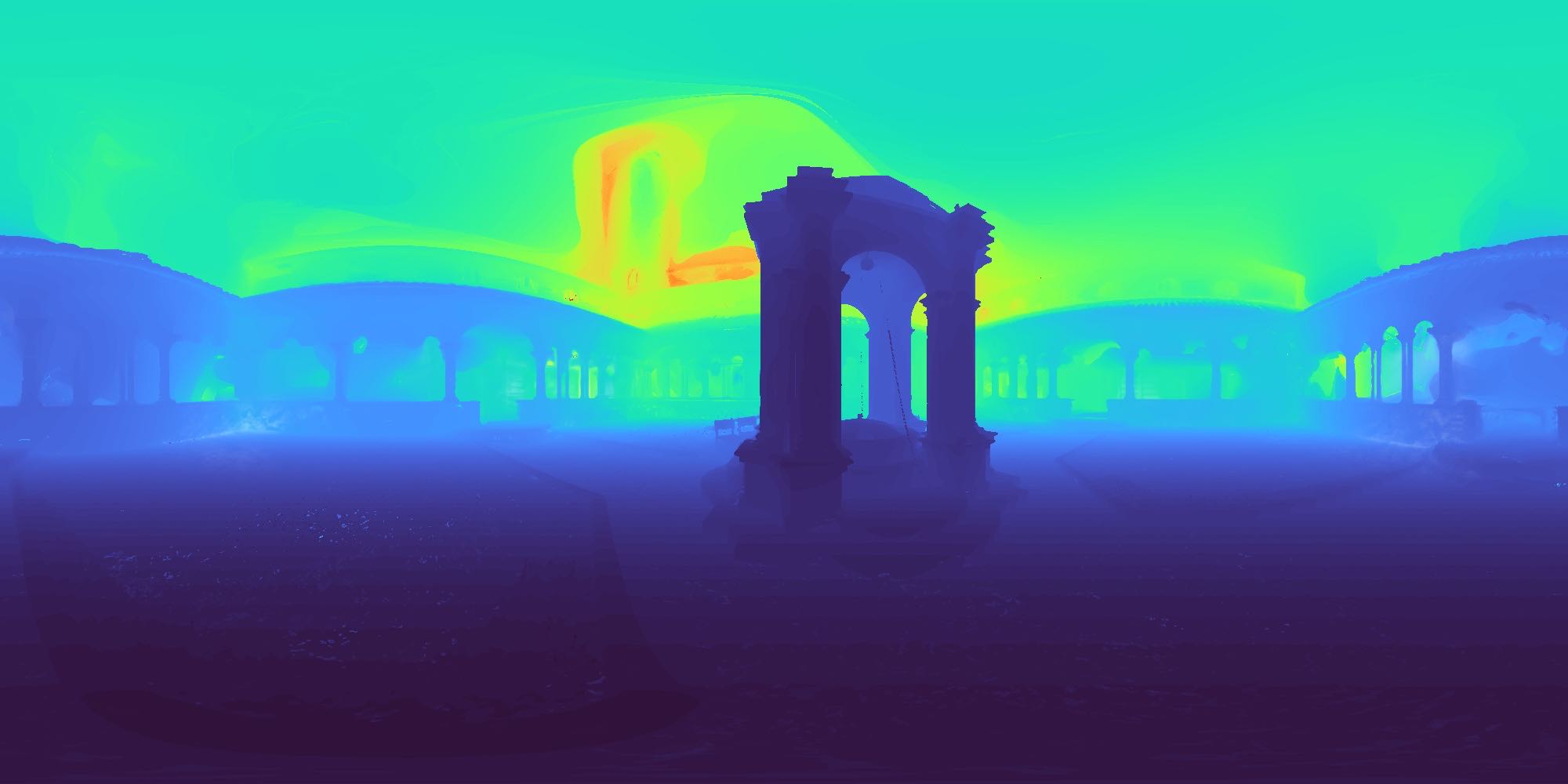} \\
\includegraphics[width=.19\textwidth]{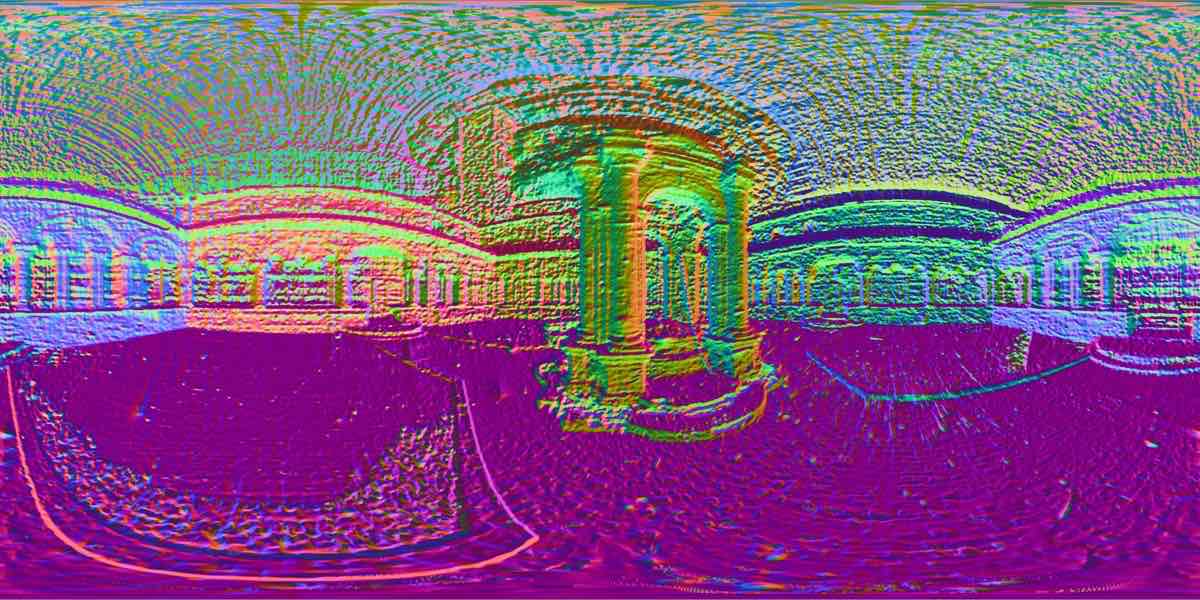} &
\includegraphics[width=.19\textwidth]{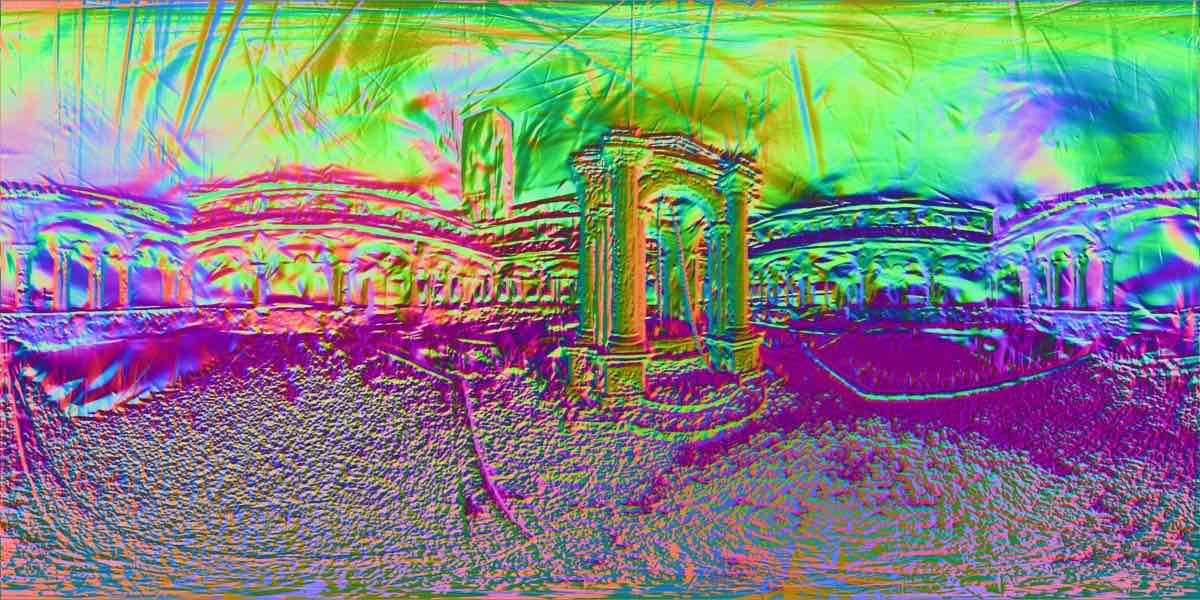} &
\includegraphics[width=.19\textwidth]{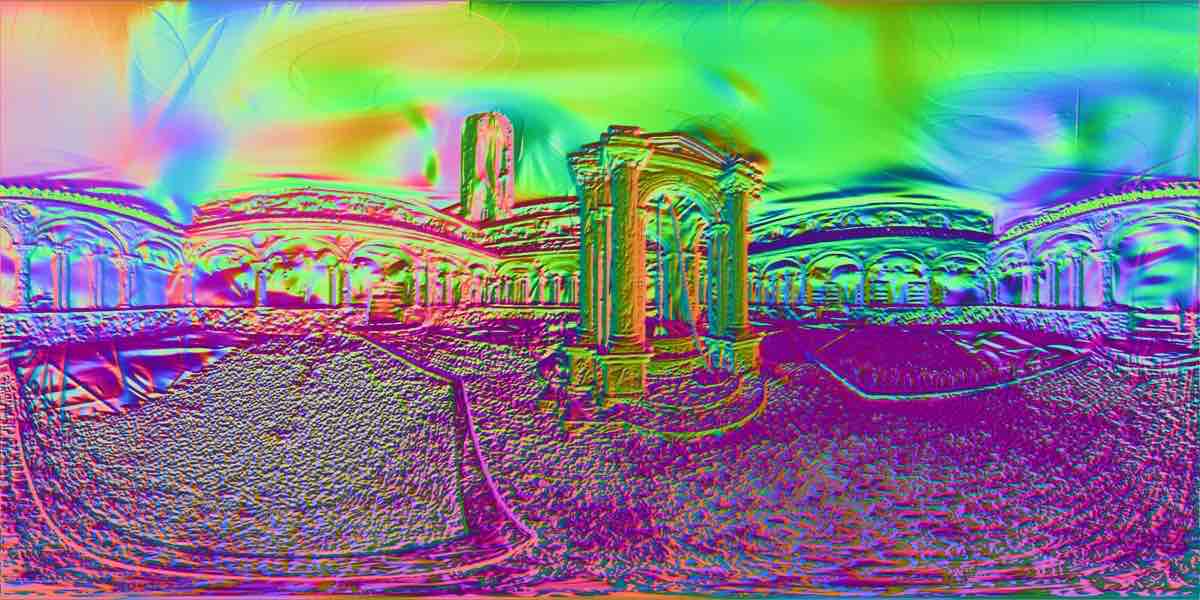} &
\includegraphics[width=.19\textwidth]{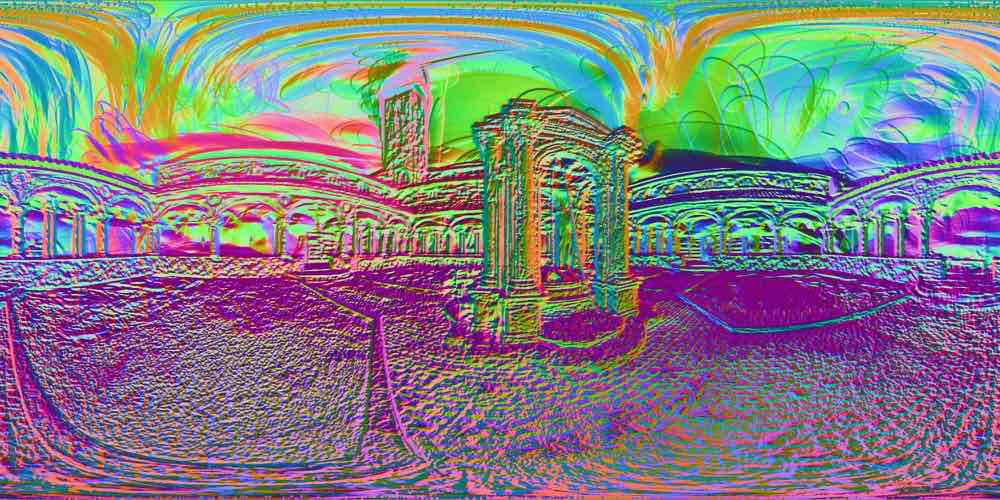} &
\includegraphics[width=.19\textwidth]{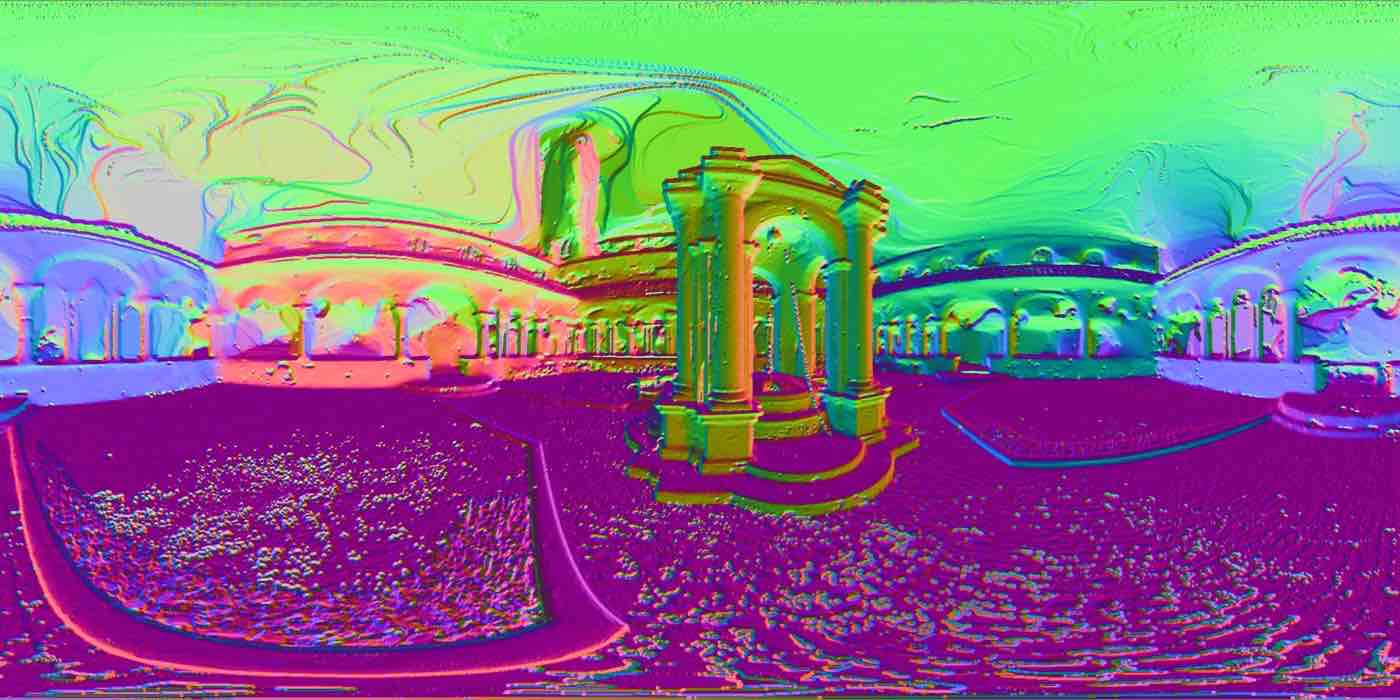} \\
\end{tabular}
\vskip-1ex
\caption{\textbf{Qualitative comparison on OmniBlender}~\cite{choi2023balanced}\textbf{.}
We show RGB renderings and geometry-related outputs for representative scenes.
Compared to projection-based panoramic 3DGS baselines, our method yields smoother and more structurally consistent depth, with fewer texture-aligned ripples on planar regions.
Normal maps are computed from the rendered depth for visualization.}
\label{fig:qual_geom}
\vskip-3ex
\end{figure*}

\subsection{Panorama-Aware Geometric Losses}
Ray-space rendering removes screen-space projection errors, but optimizing geometry from photometric supervision alone remains ill-posed. Without additional constraints, appearance variations can be partially explained by geometry, leading to high-frequency artifacts in depth and normals. We therefore augment the photometric objective with panorama-aware geometric regularizers:
\begin{equation}
\mathcal{L}
=
\mathcal{L}_{\text{rgb}}
+
\lambda_{dn} \mathcal{L}_{\text{dn}}
+
\lambda_{j1} \mathcal{L}_{\text{jump1}}
+
\lambda_{j2} \mathcal{L}_{\text{jump2}} .
\end{equation}
Here, $\mathcal{L}_{\text{rgb}}$ denotes the standard reconstruction loss used in Gaussian rendering, and the additional terms regularize normals and depth to suppress high-frequency geometric artifacts.

\noindent\textbf{Opacity-based valid mask.}
Geometric regularization is only meaningful on pixels with sufficient rendered opacity. We therefore define the valid pixel set
\begin{equation}
\Omega=\bigl\{\mathbf{u}\,\big|\,\alpha(\mathbf{u})>\tau\bigr\},
\label{eq:opacity_valid_set}
\end{equation}
where $\alpha(\mathbf{u})$ denotes the accumulated opacity at pixel $\mathbf{u}$ and $\tau$ is a fixed threshold.

\noindent\textbf{Depth-normal consistency.}
We encourage consistency between the rendered normal map $\mathbf N(\mathbf u)$ and a depth-induced normal map $\mathbf N^{d}(\mathbf u)$ computed from the rendered depth.
Let $\mathbf p(\mathbf u)$ denote the 3D point obtained by back-projecting pixel $\mathbf u$ with its rendered depth.
We estimate
\begin{equation}
\mathbf N^{d}(\mathbf u)=\mathrm{norm}\!\left(\Delta_x \mathbf p(\mathbf u)\times \Delta_y \mathbf p(\mathbf u)\right),
\end{equation}
Here, $\Delta_x$ and $\Delta_y$ denote first-order finite differences of $\mathbf p(\mathbf u)$ along the image $x$- and $y$-directions using neighboring pixels, and we minimize the angular discrepancy between $\mathbf N(\mathbf u)$ and $\mathbf N^{d}(\mathbf u)$.\begin{equation}
\mathcal L_{dn}
=\sum_{\mathbf u\in\Omega}
\omega_{\text{lat}}(\phi)\,
\Bigl(1-\bigl|\mathbf N(\mathbf u)^{\top}\mathbf N^{d}(\mathbf u)\bigr|\Bigr),
\end{equation}
Here, $\Omega$ denotes the opacity-based valid set defined in \eqref{eq:opacity_valid_set}, and we apply the latitude weight $\omega_{\text{lat}}(\phi)$ in \eqref{eq:lat_weight} to compensate for ERP's latitude-dependent distortion and balance the contributions across latitudes.

\noindent\textbf{Depth jump regularization.}
We suppress depth oscillations by applying hinge penalties to log-depth differences. Let $D(\mathbf u)$ denote the rendered depth at panorama pixel $\mathbf u$, and let $\phi(\mathbf u)$ be the corresponding latitude. 
Given a small constant $\epsilon$, we define
$z(\mathbf u)=\log(\max(D(\mathbf u),\epsilon))$ and
$s(\mathbf u)=\max(\cos\phi(\mathbf u),\epsilon)$.
We compute first-order differences
\begin{align}
\Delta_x z(\mathbf u) &=
\frac{z(\mathbf u+\Delta_x)-z(\mathbf u)}{s(\phi)}, \\
\Delta_y z(\mathbf u) &=
z(\mathbf u+\Delta_y)-z(\mathbf u),
\end{align}
and the corresponding hinge responses
\begin{align}
E_x(\mathbf u) &= \max\!\bigl(|\Delta_x z(\mathbf u)|-\tau_1,\,0\bigr), \\
E_y(\mathbf u) &= \max\!\bigl(|\Delta_y z(\mathbf u)|-\tau_1,\,0\bigr).
\end{align}
With edge-aware weights $w_x(\mathbf u)=\exp(-\beta\|\partial_x I(\mathbf u)\|)$ and
$w_y(\mathbf u)=\exp(-\beta\|\partial_y I(\mathbf u)\|)$ computed from the input panorama $I$, we define
\begin{align}
\mathcal L_{\text{jump1}}
&=\sum_{\mathbf u\in\Omega}\omega_{\text{lat}}(\phi)\,
\bigl(
w_x(\mathbf u)E_x(\mathbf u)
+w_y(\mathbf u)E_y(\mathbf u)
\bigr).
\label{eq:jump1}
\end{align}

We further penalize second-order log-depth differences to obtain $\mathcal L_{\text{jump2}}$, using the same horizontal ERP correction to reduce ripple-like artifacts.

\section{Experiments}
\subsection{Experimental Setup}
\noindent\textbf{Datasets.}

We conduct experiments on three datasets: the synthetic OmniBlender~\cite{choi2023balanced}, the real-world OmniPhotos~\cite{bertel2020omniphotos}, and OmniRob.
OmniBlender is rendered in Blender, offering controlled omnidirectional imagery without operator-induced motion artifacts.
OmniPhotos is captured using a consumer 360{\textdegree} camera in real environments.

To evaluate transfer across camera models in robotic settings, we collect OmniRob on two platforms equipped with different omnidirectional cameras.
OmniRob-UAV contains aerial sequences captured by an Antigravity UAV that provides full equirectangular panoramas.
OmniRob-Quadruped contains ground-level sequences captured by a Unitree Go2 legged robot equipped with an annular panoramic camera.
Overall, OmniRob includes four scenes, with two scenes per platform.
This setup enables evaluation under heterogeneous viewpoints and camera parameterizations.

The annular camera used in OmniRob-Quadruped has a limited vertical field of view of $[-39^\circ,6^\circ]$.
To facilitate controlled comparisons under restricted vertical coverage, we crop the UAV panoramas to $[-40^\circ,20^\circ]$ to form pseudo-annular observations.
Following ODGS~\cite{lee2024odgs}, we subsample $100$ images per scene, use $25$ views for training and $25$ for testing, and reserve the remaining views unless otherwise specified.
We initialize the sparse point cloud and camera poses using OpenMVG~\cite{moulon2016openmvg}.

\noindent\textbf{Implementation details.}
Our method is trained for $8k$ iterations on a single NVIDIA RTX 4090 GPU, with densification stopped at $4k$ iterations.
We set $\lambda_{j1}=0.45$, $\lambda_{j2}=0.32$, and $\lambda_{dn}=0.03$ for geometry regularization.
We ramp up geometry regularization during training and enable the depth-normal term only in the later stage.

\begin{table*}[!t]
\centering
\vskip-1ex
\footnotesize
\setlength{\tabcolsep}{3pt}
\renewcommand{\arraystretch}{1.05}
\begin{tabular}{c|ccccc|ccccc|ccccc}
\toprule
& \multicolumn{5}{c|}{$\theta=0^\circ$}
& \multicolumn{5}{c|}{$\theta=60^\circ$}
& \multicolumn{5}{c}{$\theta=90^\circ$} \\
\cmidrule(lr){2-6} \cmidrule(lr){7-11} \cmidrule(lr){12-16} 
Method & 
DRE$\downarrow$ & CIR$\uparrow$ & PSNR$\uparrow$ & SSIM$\uparrow$ & LPIPS$\downarrow$ & 
DRE$\downarrow$ & CIR$\uparrow$ 
& PSNR$\uparrow$ & SSIM$\uparrow$ & LPIPS$\downarrow$ & 
DRE$\downarrow$ & CIR$\uparrow$ 
& PSNR$\uparrow$ & SSIM$\uparrow$ & LPIPS$\downarrow$ \\
\midrule
ODGS~\cite{lee2024odgs} 
& 0.2717 & 43.03 & 29.25 & 0.8862 & 0.1007 & 0.2968 & 42.43 & 25.63 & 0.8622 & 0.1161 & 0.2977 & 38.81 & 24.59 & 0.8440 & 0.1295 \\
OmniGS~\cite{li2025omnigs}       
& 0.0848 & 58.42 & \textbf{33.91} & 0.9260 & 0.0969 & 0.0887 & 58.05 & 24.28 & 0.8386 & 0.1461 & 0.0857 & 60.04 & 23.01 & 0.8193 & 0.1689 \\
SPaGS~\cite{li2025spags}
& 0.0954 & 67.76 & 33.48 & \textbf{0.9262} & 0.0771 & 0.0945 & 67.92 & \textbf{34.89} & \textbf{0.9444} & 0.0589 & 0.0955 & 67.75 & \textbf{34.50} & \textbf{0.9407} & 0.0644 \\
Ours 
& \textbf{0.0326} & \textbf{87.20} & 32.61 & 0.9203 & \textbf{0.0697} & \textbf{0.0323} & \textbf{89.21} & 31.12 & 0.9188 & \textbf{0.0525} & \textbf{0.0322} & \textbf{90.36} & 30.27 & 0.9065 & \textbf{0.0585} \\
\bottomrule
\end{tabular}
\caption{Rotation robustness on OmniBlender~\cite{choi2023balanced}. 
Models are trained on canonical poses and evaluated under additional random rotations within $\pm\theta$.}
\label{tab:rot_full}
\end{table*}

\begin{figure*}[t]
\centering
\setlength{\tabcolsep}{1.5pt}
\renewcommand{\arraystretch}{1.0}

\begin{tabular}{@{}>{\raggedleft\arraybackslash}m{9pt}@{\hspace{4pt}}ccccc@{}}
& \small ODGS~\cite{lee2024odgs} &
  \small OmniGS~\cite{li2025omnigs} &
  \small SPaGS~\cite{li2025spags} &
  \small Ours &
  \small Ground truth \\

\adjustbox{valign=c}{\rotatebox{90}{\scriptsize\textit{Yaw}~$0^\circ$}} &
\includegraphics[width=.19\textwidth,valign=c]{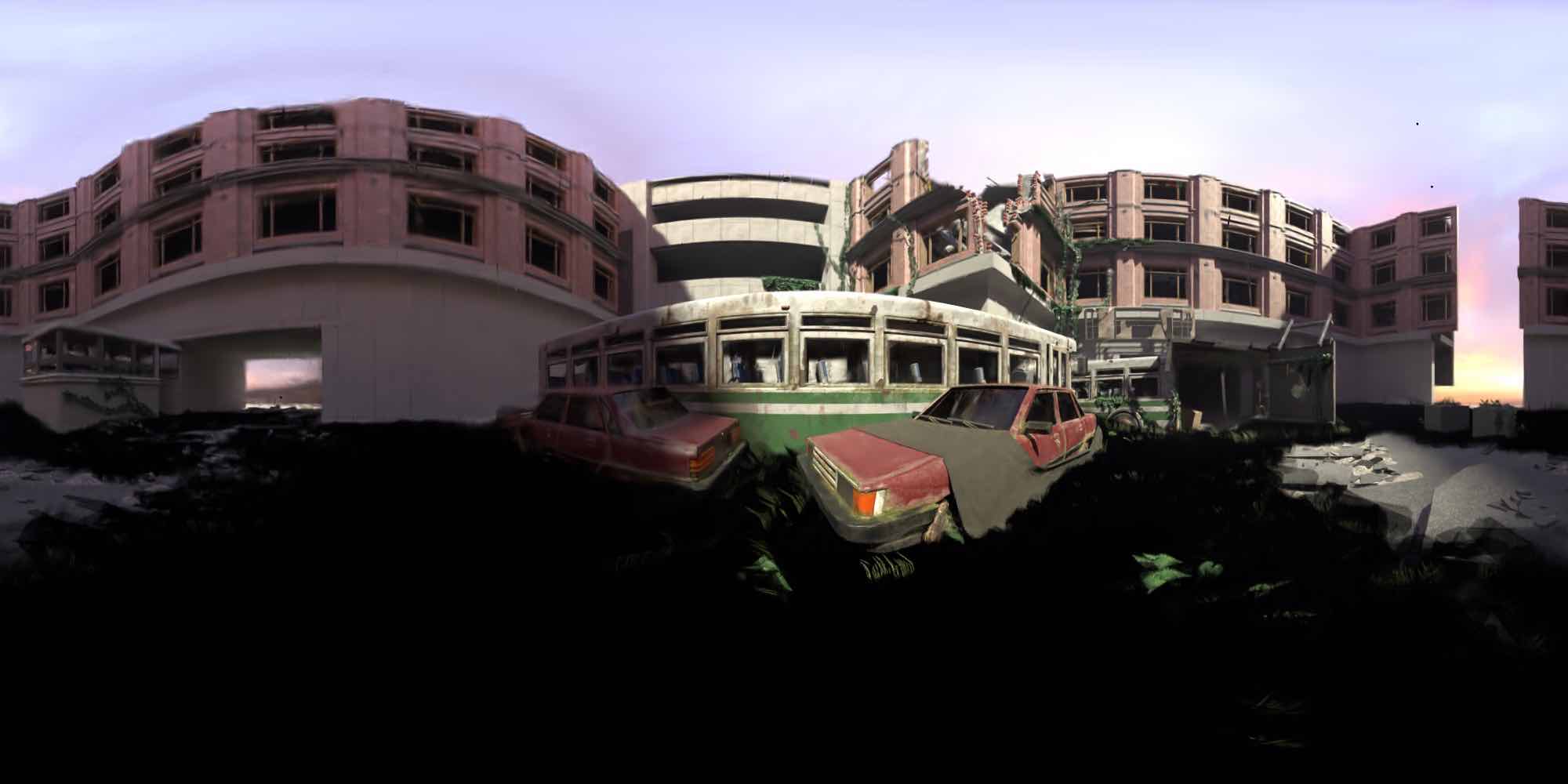} &
\includegraphics[width=.19\textwidth,valign=c]{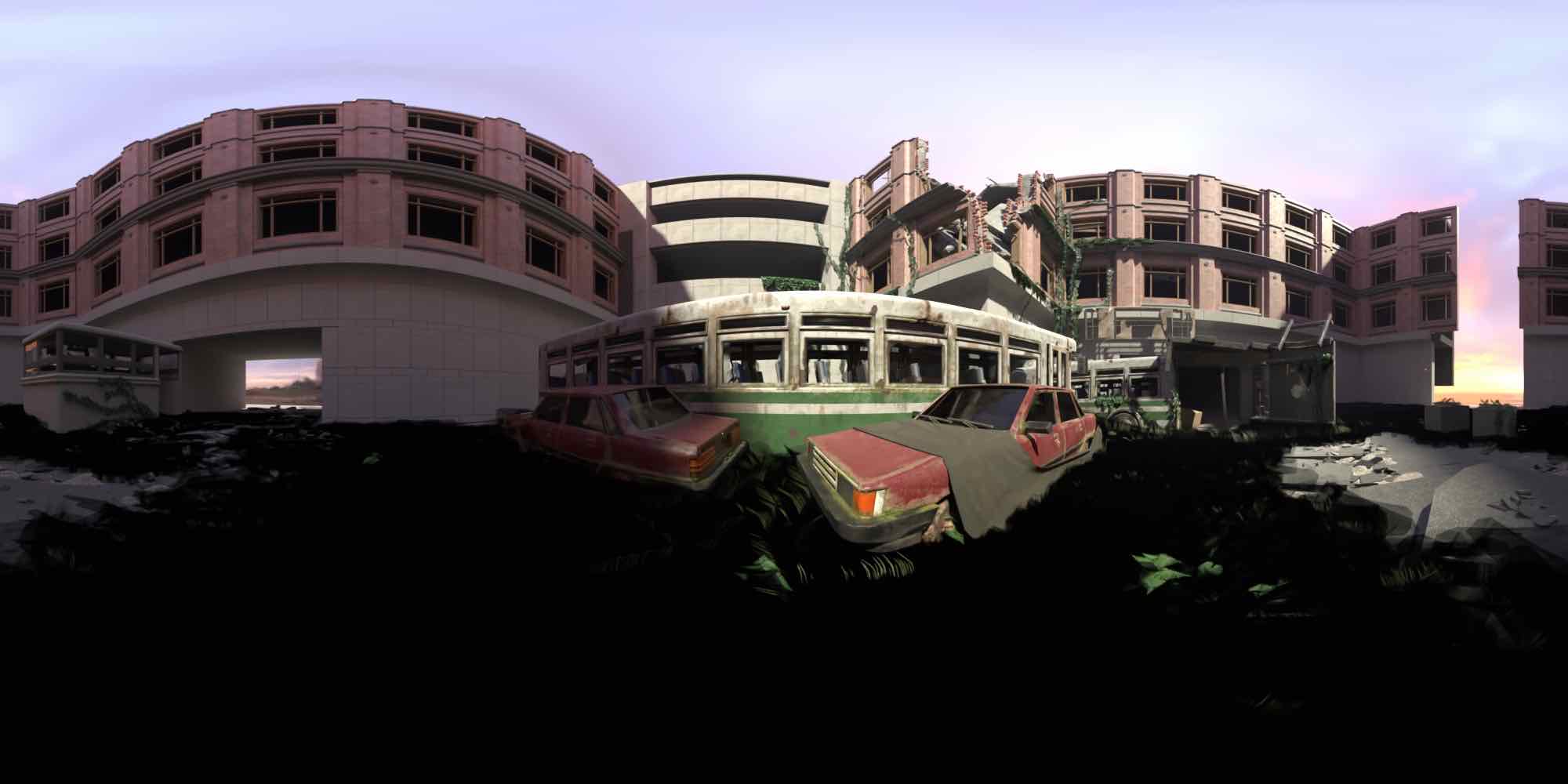} &
\includegraphics[width=.19\textwidth,valign=c]{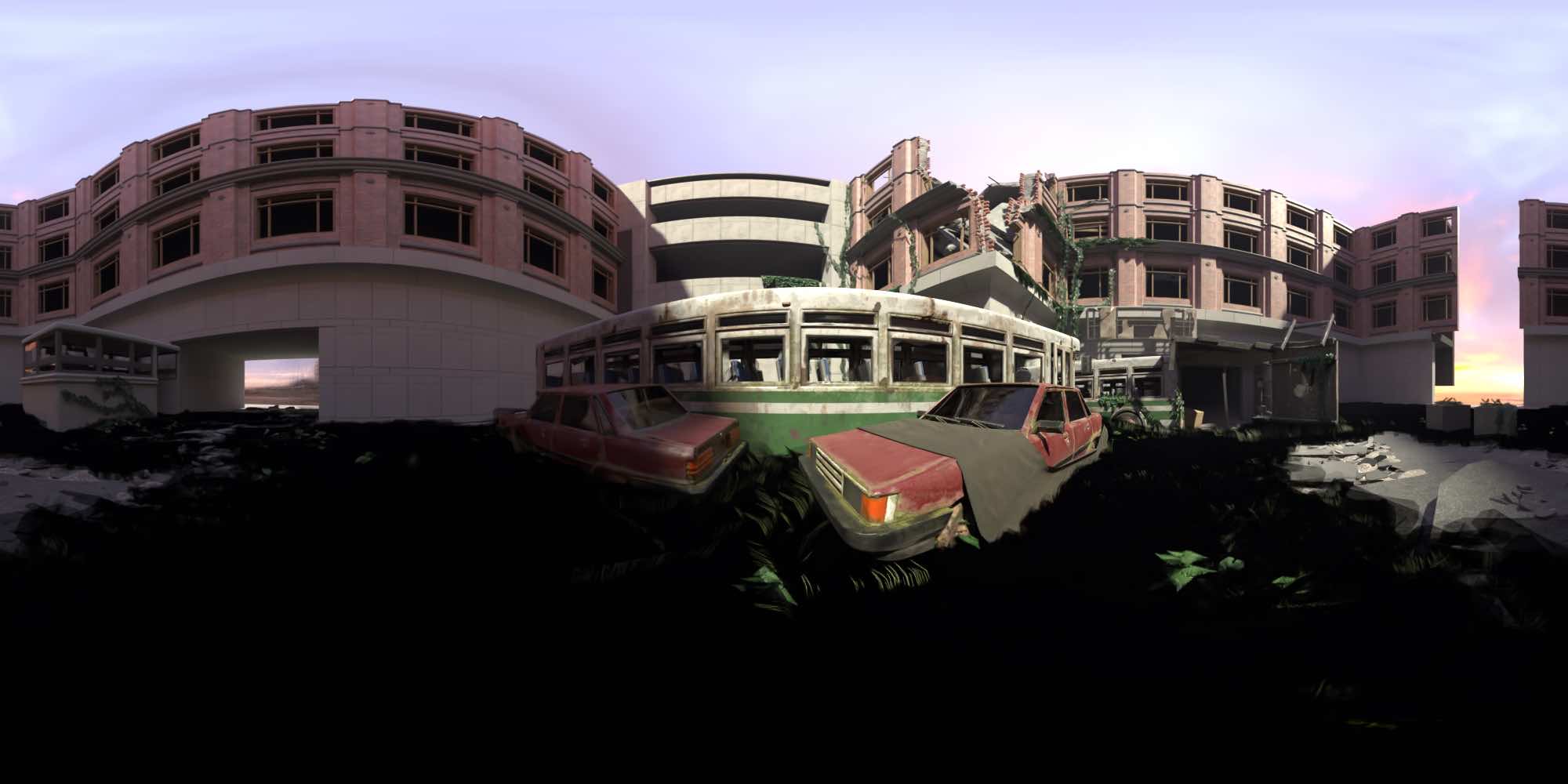} &
\includegraphics[width=.19\textwidth,valign=c]{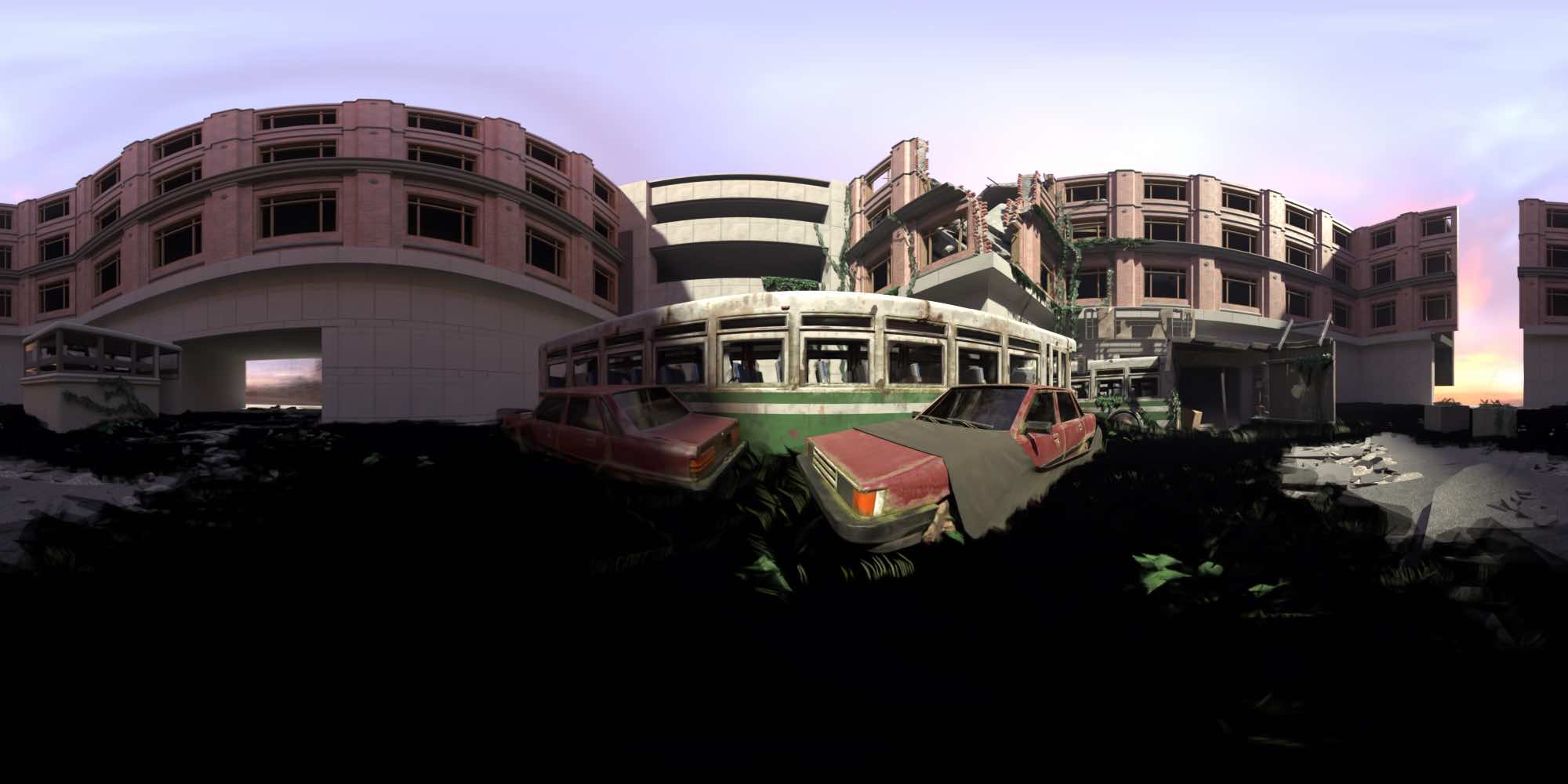} &
\includegraphics[width=.19\textwidth,valign=c]{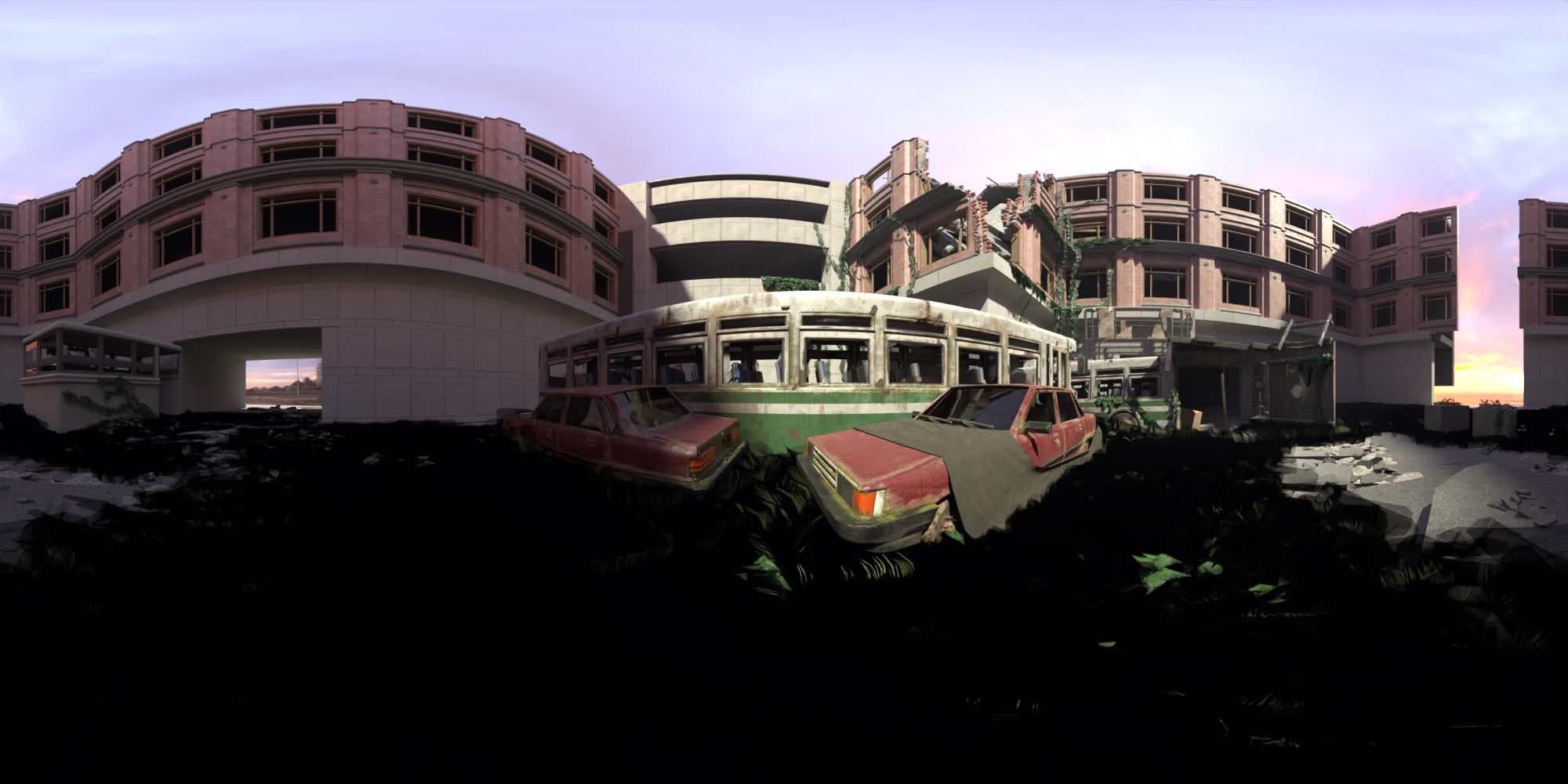} \\

\adjustbox{valign=c}{\rotatebox{90}{\scriptsize\textit{Yaw}~$+60^\circ$}} &
\includegraphics[width=.19\textwidth,valign=c]{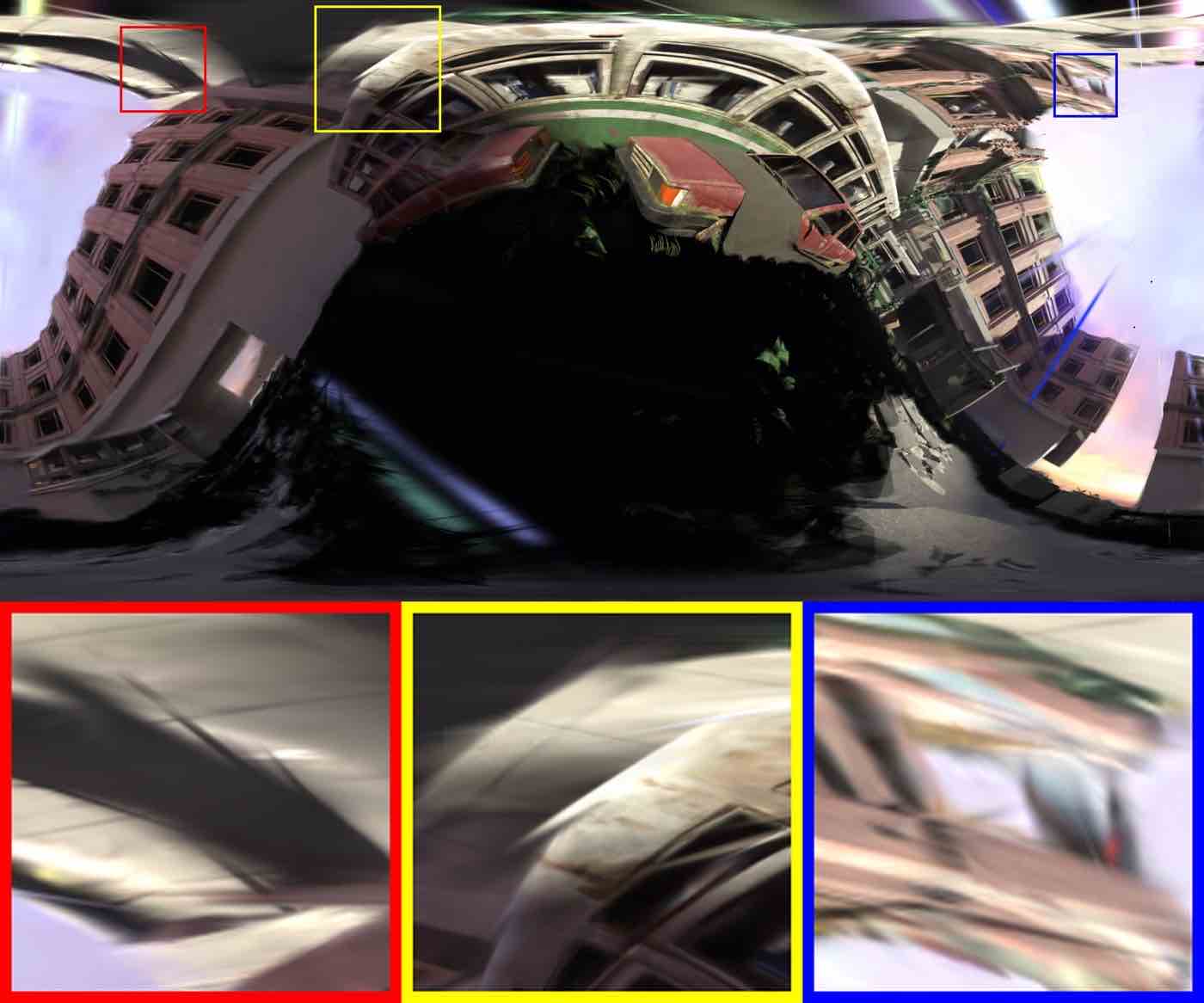} &
\includegraphics[width=.19\textwidth,valign=c]{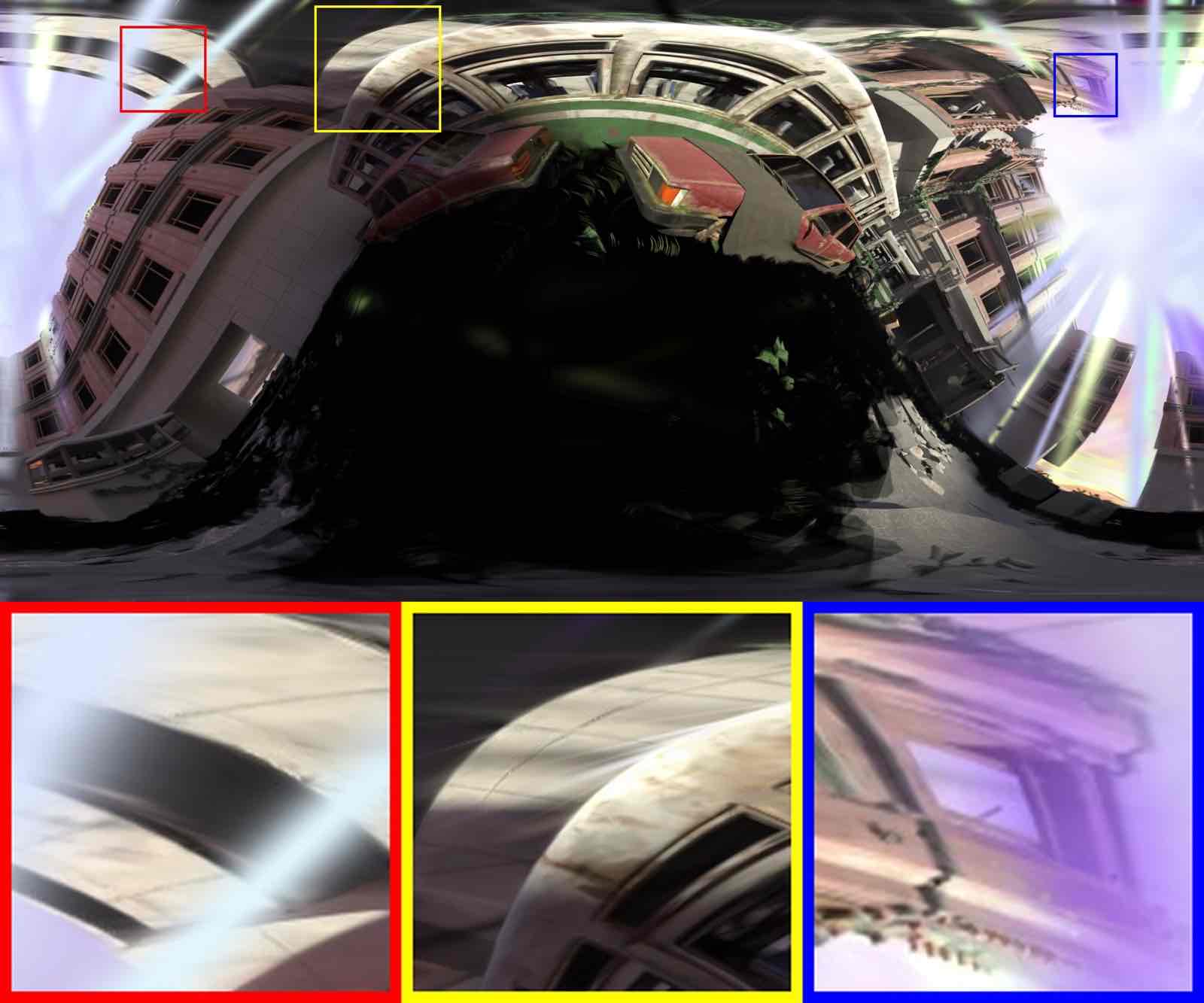} &
\includegraphics[width=.19\textwidth,valign=c]{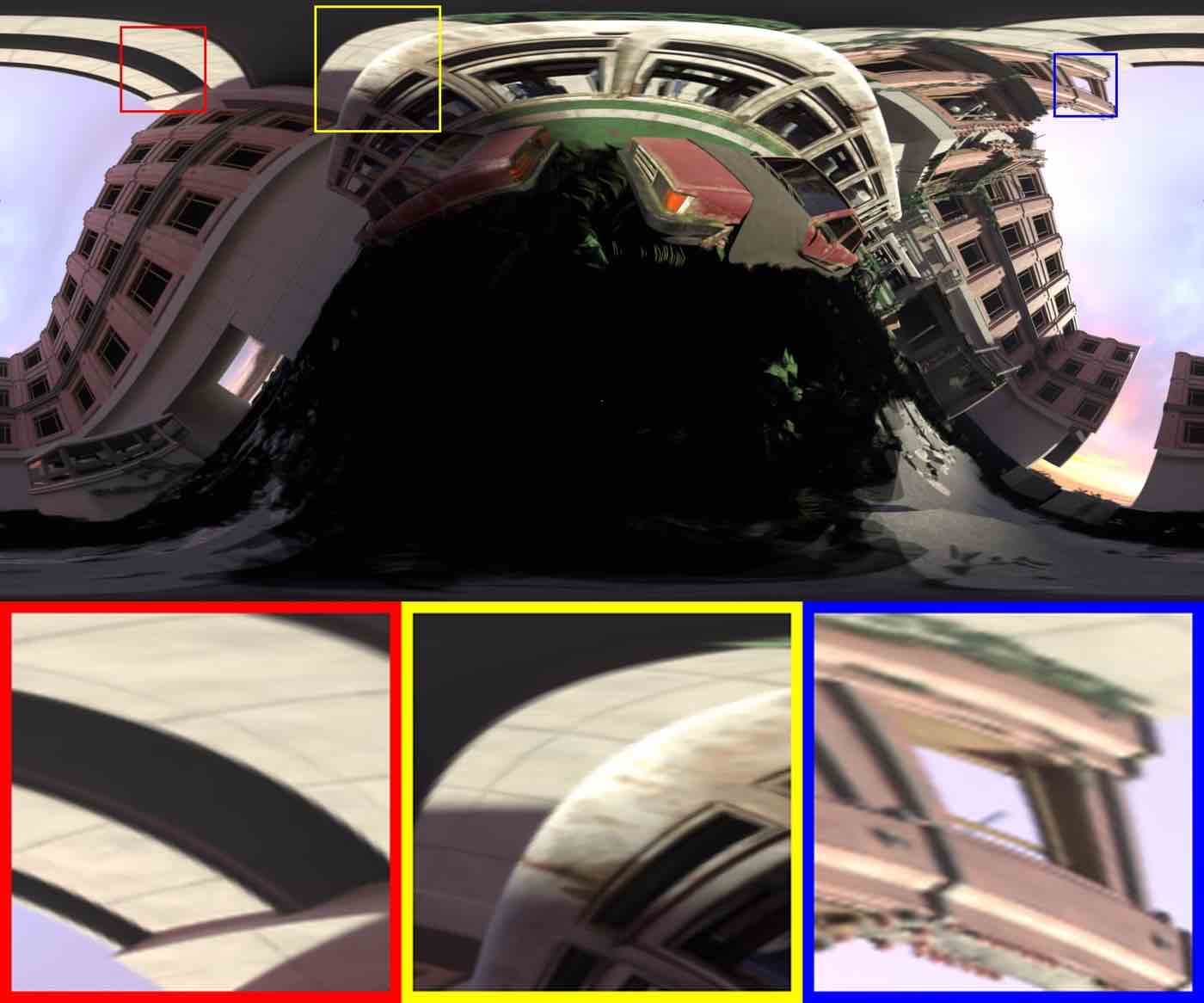} &
\includegraphics[width=.19\textwidth,valign=c]{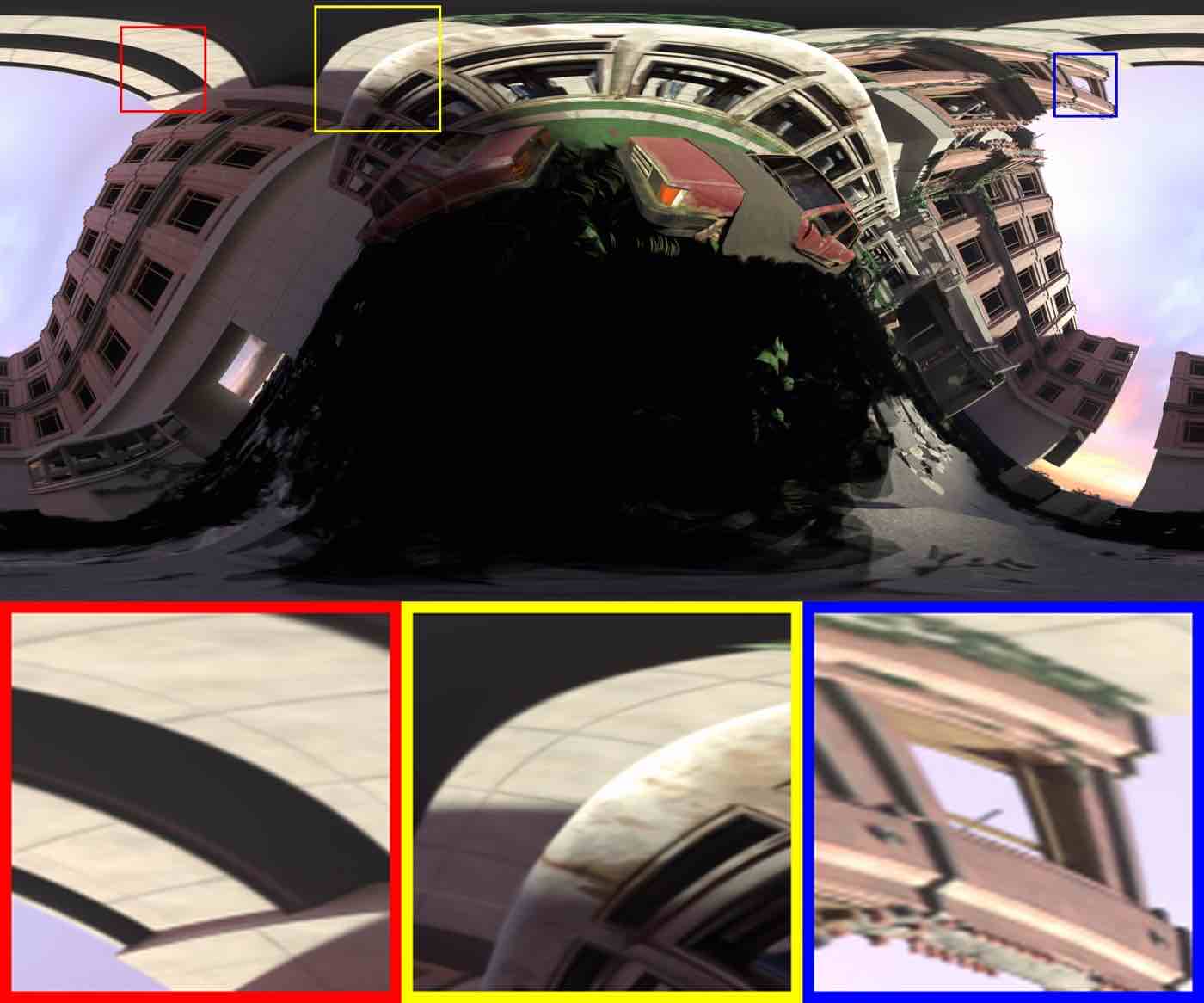} &
\includegraphics[width=.19\textwidth,valign=c]{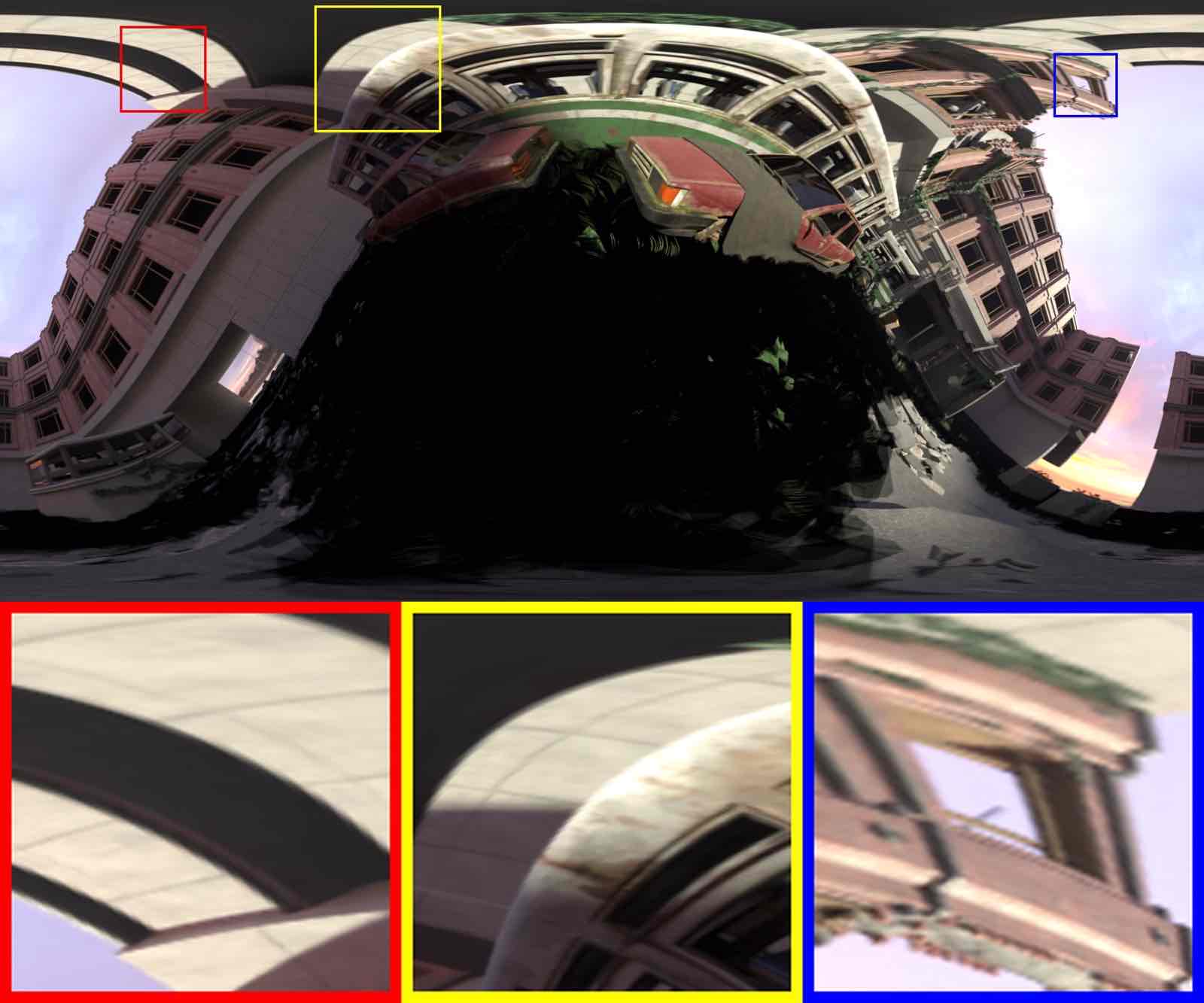} \\

\adjustbox{valign=c}{\rotatebox{90}{\scriptsize\textit{Yaw}~$+90^\circ$}} &
\includegraphics[width=.19\textwidth,valign=c]{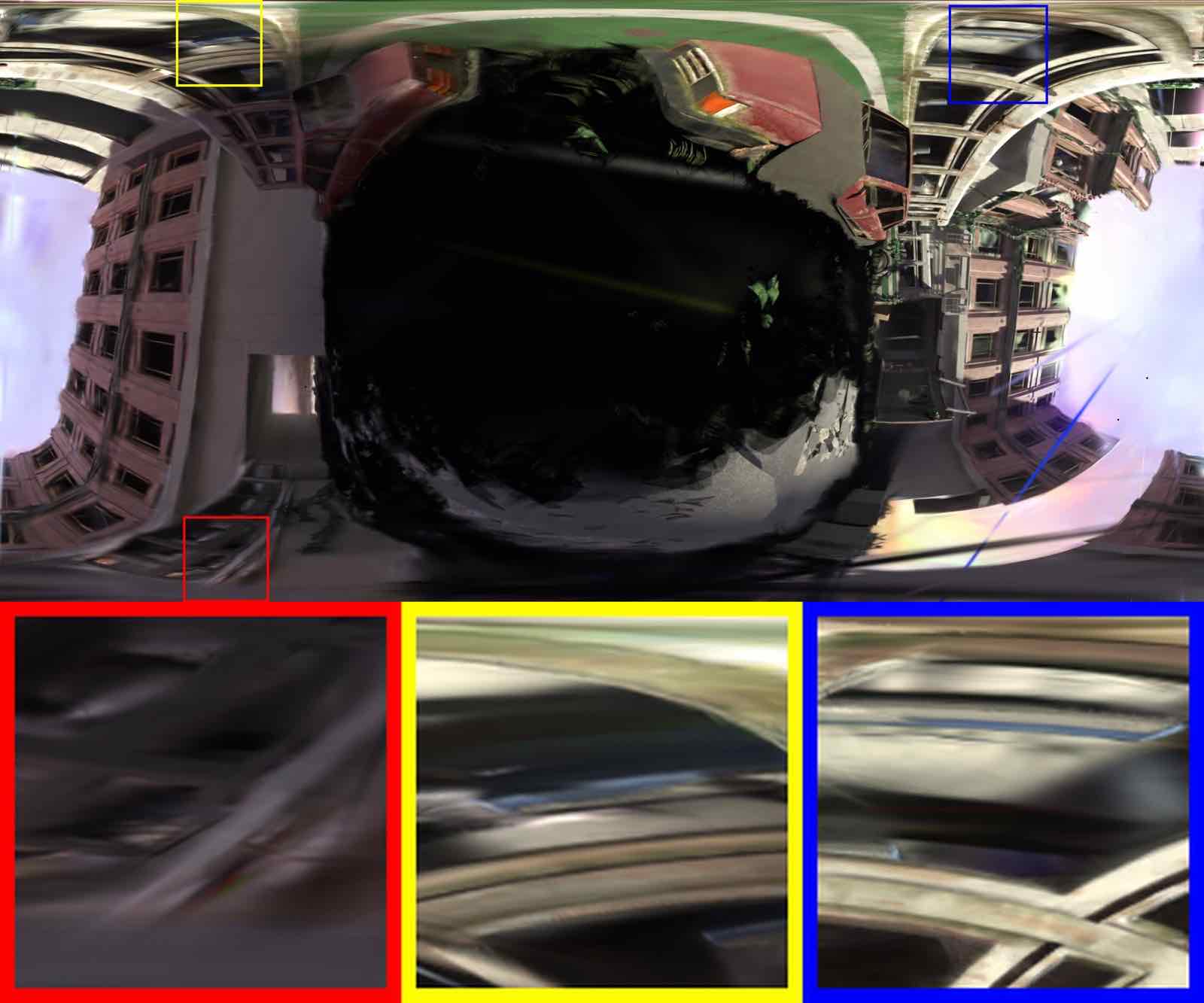} &
\includegraphics[width=.19\textwidth,valign=c]{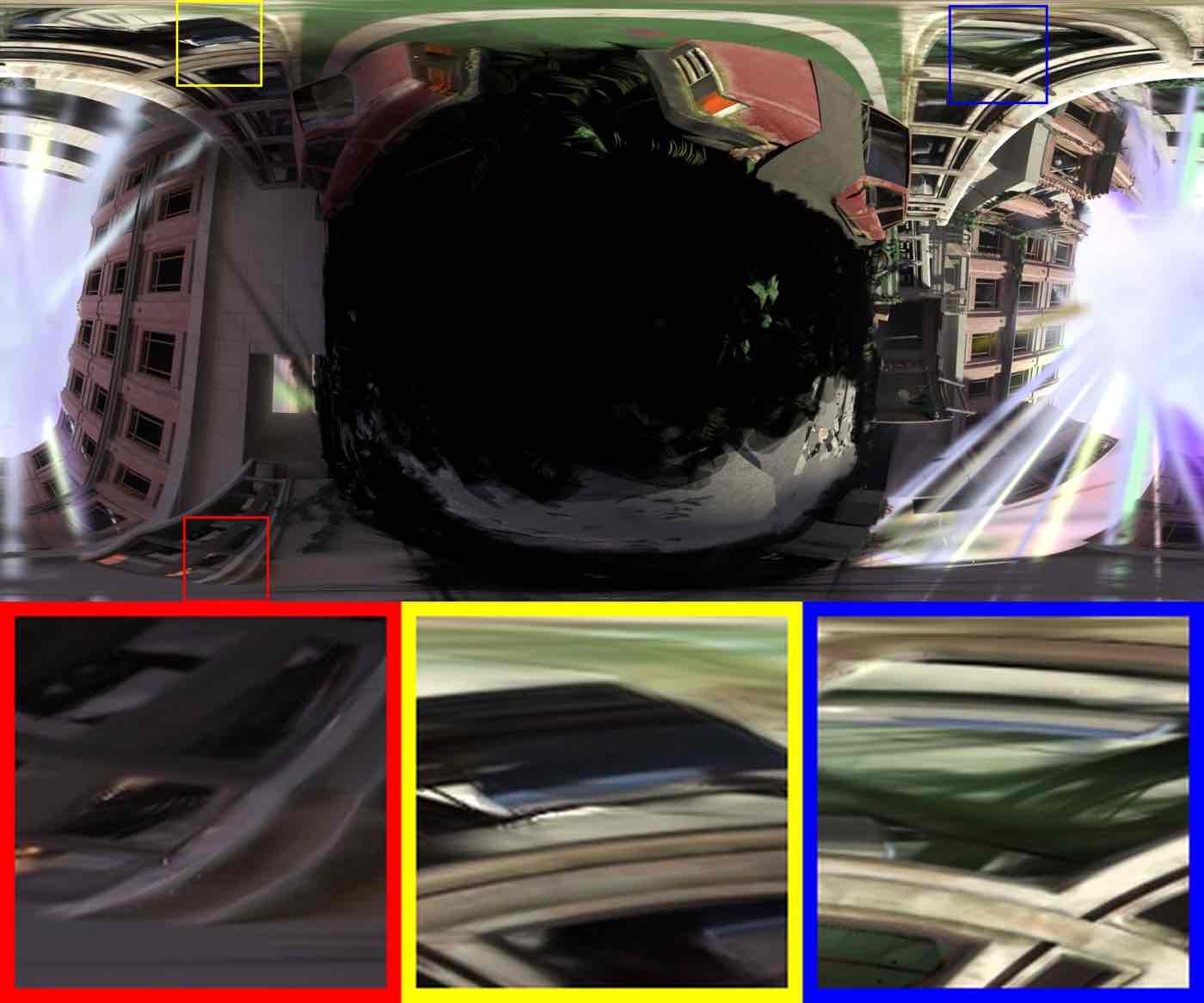} &
\includegraphics[width=.19\textwidth,valign=c]{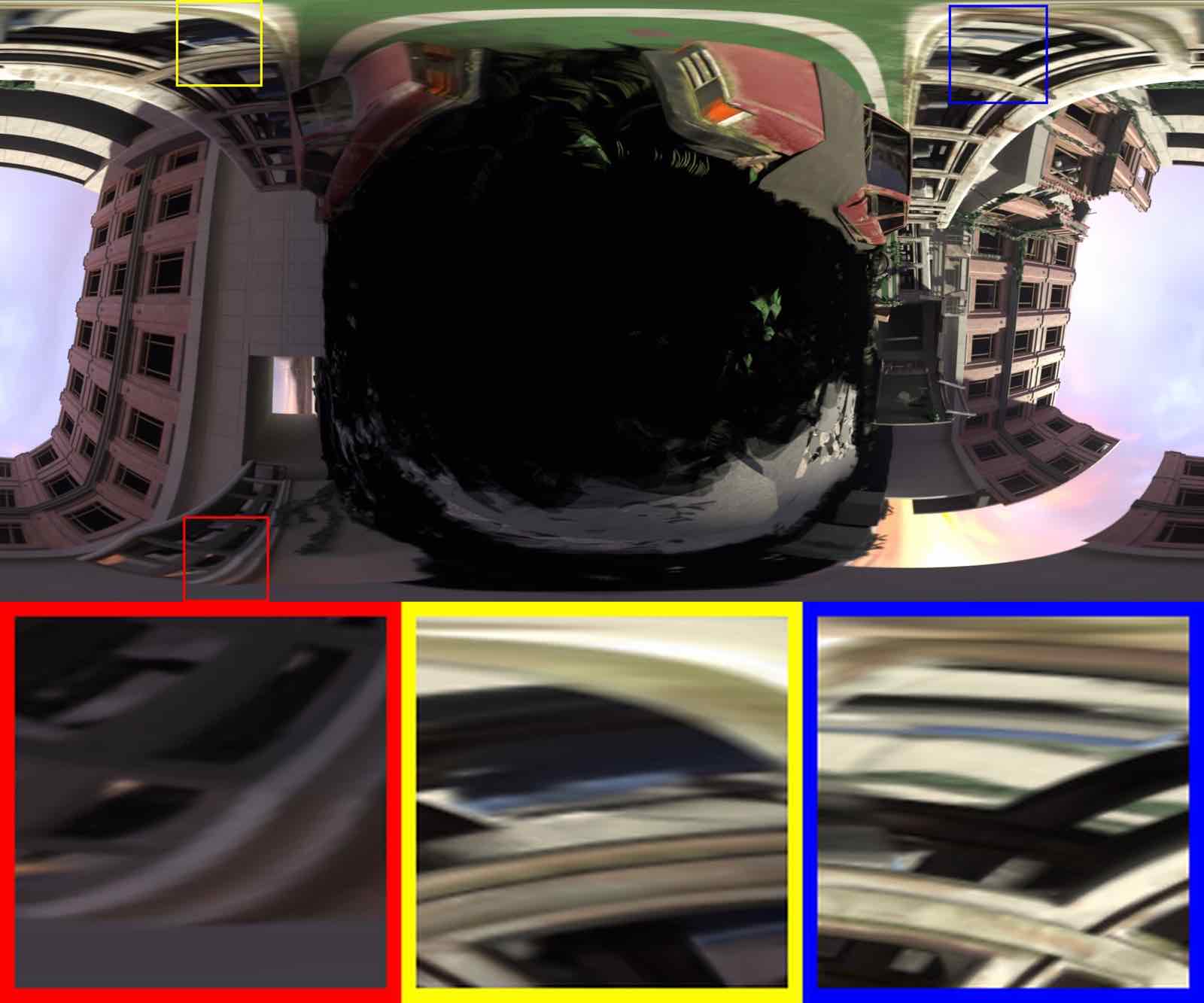} &
\includegraphics[width=.19\textwidth,valign=c]{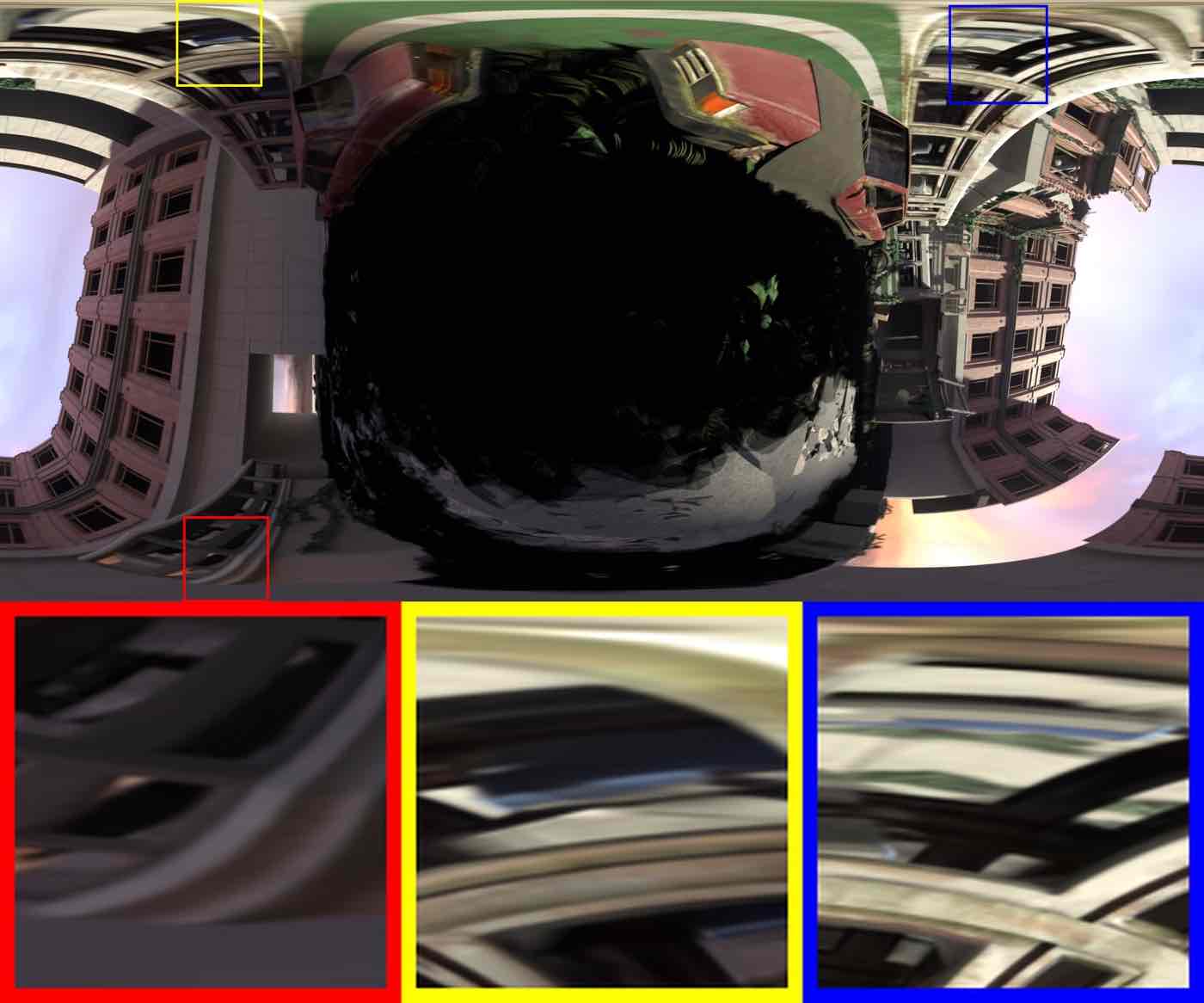} &
\includegraphics[width=.19\textwidth,valign=c]{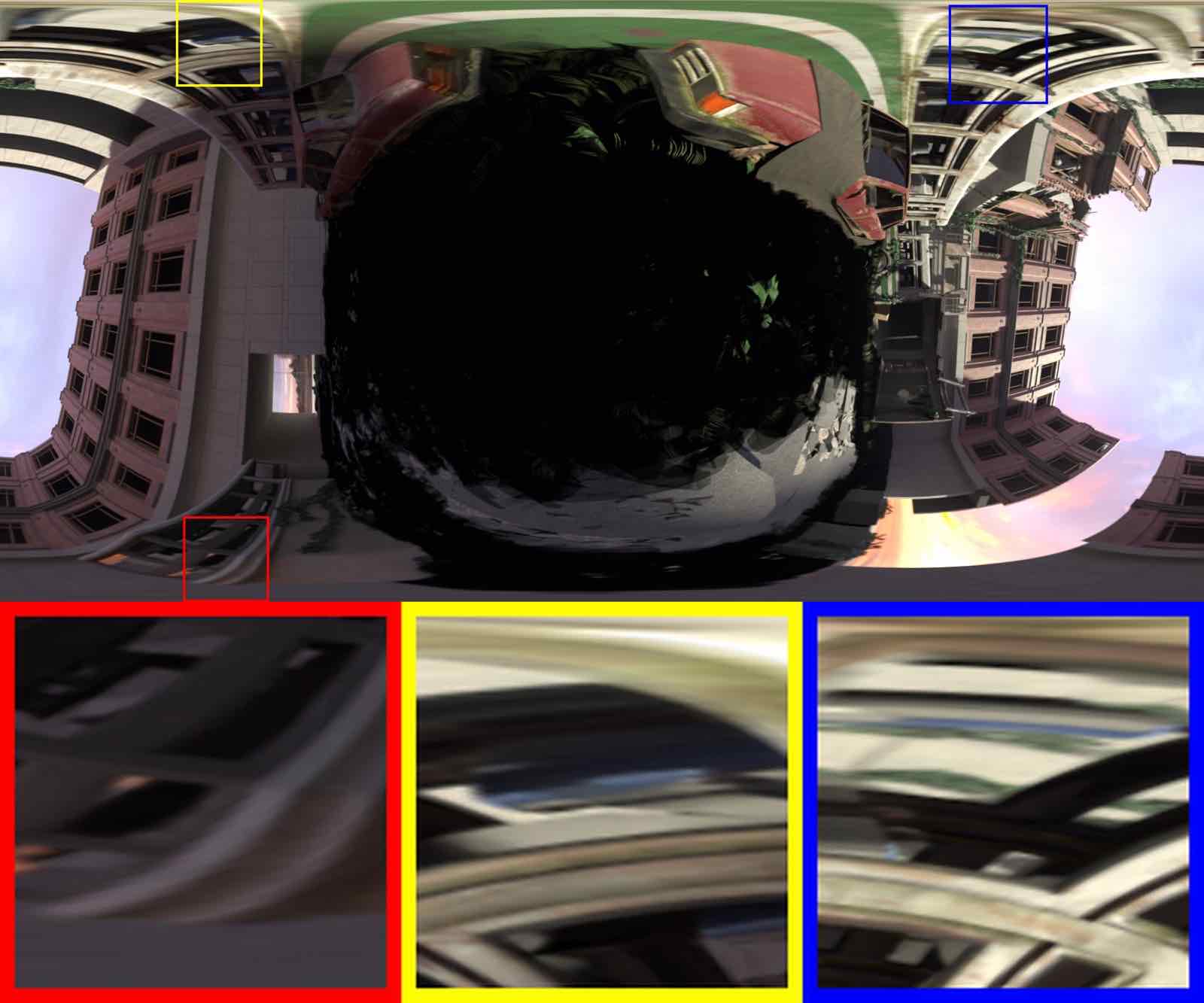} \\
\end{tabular}
\vskip-1ex
\caption{\textbf{Rotation robustness under global panorama rotations.}
We train with canonical poses and evaluate the same scene under additional global rotations of $0^\circ$, $60^\circ$, and $90^\circ$.
Each result includes a zoom-in inset from a fixed region for easier comparison.
Projection-based methods (ODGS~\cite{lee2024odgs} and OmniGS~\cite{li2025omnigs}) show increasing rotation-dependent degradation, while SPaGS~\cite{li2025spags} and our method remain more stable under large rotations.}
\label{fig:qual_rotation}
\vskip-3ex
\end{figure*}

\noindent\textbf{Evaluation metrics.}
We evaluate image-space rendering quality using PSNR, SSIM~\cite{wang2004image}, and LPIPS~\cite{zhang2018unreasonable}, where LPIPS is computed with the AlexNet backbone~\cite{krizhevsky2012imagenet}.

To assess multi-view geometric stability without ground-truth depth, we report a depth reprojection consistency error (DRE). For a pixel $\mathbf u$ in view $i$ with rendered depth $D_i(\mathbf u)$, we back-project it to 3D, transform the point to view $j$, and reproject it to obtain a correspondence $\mathbf u'$ and the predicted depth $D_{\text{proj}}(\mathbf u')$. We then sample the rendered depth of view $j$ at $\mathbf u'$ by bilinear interpolation, denoted $D_j(\mathbf u')$, and define
\begin{equation}
e(\mathbf u)=
\frac{\left|D_{\text{proj}}(\mathbf u')-D_j(\mathbf u')\right|}
{D_j(\mathbf u')+\epsilon}.
\end{equation}
We report the mean of $e(\mathbf u)$ over all valid overlapping pixels across the same set of view pairs. A pixel is valid if $\mathbf u'$ lies inside the image domain of view $j$ and both $D_i(\mathbf u)$ and $D_j(\mathbf u')$ are defined.

We additionally report the Cycle Inlier Ratio (CIR), which measures the fraction of pixels that remain consistent under a forward-backward reprojection cycle. Starting from $\mathbf u$ in view $i$, we reproject to view $j$ to obtain $\mathbf u'$, and then reproject $\mathbf u'$ back to view $i$ to obtain $\hat{\mathbf u}$. We count $\mathbf u$ as an inlier if the round-trip pixel error $\|\hat{\mathbf u}-\mathbf u\|_2$ is below a threshold $\tau_{\text{cyc}}$ (we use $\tau_{\text{cyc}}=2$ pixels). CIR is computed as the inlier ratio over the same valid overlapping pixels and view pairs. Lower DRE and higher CIR indicate more stable, view-consistent geometry. Intuitively, DRE measures the severity of cross-view depth drift, while CIR reflects the coverage of view-consistent geometry. 

\subsection{Experiment Results}
\noindent\textbf{Quantitative comparison.}
Table~\ref{tab:main} summarizes photometric quality and multi-view geometry consistency on OmniBlender and OmniPhotos.
Overall, OmniGS~\cite{li2025omnigs} and SPaGS~\cite{li2025spags} achieve strong photometric performance, and our method remains competitive on image-space metrics.
More importantly, our method consistently improves geometric consistency across all benchmarks, achieving the lowest DRE and the highest CIR in every setting.
On OmniBlender-Indoor, our method reduces DRE by $62.7\%$ and increases CIR by $22.6\%$ relative to SPaGS, indicating substantially more view-consistent depth on scenes with large planar structures.
Similar improvements are observed on OmniBlender-Outdoor and OmniPhotos, demonstrating stable cross-view geometry under both synthetic and real-world panoramas. This gain comes with a trade-off in training efficiency: SPaGS is the most training-efficient baseline and typically converges in tens of minutes, whereas our current implementation converges in about 1 hour.

\noindent\textbf{Qualitative comparison.}
Fig.~\ref{fig:qual_geom} presents qualitative comparisons on OmniBlender, where large planar surfaces make geometric artifacts visible. Although all methods achieve plausible RGB renderings, their geometry noticeably. Projection-based panoramic 3DGS baselines often produce texture-aligned depth ripples, causing planar regions to become wavy or distorted. In contrast, our method yields smoother depth maps with cleaner discontinuities and fewer appearance-induced artifacts. We further visualize surface orientation using normal maps computed from rendered depth for all methods to ensure a fair comparison. Despite the sensitivity of depth-to-normal conversion under ERP sampling, our results show more coherent normals on planar areas and are less correlated with image textures, which is consistent with the geometric improvements reported in Table~\ref{tab:main}.

\noindent\textbf{Rotation robustness.}
Panoramic projection is highly nonlinear, and projection-based 3DGS variants often approximate it using screen-space local linearization.
This approximation becomes less accurate in strongly distorted regions and can induce orientation-dependent rendering.
We evaluate rotation robustness on OmniBlender by training with canonical poses and testing with additional random global rotations within $\pm\theta$.
Quantitative results are reported in Table~\ref{tab:rot_full}, and qualitative comparisons are shown in Fig.~\ref{fig:qual_rotation}.

\begin{table}[!t]
\centering
\footnotesize
\setlength{\tabcolsep}{3.5pt}
\renewcommand{\arraystretch}{1.05}
\begin{tabular}{l|ccc|cc}
\toprule
Variant
& DRE$\downarrow$ & CIR$\uparrow$ 
& PSNR$\uparrow$ & SSIM$\uparrow$ & LPIPS$\downarrow$\\
\midrule
Full (Ours)  & 0.0439 & 84.87 & 32.61 & 0.9195 & 0.0687 \\
w/o $\mathcal{L}_{dn}$  & 0.0397 & 85.25 & 32.70 & 0.9203 & 0.0669 \\
w/o $\mathcal{L}_{\text{jump}}$  & 0.0692 & 76.87 & 32.75 & 0.9212 & 0.0668 \\
w/o $\mathcal{L}_{dn}$ and $\mathcal{L}_{\text{jump}}$  & 0.0577 & 77.66 & 32.84 & 0.9222 & 0.0648 \\
\bottomrule
\end{tabular}
\caption{Ablation study on OmniBlender~\cite{choi2023balanced}.}
\label{tab:ablation}
\end{table}

\begin{figure}[!t]
  \centering
  \setlength{\tabcolsep}{1.5pt}
  \renewcommand{\arraystretch}{1.0}
  \begin{tabular}{cc}
    \includegraphics[width=0.49\columnwidth]{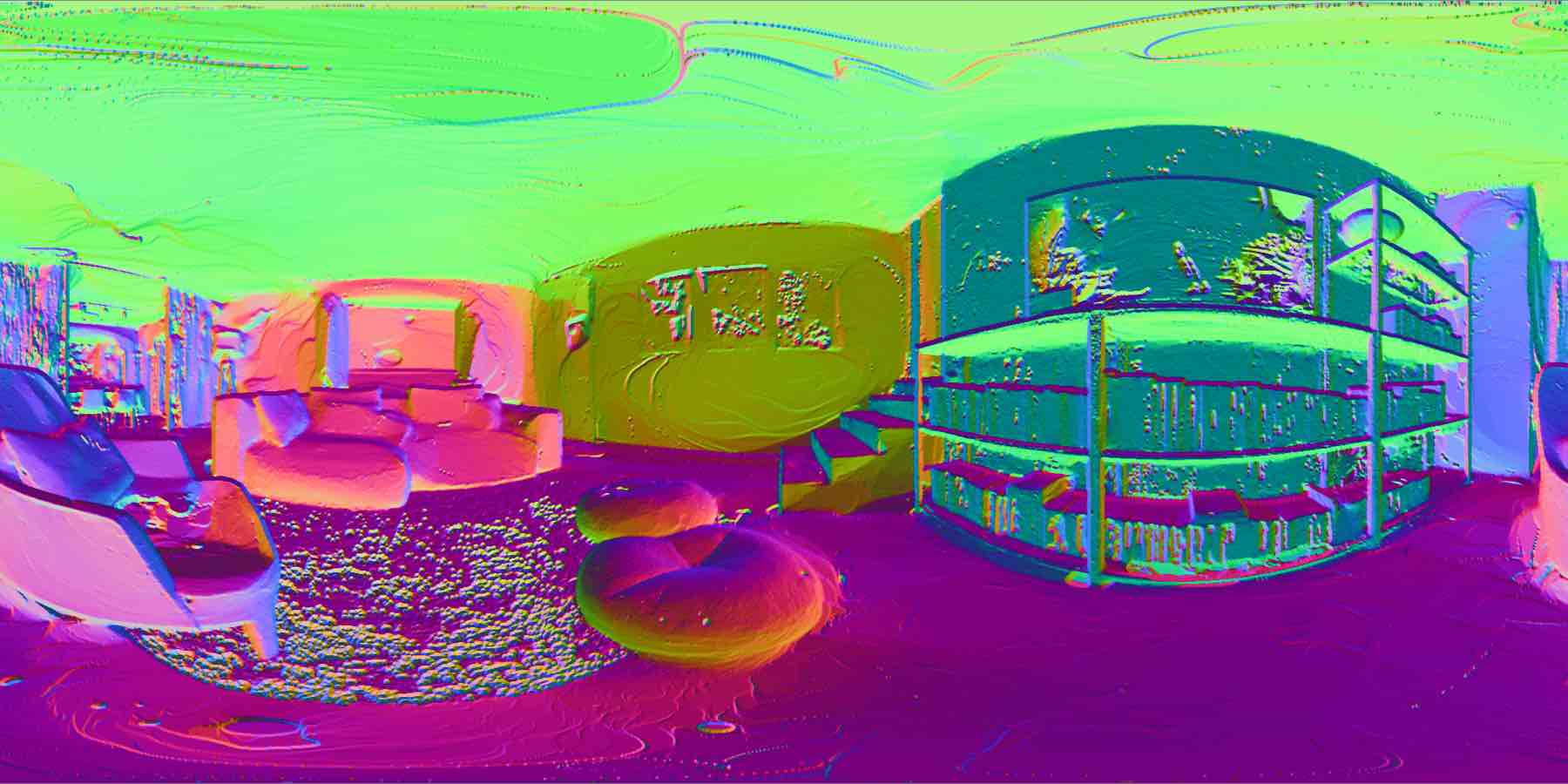} &
    \includegraphics[width=0.49\columnwidth]{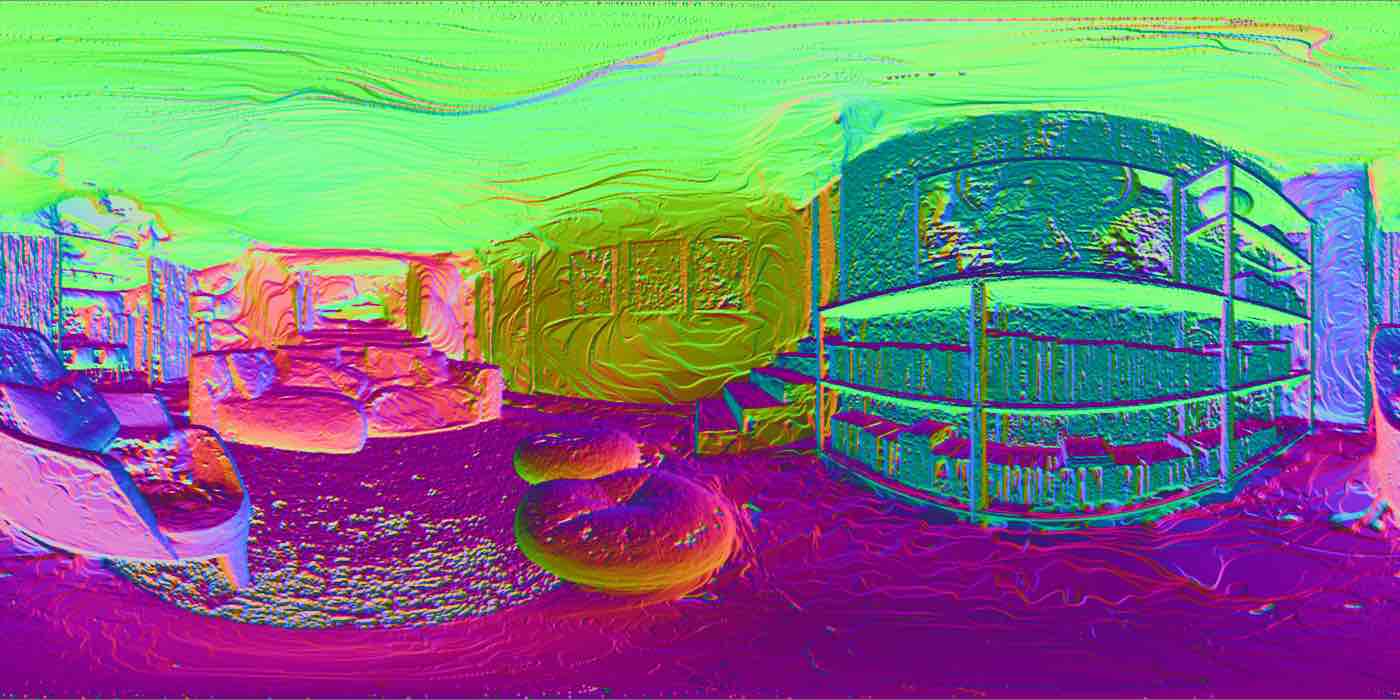} \\
    \scriptsize (a) Full  &
    \scriptsize (b) w/o $\mathcal{L}_{dn}$ \\[0.6ex]
    \includegraphics[width=0.49\columnwidth]{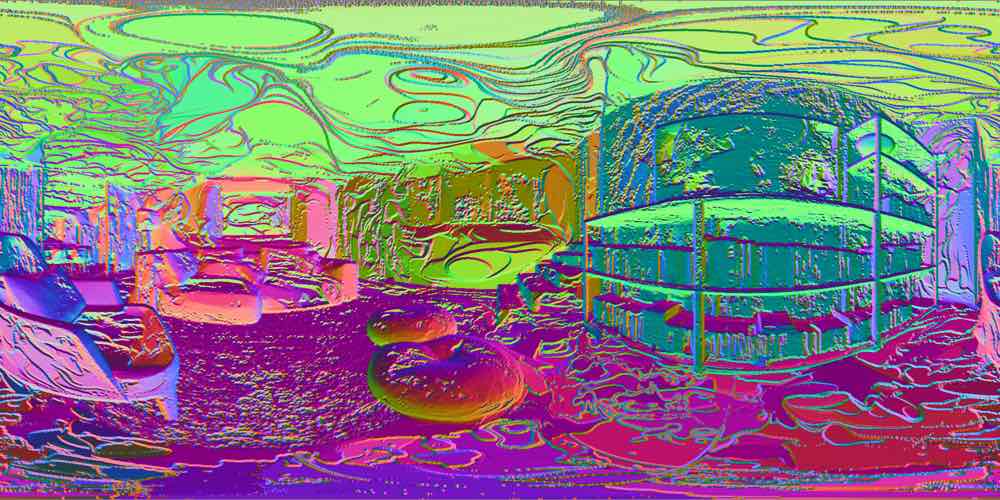} &
    \includegraphics[width=0.49\columnwidth]{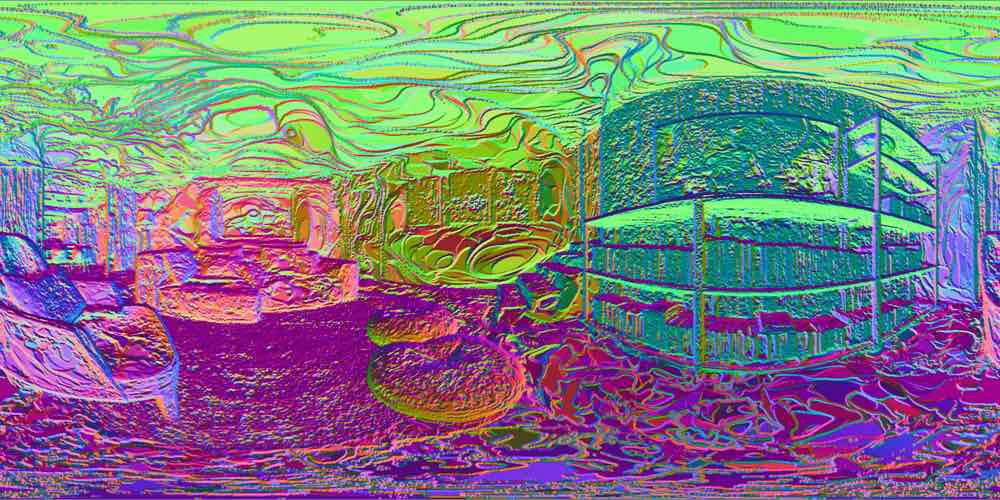} \\
    \scriptsize (c) w/o $\mathcal{L}_{\text{jump}}$ &
    \scriptsize (d) w/o $\mathcal{L}_{dn}$ and $\mathcal{L}_{\text{jump}}$ \\
  \end{tabular}
  \caption{Qualitative ablation on OmniBlender~\cite{choi2023balanced}.}
  \label{fig:ablation_vis}
\end{figure}

Projection-based methods, including ODGS and OmniGS, degrade progressively as the rotation magnitude increases.
At $\theta{=}90^\circ$, OmniGS suffers a $32\%$ PSNR drop and a $74\%$ LPIPS increase, while ODGS shows a smaller but still notable $16\%$ PSNR degradation.
Qualitatively, Fig.~\ref{fig:qual_rotation} shows blurred zoom-in regions and high-latitude distortions after rotation, which are consistent with orientation-sensitive screen-space linearization.
In contrast, SPaGS and our method remain much more stable under large rotations.
At $\theta{=}90^\circ$, our method reduces PSNR by only about $7\%$, while maintaining consistently low LPIPS, and SPaGS stays nearly invariant.
DRE and CIR fluctuate less across rotations because they measure cross-view consistency with outlier filtering and are less sensitive to purely photometric degradation.
Overall, these results indicate that ray-space Gaussian evaluation improves rotation robustness for panoramic rendering.

\noindent\textbf{Ablation.}
Table~\ref{tab:ablation} reports the ablation of $\mathcal{L}_{dn}$ and $\mathcal{L}*{\text{jump}}$. Removing $\mathcal{L}*{\text{jump}}$ clearly harms geometric consistency, as indicated by higher DRE and lower CIR. Removing $\mathcal{L}*{dn}$ yields slightly better numerical scores, because $\mathcal{L}*{dn}$ is a perceptual geometry regularizer that promotes locally smooth depth and normal fields rather than directly optimizing the proxy metrics. Since DRE/CIR mainly measure cross-view consistency, they do not fully capture local surface smoothness or ripple-like artifacts. As shown in Fig.~\ref{fig:ablation_vis}, $\mathcal{L}*{dn}$ improves the visual quality of planar regions and normals, despite a small trade-off in the quantitative scores.

\begin{table}[!t]
\centering
\vspace{-1ex}
\footnotesize
\setlength{\tabcolsep}{3.5pt}
\renewcommand{\arraystretch}{1.05}
\begin{tabular}{l|ccc|cc}
\toprule
Setting / Method
& DRE$\downarrow$ & CIR$\uparrow$ 
& PSNR$\uparrow$ & SSIM$\uparrow$ & LPIPS$\downarrow$\\
\midrule
\multicolumn{6}{l}{\textbf{OmniRob-UAV (full panorama)}} \\
SPaGS~\cite{li2025spags} & 0.0715 & 73.44 & 35.28 & 0.9515 & 0.1908 \\
Ours  & 0.0184 & 92.34 & 33.05 & 0.9297 & 0.0858 \\
\midrule
\multicolumn{6}{l}{\textbf{OmniRob-Quadruped (annular camera)}} \\
SPaGS~\cite{li2025spags} & 0.2614 & 19.49 & 19.90 & 0.7492 & 0.3272 \\
Ours  & 0.1568 & 47.63 & 19.60 & 0.7517 & 0.2077 \\
\midrule
\multicolumn{6}{l}{\textbf{Cropped-UAV (pseudo-annular)}} \\
SPaGS~\cite{li2025spags} & 0.0770 & 83.20 & 35.24 & 0.9550 & 0.1098 \\
Ours  & 0.0270 & 90.82 & 30.72 & 0.8879 & 0.1210 \\
\bottomrule
\end{tabular}
\caption{Analysis on OmniRob. 
We compare SPaGS and ours on OmniRob-UAV, OmniRob-Quadruped, and the Cropped-UAV annular setting.}
\label{tab:omnirob_transfer}
\vspace{-2ex}
\end{table}

{%
\begin{figure}[!t]
  \centering
  \setlength{\tabcolsep}{0.6pt}
  \renewcommand{\arraystretch}{1.0}

  \newlength{\imgw}
  \setlength{\imgw}{\dimexpr\columnwidth-9pt-9pt-2pt-2pt\relax}

  \begin{tabular}{@{}>{\centering\arraybackslash}m{9pt}@{\hspace{2pt}}>{\centering\arraybackslash}m{9pt}@{\hspace{2pt}}c@{}}
    & & \small OmniRob-UAV \\[0.2ex]

    \mbox{} &
    \adjustbox{valign=c}{$\vcenter{\hbox{\rotatebox{90}{\scriptsize GT}}}$} &
    \makebox[\imgw][c]{\includegraphics[width=\imgw,keepaspectratio,valign=c]{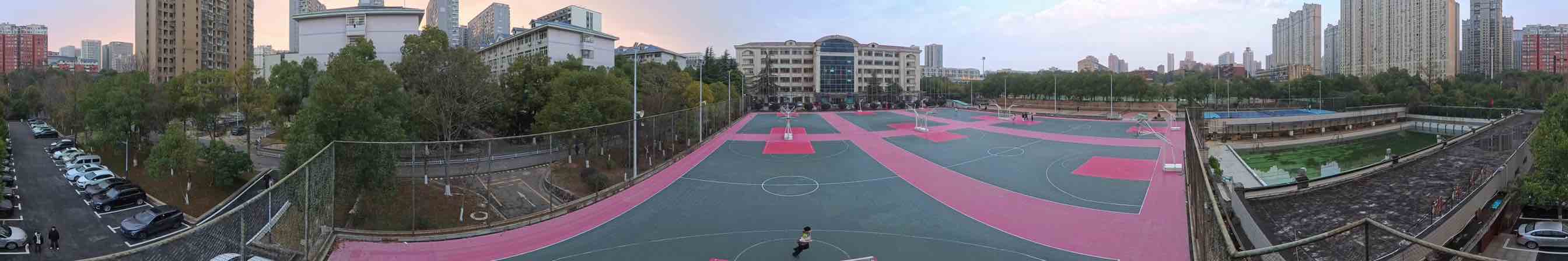}} \\[0.45ex]

    \multirow[c]{2}{*}[-3.0ex]{\adjustbox{valign=c}{$\vcenter{\hbox{\rotatebox{90}{\scriptsize SPaGS}}}$}} &
    \adjustbox{valign=c}{$\vcenter{\hbox{\rotatebox{90}{\scriptsize RGB}}}$} &
    \makebox[\imgw][c]{\includegraphics[width=\imgw,keepaspectratio,valign=c]{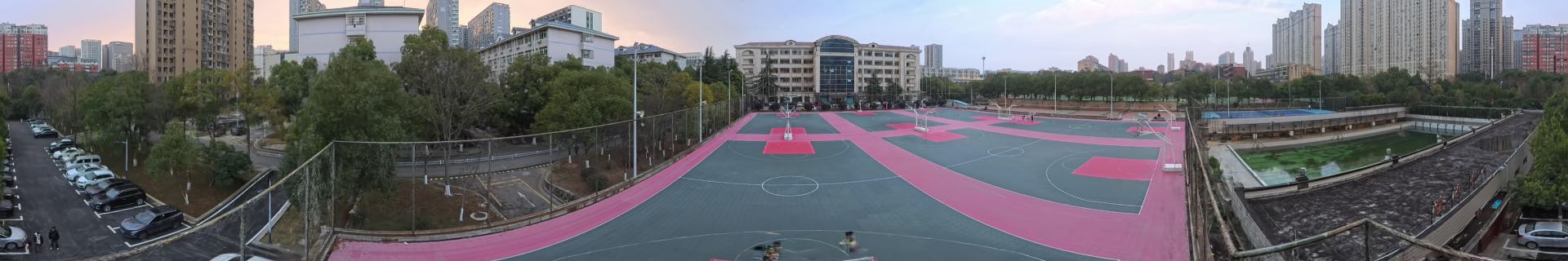}} \\
    &
    \adjustbox{valign=c}{$\vcenter{\hbox{\rotatebox{90}{\scriptsize Normal}}}$} &
    \makebox[\imgw][c]{\includegraphics[width=\imgw,keepaspectratio,valign=c]{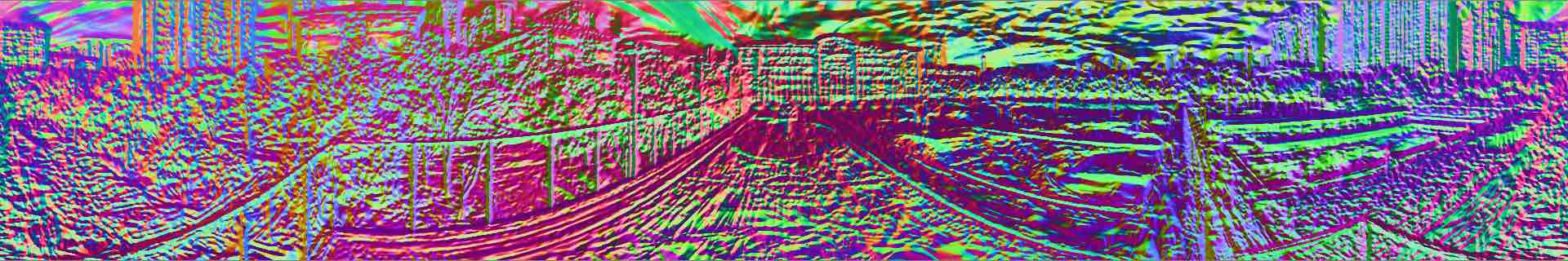}} \\[0.45ex]

    \multirow[c]{2}{*}[-3.0ex]{\adjustbox{valign=c}{$\vcenter{\hbox{\rotatebox{90}{\scriptsize Ours}}}$}} &
    \adjustbox{valign=c}{$\vcenter{\hbox{\rotatebox{90}{\scriptsize RGB}}}$} &
    \makebox[\imgw][c]{\includegraphics[width=\imgw,keepaspectratio,valign=c]{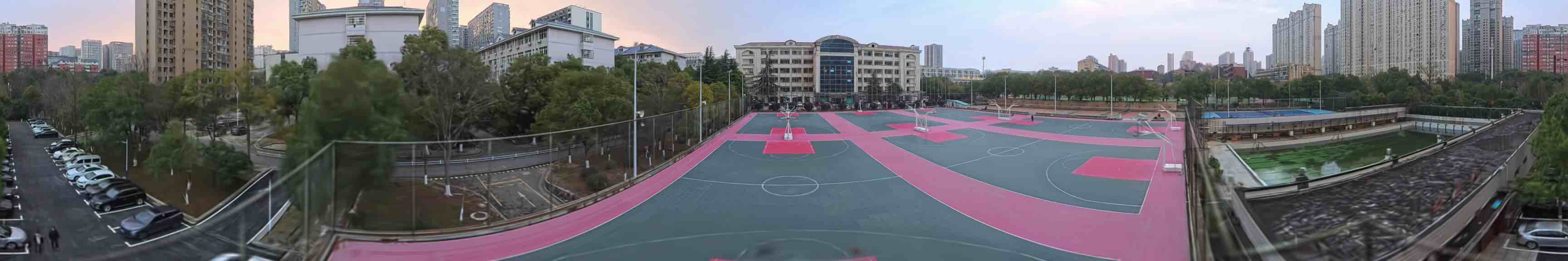}} \\
    &
    \adjustbox{valign=c}{$\vcenter{\hbox{\rotatebox{90}{\scriptsize Normal}}}$} &
    \makebox[\imgw][c]{\includegraphics[width=\imgw,keepaspectratio,valign=c]{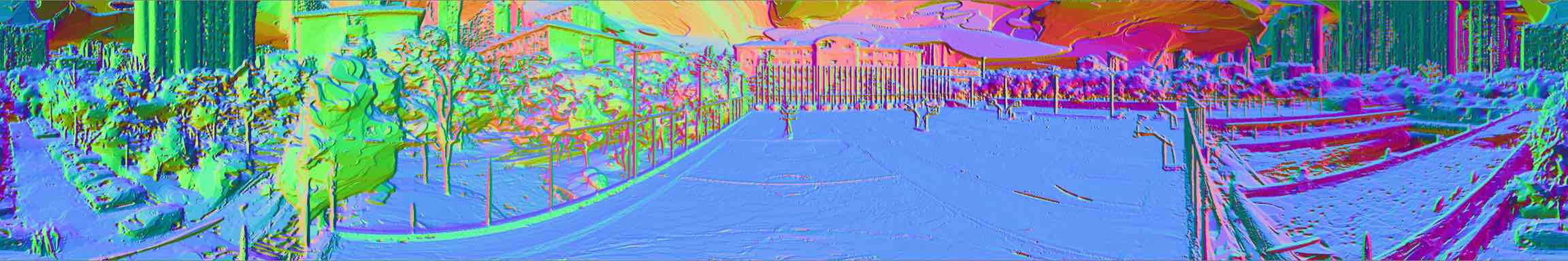}} \\
  \end{tabular}
  \caption{Qualitative results on OmniRob annular panoramas. Rows show GT, SPaGS (RGB/normal), and Ours (RGB/normal).}
  \label{fig:omnirob_ring_qual}
\end{figure}
}%

\noindent\textbf{Adaptation to annular panoramic cameras.}
We further evaluate whether the proposed spherical formulation can adapt to different omnidirectional camera parameterizations. Unlike stitched panoramas, annular observations avoid stitching seams and photometric inconsistencies since they are obtained from a single omnidirectional capture~\cite{shin2025seam360gs}. OmniRob contains full equirectangular panoramas from the UAV platform and real annular observations from the quadruped platform. Since the quadruped sequences are affected by capture artifacts and annular unwrapping, we additionally construct Cropped-UAV as a controlled pseudo-annular setting by cropping full panoramas to a similar vertical FoV. As shown in Table~\ref{tab:omnirob_transfer} and Fig.~\ref{fig:omnirob_ring_qual}, both SPaGS and our method can be applied to these settings with only minor changes to the ray generation range. SPaGS achieves higher PSNR/SSIM, whereas our method obtains lower DRE and higher CIR, indicating better cross-view geometric consistency.

\subsection{Implications for Embodied Intelligence}
\begin{figure}[!t]
    \centering
    \includegraphics[width=0.49\columnwidth]{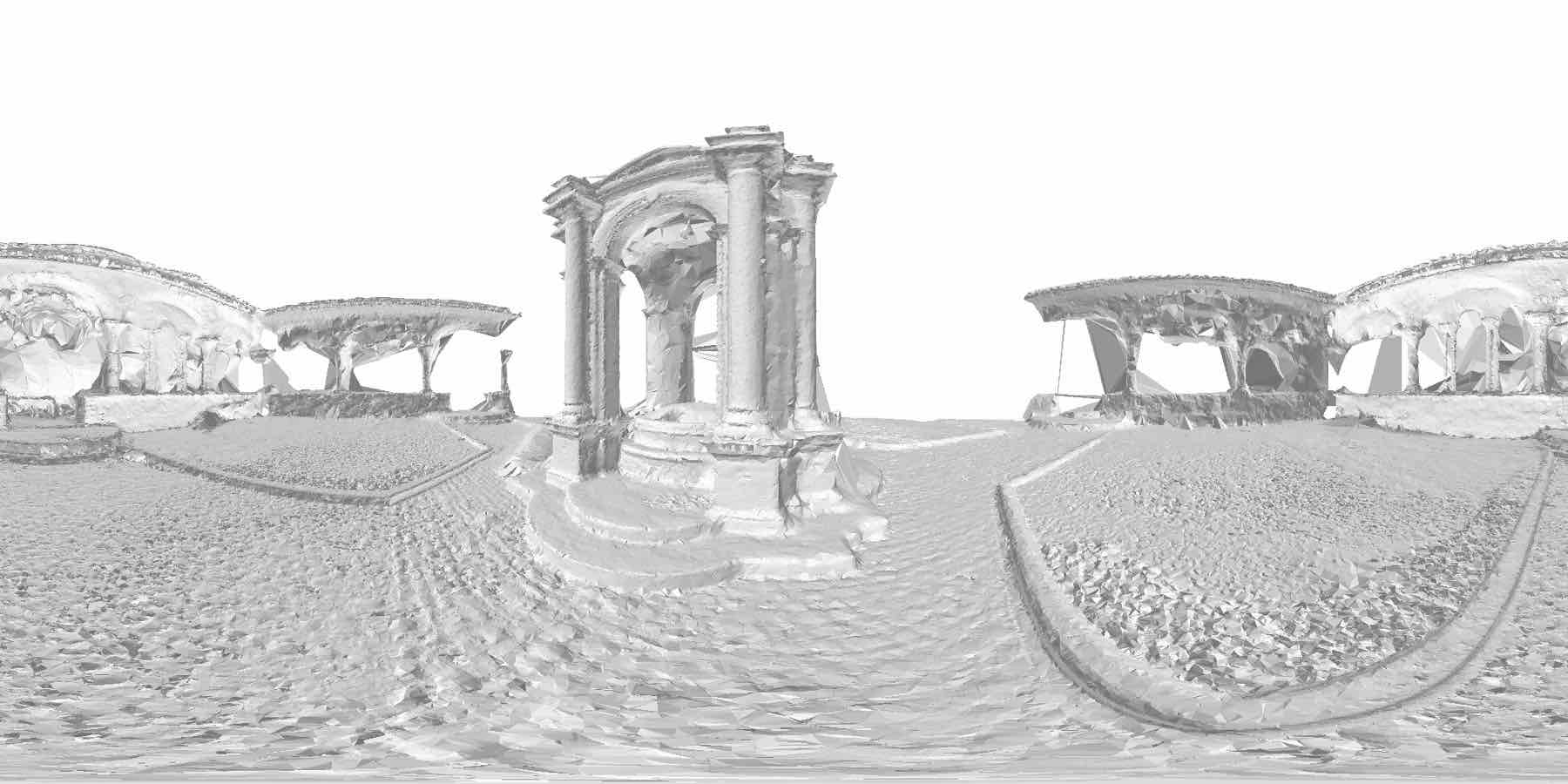}
    \includegraphics[width=0.49\columnwidth]{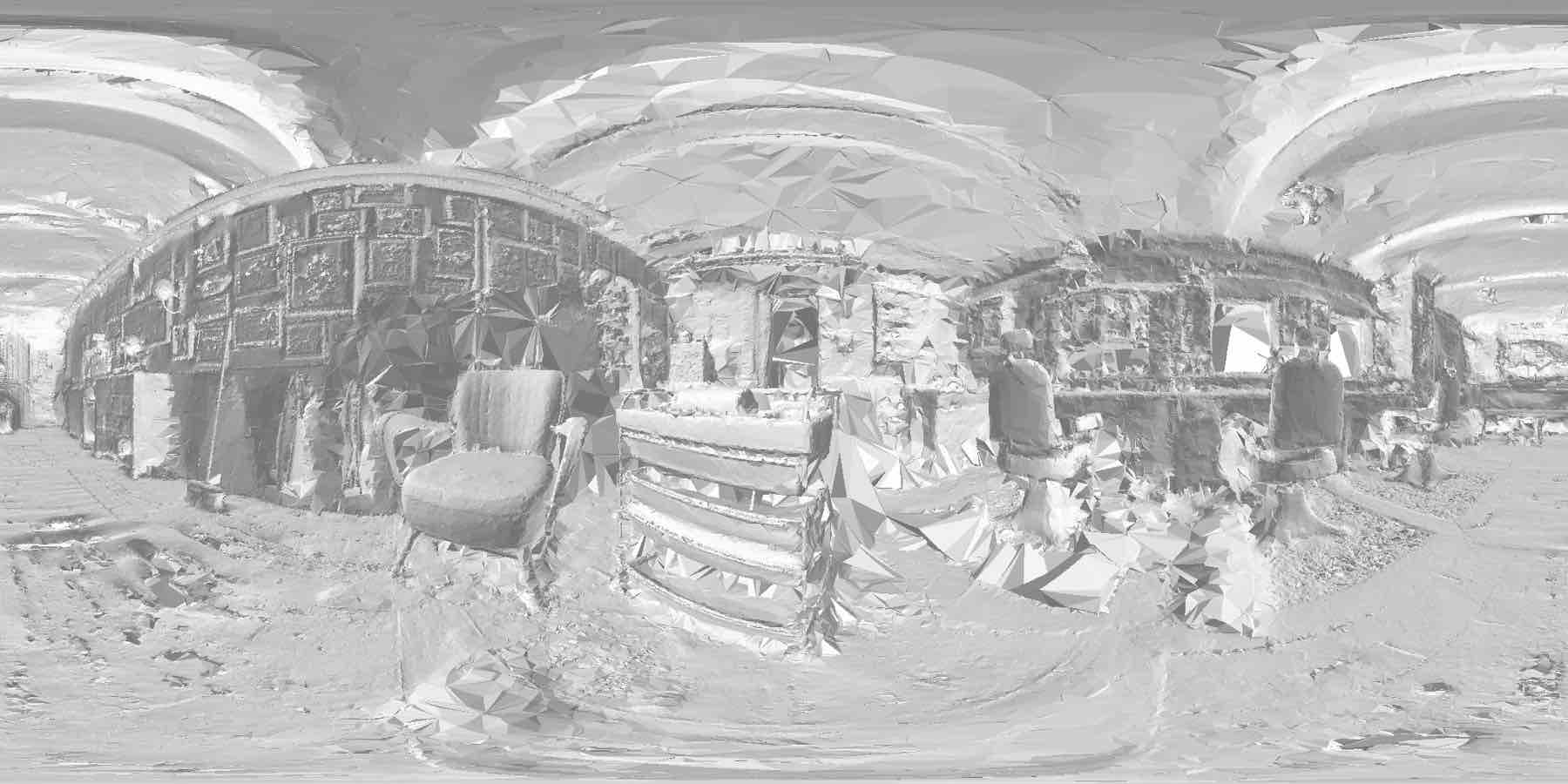}
    \vspace{-3ex}
    \caption{Mesh extraction results on OmniBlender~\cite{choi2023balanced}. 
    Our method reconstructs clean surfaces with fewer holes and reduced texture-induced artifacts.}
    \label{fig:mesh_vis}
    \vspace{-3ex}
\end{figure}

Our method produces geometrically consistent omnidirectional depth, enabling mesh extraction for downstream embodied tasks. As shown in Fig.~\ref{fig:mesh_vis}, reducing texture-induced depth ripples leads to smoother surfaces with fewer artifacts. Although panoramic images have limited spatial resolution due to their large FoV, the reconstructed meshes preserve the main scene layout and can support navigation, obstacle avoidance, and motion planning.

\section{Conclusion}
In this work, we have presented Spherical-GOF, a spherical ray-space GOF sampling and omnidirectional Gaussian rendering framework for ERP panoramas.
By using an ERP-aware footprinting strategy and sphere-metric-consistent geometric regularization, our method achieves competitive photometric quality while significantly improving multi-view geometric consistency, producing cleaner depth and more coherent normal maps with reduced sensitivity to high-frequency appearance textures, and remaining robust under large global panorama rotations.
We further verified its generality on real-world robot-captured omnidirectional data from a panoramic UAV and a quadruped equipped with an annular panoramic camera.

Future work will explore higher-quality and more efficient geometric reconstruction under omnidirectional imaging, including improved geometry priors and faster spherical sampling/rendering strategies.

{\small
\bibliographystyle{IEEEtran}
\bibliography{bib}
}

\end{document}